\documentclass[runningheads]{llncs}
\usepackage{eccv}

\usepackage{eccvabbrv}

\usepackage{graphicx}
\usepackage{booktabs}

\usepackage{hyperref}

\usepackage[accsupp]{axessibility}  %
\usepackage{stfloats}
\usepackage{listings}
\usepackage{mathtools}
\usepackage{amsmath}
\usepackage{placeins}
\usepackage[export]{adjustbox}
\usepackage{subcaption}
\usepackage{graphicx}
\usepackage{pgfplots}
\usepackage{enumitem}
\pgfplotsset{width=10cm,compat=1.9}
\usepackage{multicol}
\usepackage{multirow}
\usepackage{subcaption} 
\usepackage{colortbl}
\usepackage{floatrow}

\makeatletter
\robustify\@latex@warning@no@line
\makeatother
\usepackage{authblk}

\definecolor{turquoise}{rgb}{0.25, 0.88, 0.82}
\definecolor{cvprblue}{rgb}{0.21,0.49,0.74}

\usepackage{orcidlink}

\begin{document}

\title{Identification of Fine-grained Systematic Errors via Controlled Scene Generation}

\author{Valentyn Boreiko\inst{1,2} \and
Matthias Hein\inst{2} \and
Jan Hendrik Metzen\inst{1}}

\authorrunning{V. Boreiko et al.}

\institute{Bosch Center for Artificial Intelligence, Robert Bosch GmbH \and
University of Tübingen}

\maketitle

\begin{abstract}
Many safety-critical applications, especially in autonomous driving, require reliable object detectors. 
They can be very effectively assisted by a method to search for and identify potential failures and systematic errors before these detectors are deployed. Systematic errors are characterized by combinations of attributes such as object location, scale, orientation, and color, as well as the composition of their respective backgrounds.
To identify them, one must rely on something other than real images from a test set because they do not account for very rare but possible combinations of attributes. To overcome this limitation, we propose a pipeline for generating realistic synthetic scenes with fine-grained control, allowing the creation of complex scenes with multiple objects.
Our approach, BEV2EGO, allows for a realistic generation of the complete scene with road-contingent control that maps 2D bird's-eye view (BEV) scene configurations to a first-person view (EGO). In addition, we propose a benchmark for controlled scene generation to select the most appropriate generative outpainting model for BEV2EGO. We further use it to perform a systematic analysis of multiple state-of-the-art object detection models and discover differences between them. Code is available under \url{https://github.com/valentyn1boreiko/BEV2EGO}.
  \keywords{Systematic errors \and Controlled generative models \and Object detectors}
\end{abstract}

\begin{figure}
\centering
\begin{subfigure}{\textwidth}
\centering
\includegraphics[width=1\linewidth,height=1.35\linewidth]{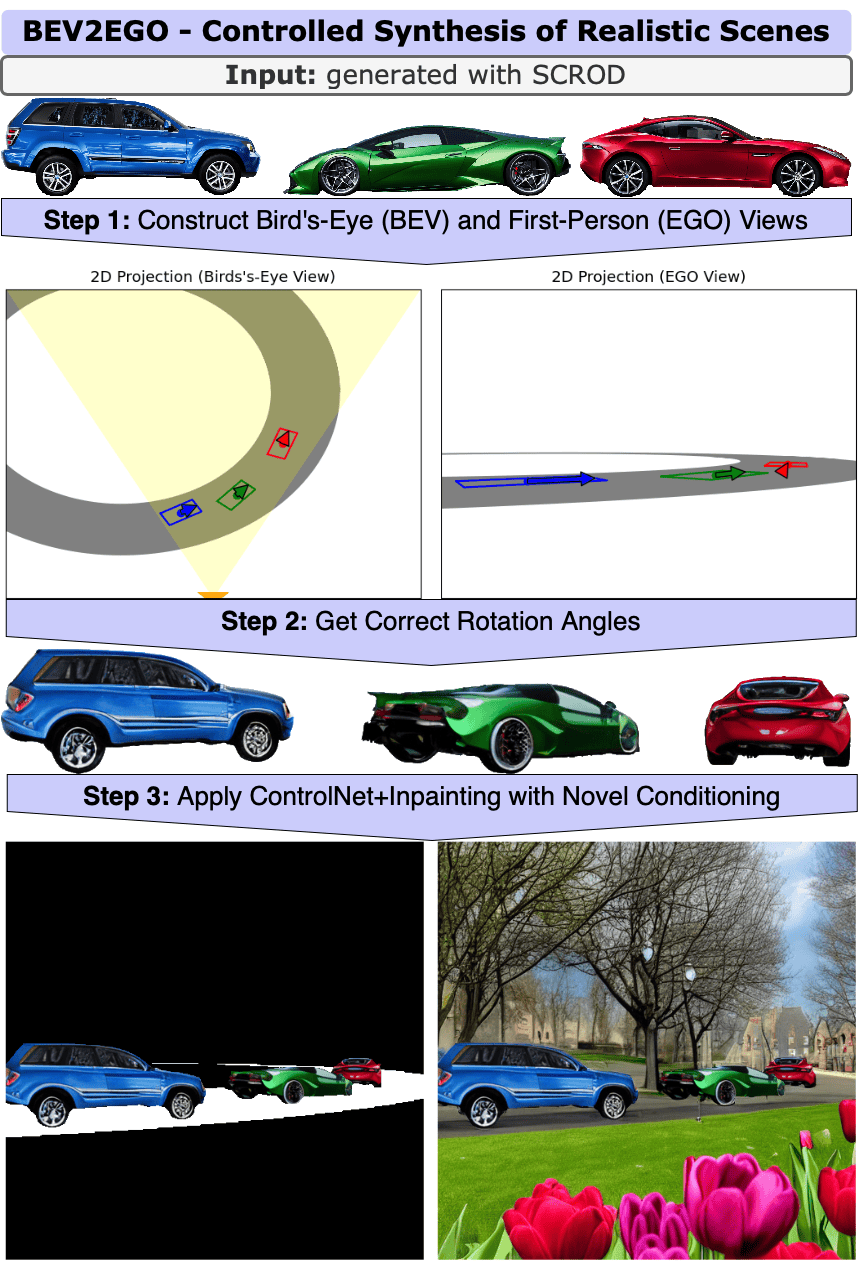}
\end{subfigure}
\caption{\textbf{The BEV2EGO Method for Controlled Synthesis of Realistic Scenes}. Our BEV2EGO method takes images generated with SCROD \cite{boreiko2023SCROD} as input and consists of i) using the camera matrix $\mathbf{P}$, as described in Section~\ref{sec:BEV_method}, to transform a bird's-eye view (BEV) of any scene into a first-person view (EGO); ii) computing the correct rotation angles under which a car is visible for a translated object (see Fig.~\ref{fig:BEV2EGO_base_2D_birds_eye_view} for the discussion on why using the original angle is not appropriate); iii) conditioning with our approach towards realistic outpainting using ControlNet+Inpainting \cite{von-platen-etal-2022-diffusers}, as described in Section~\ref{sec:generative_outpaintings_comparison}.%
}
\label{fig:BEV_teaser}
\end{figure}

\begin{figure}[h]
\scriptsize
\centering
\setlength{\tabcolsep}{0pt}
\begin{tabular}{cc|c|cc}
       \multicolumn{1}{c}{BEV} &
       \multicolumn{1}{c|}{EGO}  & 
       \multicolumn{1}{c|}{Input} & 
        \multicolumn{1}{c}{Car:32\%} &
        \multicolumn{1}{c}{Car: 0\%}
        \\
\hline        
\includegraphics[height=2cm,width=.2\textwidth]{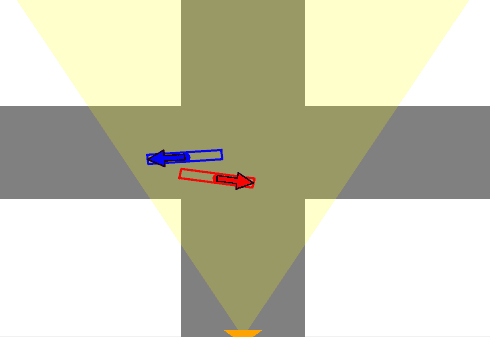} &
\includegraphics[height=2cm,width=.2\textwidth] {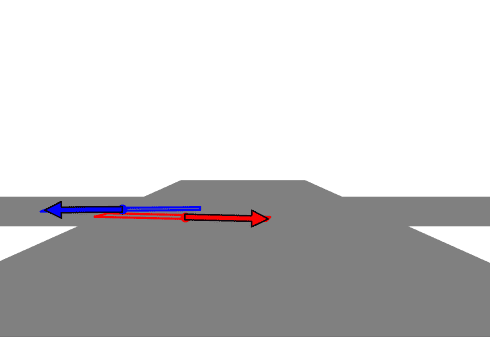} &
\includegraphics[height=2cm,width=.2\textwidth] 
{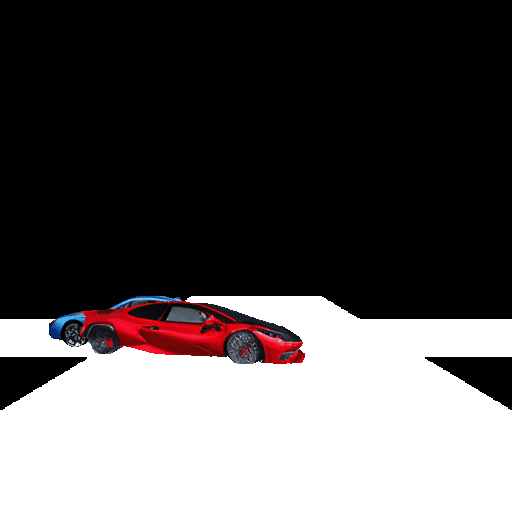} &
\includegraphics[height=2cm,width=.2\textwidth]{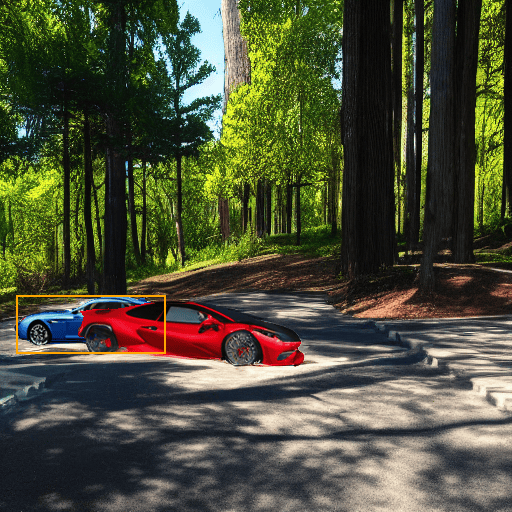}&
\includegraphics[height=2cm,width=.2\textwidth]{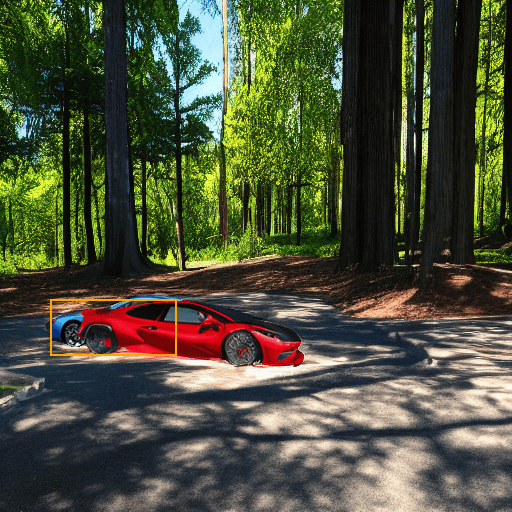} 
\\
\end{tabular}
\caption{\textbf{BEV2EGO enables fine-grained analysis of object detectors}. \textit{The two left images} illustrate the projection from a BEV scene to an EGO perspective according to the pinhole camera model. \textit{The middle image} displays the conditioning used for our outpainting approach with ControlNet+Inpainting~\cite{von-platen-etal-2022-diffusers}. \textit{The two right images} show the outpainted version of the camera view as described in Section~\ref{sec:BEV_method}. Occlusion can significantly degrade the performance of object detectors, such as shown here for \emph{YOLOv5n} (refer to Table~\ref{tab:reproduced_AP} and Section~\ref{sec:systematic_errors} for more details): while in the second image from the right, the probability of the class ``car'' for the partially occluded blue car is 32\%, this probability drastically decreases in the rightmost image with slightly increased occlusion, dropping to 0\%. The resolution here is 512x512, and the prompt used is ``cars are driving in a forest, high resolution, high definition, high quality.''}
\label{fig:BEV2EGO_motivation}
\end{figure}

\section{Introduction}
\label{sec:intro}
The detection of errors in computer vision models has recently made significant progress. Regarding vision model failures, the most attention has been given to classifier failures, especially those identified in ImageNet \cite{peychev2023automated,vasudevan2022does,Metzen_Systematic_Errors,singla2022salient,yannic2023spurious,geirhos2020shortcut,li_2021_discover}. Vasudevan et al. \cite{vasudevan2022does} categorize classifier errors into four types: i) fine-grained errors, ii) fine-grained out-of-vocabulary errors, iii) spurious correlations \cite{singla2022salient,yannic2023spurious}, and iv) non-prototypical errors.

Furthermore, Peychev et al. \cite{peychev2023automated} propose an automated pipeline for identifying different types of errors. These errors can lead to computer vision models performing unevenly across various data subgroups, a critical concern in fields like autonomous driving \cite{Blank_Hueger_Mock_Stauner_2022}. This issue underscores the need to investigate data subgroups where computer vision models are less effective, known as \textit{systematic errors} \cite{eyuboglu_domino:_2022, jain_distilling_2022, Metzen_Systematic_Errors, Tong_Mass_Producing, wiles2022discovering, bitterwolf2022classifiers, boreiko2023SCROD}. Systematic errors have been explored in various computer vision models \cite{Tong_Mass_Producing, eyuboglu_domino:_2022, boreiko2023SCROD, jain_distilling_2022, Metzen_Systematic_Errors, wiles2022discovering, gao2022adaptive}, but research on their discovery in object detectors is limited \cite{boreiko2023SCROD, gao2022adaptive}.

The limitations in capturing all rare but relevant data subgroups in real-world imagery pave the way for synthetic data as an alternative for evaluating object detectors. Synthesis methods generally fall into three categories: i) those utilizing physical simulators and manual 3D designs \cite{leclerc2021three,Resnick2021CBERT_Wokrshop}, which are resource-intensive and not easily scalable; ii) approaches based on neural radiance fields \cite{Xu_2023_CVPR,Niemeyer2020GIRAFFE}, limited in scalability for diverse object configurations and scene types; iii) single-model generative techniques like Stable Diffusion \cite{rombach2021highresolution}, where control over geometric properties through text prompts is challenging.

The pipeline Segment-Control-Rotate-Outpaint-Detect (SCROD) \cite{boreiko2023SCROD} addresses these and other issues related to using a single generative model for the controlled image synthesis. SCROD, however, can only generate a single centered object, which prohibits testing computer vision models on realistic scene compositions that involve having multiple objects rotated and translated in the scene.

We address this challenge by proposing the BEV2EGO method that allows mapping from a 2D BEV configuration of the scene to a realistic image of an EGO without assuming knowledge of the 3D models of the underlying objects (see Section~\ref{sec:BEV_method}, Fig.~\ref{fig:BEV_teaser}, and Fig.~\ref{fig:BEV2EGO_motivation}). While we focus on the application of automated driving and thus on car objects, our method can be extended to other types of objects that are well represented in the distribution of the training data of the generative models we use in our pipeline. With this pipeline, we can identify systematic errors of object detectors (see Section~\ref{sec:systematic_errors}).
\newpage
We make the following contributions:
\begin{itemize}[noitemsep,topsep=0pt,parsep=0pt,partopsep=0pt]
\item In Section \ref{sec:BEV_method}, we propose BEV2EGO, a \emph{novel pipeline for general scene synthesis}, by performing a mapping of a 2D bird's eye view (BEV) to a first-person view (EGO).
\item In Section \ref{sec:generative_outpaintings_comparison}, we propose an approach for outpainting images with ControlNet+Inpainting~\cite{von-platen-etal-2022-diffusers} and a benchmark for comparing different outpainting methods motivated by the criteria introduced in Section \ref{sec:generative_outpaintings_comparison}.
\item In Section \ref{sec:systematic_errors}, we evaluate the state-of-the-art (SOTA) object detectors using our pipeline BEV2EGO by freely positioning two cars in the scene. We show some examples of their systematic errors in Fig.~\ref{fig:systematic_errors_detection} and Fig.~\ref{fig:systematic_errors_detection_VCEs} as well as perform the quantitative evaluation of the overall behavior of these detectors. Results for three cars and more complex scenes are in Appendix~\ref{app:systematic_errors_extended}.
\item In Section \ref{sec:Sym2Real}, we evaluate the Sim2Real gap for our synthetic images by comparing the performance of object detectors on both them and real images.
\end{itemize}

\begin{figure}[hbt!]
\includegraphics[width=1.00\textwidth]{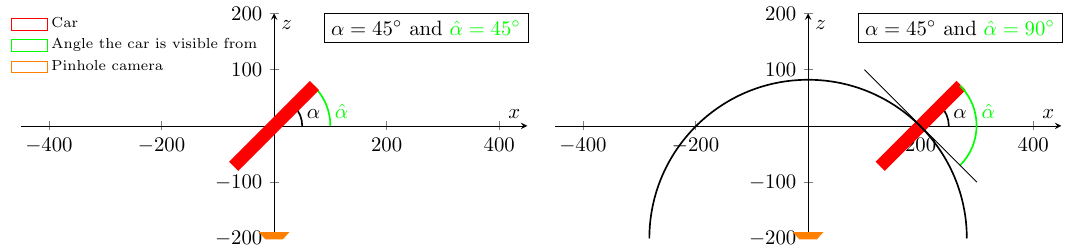}
\caption{\textbf{Computing rotation angles from the camera's reference system to the car's reference system.} \textit{On the left}, the object is centered, and thus the rotation angle $\alpha = 45^{\circ}$ around its axis is identical to the angle $\hat{\alpha}$ from which it is visible. We define $\hat{\alpha}$ as the angle between the tangent to the circle around the camera's pinhole at the center of the car $(m_x, m_z)$ and the line from the camera's pinhole to $(m_x, m_z)$. \textit{On the right}, the object is shifted to the right, resulting in $\hat{\alpha} = 90^{\circ}$, while $\alpha$ remains $45^{\circ}$.}
\label{fig:BEV2EGO_base_2D_birds_eye_view}
\end{figure}

\section{Related Works}
\textbf{Testing Computer Vision Models.}
 Common approaches for this include: i) benchmarking these models on improved test datasets \cite{vasudevan2022does, beyer2020imagenet, tsipras2020imagenet}; ii) evaluating under distribution shifts, such as attribute editing with ImageNet-E \cite{li2023imagenete}, common corruptions with ImageNet-C \cite{hendrycks2018benchmarking}, image renditions with ImageNet-R \cite{hendrycks2021manyfaces}, texture removal with Stylized-ImageNet \cite{geirhos2018}, as well as testing object detectors under weather and style changes with COCO-O \cite{mao2023coco}; iii) doing adversarial attacks \cite{croce2021robustbench}; and iv) testing for spurious correlations \cite{singla2022salient, yannic2023spurious}. However, these approaches do not allow testing the performance on arbitrarily defined subgroups, which can be more fine-grained than class labels.

\noindent \textbf{Identification of Systematic Errors.}
Investigating the behavior of computer vision models on specific subgroups requires a different approach. One solution is creating a dataset that annotates a wide range of attributes or subgroups. Examples include PUG \cite{bordes2023pug} (annotating attributes such as pose, background, size, texture, and lighting), DeepFashion2 \cite{DeepFashion2} (focusing on attributes like occlusion, segmentation, and viewpoint), and ImageNet-X \cite{Idrissi2022ImageNetX} (concentrating on pose and background), along with several examples pertinent to autonomous driving \cite{nuscenes, Marathe_2023_CVPR, MouradDAWN, bdd100k, Cordts2016Cityscapes} (including weather conditions in their metadata labels). However, the scalability of this method is constrained by the practicality of covering all possible attribute combinations.

Another approach involves automatically detecting %
subgroups, which might be prone to errors, potentially covering more attribute combinations without %
specialized datasets. Previous works \cite{eyuboglu_domino:_2022,jain_distilling_2022,Metzen_Systematic_Errors,Tong_Mass_Producing,wiles2022discovering} have mainly focused on classifiers or multi-modal models like CLIP \cite{pmlr-v139-radford21a}. Few studies, such as AdaVision \cite{gao2022adaptive} and SCROD \cite{boreiko2023SCROD}, have explored systematic error detection in object detectors without subgroup annotations. Unlike SCROD, AdaVision involves a human-guided search through existing images, which limits its applicability.

\noindent \textbf{Controlled Image Synthesis.}
Controlled synthesis in image generation can be achieved through methods based on diffusion models, GANs \cite{Shoshan_2021_ICCV, explaining_in_style, gansteerability, shen2020interpreting}, or neural radiance fields \cite{Xu_2023_CVPR, Niemeyer2020GIRAFFE}. Since diffusion models scale effectively to large datasets and offer various forms of control, especially following the introduction of Stable Diffusion \cite{rombach2021highresolution}, our focus is on such methods. They enable diverse guidance forms, such as cropped objects, Canny edges, depth maps, segmentation maps, bounding boxes, and classifier guidance \cite{zhang2023adding,hu2022lora,mou2023t2iadapter,li2023gligen,xue2023freestylenet,cheng2023layoutdiffuse, wallace2023endtoend}. Achieving control over object orientation can be done using 3D assets and simulators or through Stable Diffusion-based models like Zero123 \cite{liu2023zero1to3}, its successor Zero123-XL \cite{objaverseXL}, and other related works \cite{qian2023magic123, liu2023one, gao2023magicdrive, chen2023integrating}. Nevertheless, simultaneously attaining precise control over critical attributes for object detection, such as position, rotation angle, type, color, size of every car, road structure, and background remains a challenge.

\section{Method}
The SCROD pipeline, introduced in \cite{boreiko2023SCROD}, %
enables us to obtain starting images for realistic street synthesis (input to Step $1$ in Fig.~\ref{fig:BEV_teaser}). %
The pipeline relies on various models, including SAM \cite{kirillov2023segany}, ControlNet \cite{zhang2023adding}, Zero123-XL \cite{objaverseXL} (replacing Zero123 \cite{liu2023zero1to3} used in SCROD), and a custom LoRA fine-tuned outpainting model \cite{hu2022lora}, as delineated in \cite{boreiko2023SCROD}. The pipeline's advantage lies in its role as a crucial pre-processing step for various controllable image generation tasks and its extensibility to any object type from the training data distribution of used generative models. Its limitations, however, include: i) only generating one centered object (requiring computation of the rotation angle to allow object translations, as discussed in Fig.~\ref{fig:BEV2EGO_base_2D_birds_eye_view} and Section~\ref{sec:BEV_method}); ii) the LoRA used during outpainting is not constrained to produce realistic scenes, such that, in the context of autonomous driving, a car may appear driving on trees or sea (see Fig.~\ref{fig:outpainting_models_qualitative_comparison}, fifth column from the left, rows one and two); iii) moreover, SCROD does not automatically produce complex scenes, a capability we achieve with BEV2EGO by sampling any scene in BEV and projecting it to the EGO (see examples of this process in Fig.~\ref{fig:BEV_teaser} and Fig.~\ref{fig:BEV2EGO_motivation}).

\subsection{BEV2EGO}\label{sec:BEV_method}
BEV2EGO works by first constructing the scene in BEV, which, in our case, consists of a road and cars on it. Then, we project the scene in the EGO and perform outpainting.

To accurately synthesize scenes from the BEV in the EGO, we utilize camera matrix \cite{hartley_zisserman_2004}. As we want the paper to be self-contained, we list explicitly the steps that we are performing: i) utilizing the \emph{camera matrix} \(\mathbf{P}\) to project the 3D scene correctly onto a 2D plane, and ii) determining the correct \emph{azimuth angle} \(\hat{\alpha}\) from which the object is observed after translations. The latter is necessary as we do not possess a 3D model of the objects to use \(\mathbf{P}\) directly. 

\noindent \textbf{Camera matrix} \(\mathbf{P}\). We focus on two perspectives: the BEV, achieved by a rotation of $90^{\circ}$ around the $x$-axis%
, and the EGO, where no rotation is done. %
Our calibration matrix \(\mathbf{K}\) is simplified as follows:
\begin{equation}
\mathbf{K} = 
\begin{bmatrix}
    f & 0 & 0 \\
    0 & f & 0 \\
    0 & 0 & 1 \\
\end{bmatrix},
\end{equation}
where \(f\) denotes the focal length. 
The final camera matrix is then formulated as \(\mathbf{P} = \mathbf{K} (\mathbf{R} | -\mathbf{R}\mathbf{C})\) \cite{hartley_zisserman_2004}, where \(\mathbf{R}\) is the rotation matrix for camera orientation, and \(-\mathbf{C}\) is the negated camera position vector which translates world coordinates to camera coordinates. Fig.~\ref{fig:BEV2EGO_base_2D_birds_eye_view} demonstrates our approach to mapping the 2D BEV map to the EGO.
More details are in Appendix~\ref{app:BEV2EGO_detailed}.

\noindent \textbf{Azimuth angle}. The SCROD pipeline~\cite{boreiko2023SCROD} presumes a single central object rotated by an angle \(\alpha\) around the $y$-axis, which is perpendicular to the BEV plane, as depicted in Fig.~\ref{fig:BEV2EGO_base_2D_birds_eye_view}. While in the case considered in SCROD, the angle that the car is visible from $\hat{\alpha}$ coincides with $\alpha$, this changes when the car is translated, as can be seen in Fig.~\ref{fig:BEV2EGO_base_2D_birds_eye_view}, right side. Here, we assume the access only to the angle $\alpha$, with which the car is rotated around its axis, and the position of the center of the car $(m_x, m_z)$ after translation. Further, we define $\hat{\alpha}$ to be the angle between the tangent of the circle around the pinhole of the camera at the center of the car $(m_x, m_z)$ and the line from the pinhole of the camera to $(m_x, m_z)$ as can be seen in Fig.~\ref{fig:BEV2EGO_base_2D_birds_eye_view}, on the right. 

\subsection{Generative Outpainting Models}\label{sec:generative_outpaintings_comparison}
\begin{figure}[h]
  \centering
\includegraphics[width=\textwidth]{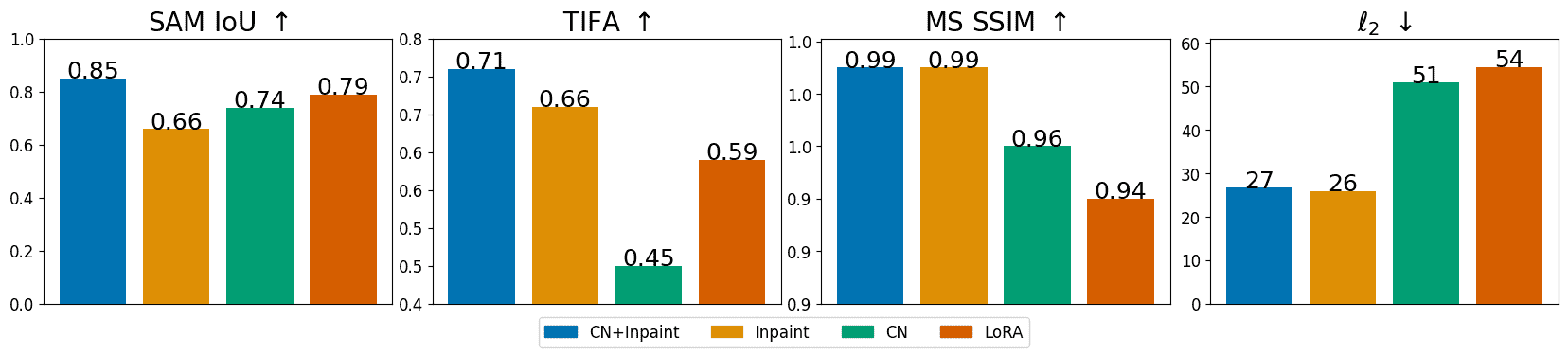}
  \caption{\textbf{Our suggested ``CN+Inpaint'' outpainting method that combines ControlNet (CN) with an Inpainting model \cite{von-platen-etal-2022-diffusers} outperforms alternative outpainting approaches}. We evaluate 120 synthetic scenes in which two cars are randomly positioned on streets. For each scene, we generate nine images with our BEV2EGO method. Examples of these images, which support the quantitative evaluation, are in Fig.~\ref{fig:outpainting_models_qualitative_comparison}. \emph{SAM IoU} measures the degree to which a method preserves the area of the masked cars after the outpainting. \emph{TIFA} \cite{hu2023tifa} measures the fine-grained alignment between the questions about the quality of the image and the generated images after outpainting. We evaluate it on the same set of questions, which we put in Appendix~\ref{app:questions_TIFA}. \emph{MS SSIM} \cite{msssim}, and the $\mathit{l_2}$ norm (of the difference) measure the degree to which a method preserves the area inside of the masked cars after the outpainting.}
\label{fig:outpainting_models_quantitative_comparison}
\end{figure}

Boreiko et al.\,\cite{boreiko2023SCROD} have fine-tuned an outpainting model with LoRA \cite{hu2022lora} from the Inpainting model \cite{von-platen-etal-2022-diffusers} for street scenes in the SCROD pipeline. However, they did not consider conditioning for the road on which cars drive. This leads to unrealistic scenes, as can be seen in our \emph{qualitative analysis} in Fig.~\ref{fig:outpainting_models_qualitative_comparison} (all rows in the fifth column from the left). Furthermore, they did not perform a quantitative comparison of different outpainting models, which we do in our \emph{quantitative analysis} in this section (see also Fig.~\ref{fig:outpainting_models_quantitative_comparison}). 

\noindent \textbf{Models.} For a thorough comparison, we investigate different models that can be adapted for the outpainting of street scenes: i) baseline \emph{Inpainting} model from \cite{von-platen-etal-2022-diffusers}, which receives masked images as an input and has to both preserve the masked images and complete their complement; ii) \emph{ControlNet} \cite{zhang2023adding}, which receives both masked images and the mask of the road as an input, which constrains the street scene to a more realistic one (it is trained as detailed below); iii) \emph{LoRA}, which is fine-tuned from the baseline Inpainting model as in SCROD \cite{boreiko2023SCROD}, using the same dataset as for ControlNet; %
iv) \emph{ControlNet+Inpainting} \cite{von-platen-etal-2022-diffusers}, which is a linear combination of a ControlNet (we set the weight parameter of ControlNet to 1) and the Inpainting model in the weight space.

\noindent \textbf{Training.} Following \cite{boreiko2023SCROD}, we use LoRA fine-tuning \cite{hu2022lora} with rank 32 of the Inpainting model \cite{rombach2021highresolution} and additionally we perform ControlNet training of the SD-v1.5 on the BDD100k \cite{bdd100k}. We use the dataset $(\mathcal{X}, \mathcal{C}_1, \mathcal{C}_2, \hat{\mathcal{X}})$ with $N=5.400$ datapoints, where input to the ControlNet is an element of the Cartesian product $\mathcal{X} \times \mathcal{C}_1 \times \mathcal{C}_2$ of sets of i) cropped cars with background masking $\mathcal{X}$ provided from BDD100k (note that only around $5.400$ images from the whole BDD100k dataset have instance and road segmentations) with augmentations following \cite{boreiko2023SCROD}; ii) captions $\mathcal{C}_1$, automatically generated using 
BLIP-2 \cite{li2023blip2}, OPT-2.7b model, following \cite{boreiko2023SCROD}; iii) road segmentations $\mathcal{C}_2$. Output is a full image from $\hat{\mathcal{X}}$.

\section{Experiments}
\subsection{Analysis of the Generative Outpainting Models}\label{sec:quantitative_analysis_generative}
\begin{figure}[htb!]
  \centering
  \begin{subfigure}{1.0\textwidth}
    \centering
    \includegraphics[width=\linewidth]{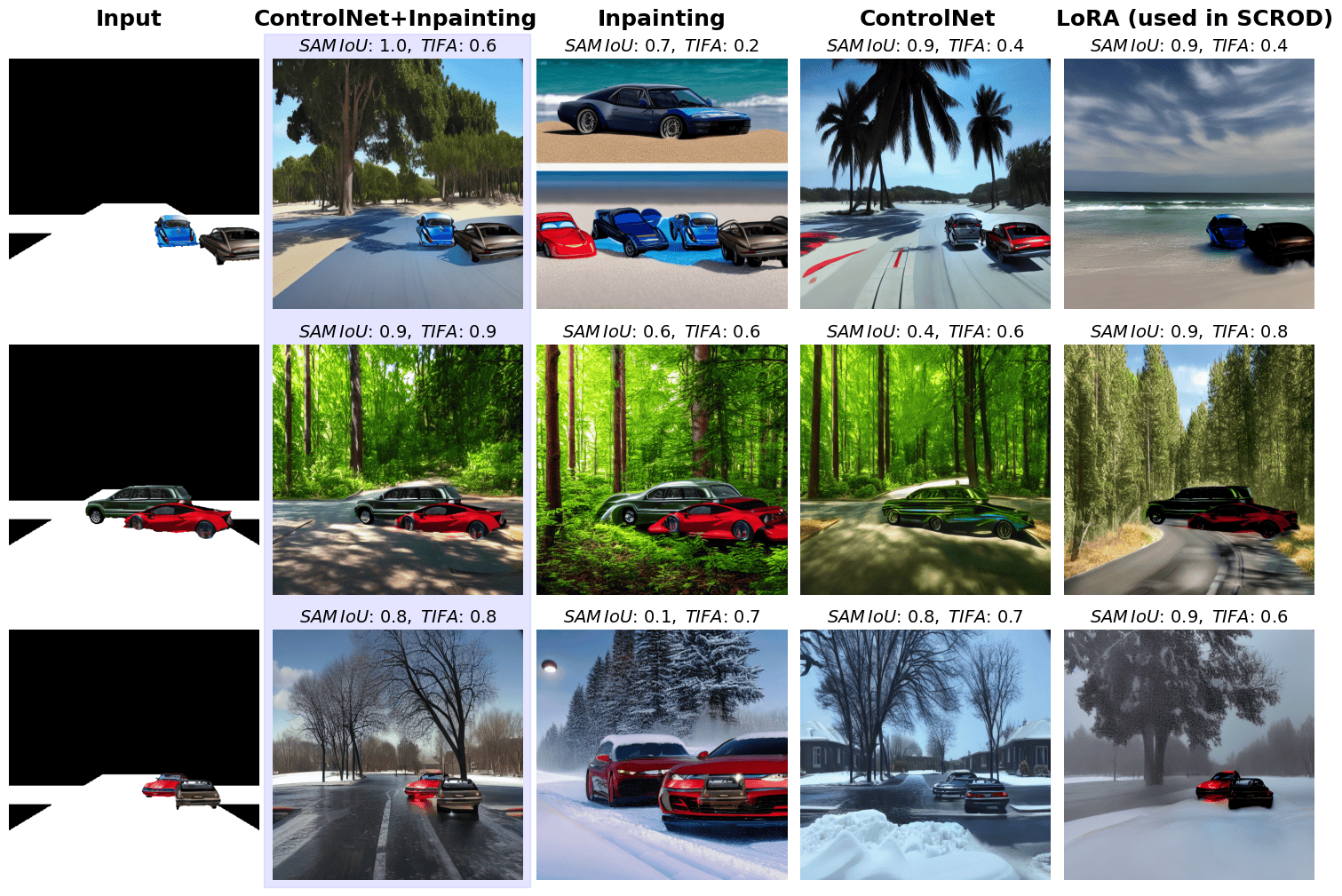}
  \end{subfigure}
  \caption{\textbf{Limitations of generative outpainting methods (Section~\ref{sec:generative_outpaintings_comparison}).} 
\emph{Inpainting} and \emph{LoRA} lack road segmentation control, leading to unrealistic outputs and lower \emph{TIFA} scores, as seen in all the rows of the respective columns. \emph{Inpainting} additionally enlarges main object areas, reducing \emph{SAM IoU} scores. \emph{ControlNet}, although incorporating road segmentation, fails to preserve the image within the mask, causing artifacts and changing the color of the object inside of the mask. These deficiencies in criteria are not present in the proposed \emph{ControlNet+Inpainting} (detailed in Section~\ref{sec:quantitative_analysis_generative}). Quantitative evaluation of these methods is in Fig.~\ref{fig:outpainting_models_quantitative_comparison}.}
\label{fig:outpainting_models_qualitative_comparison}
\end{figure}

\textbf{Goal.} To test object detectors reliably with synthetic data, a suitable outpainting model (Step 3 in our method in Fig.~\ref{fig:BEV_teaser}) should satisfy the following \emph{criteria} (in the context of the autonomous driving application, but these criteria should also be satisfied for more general realistic scene generation cases): a) cars should preserve their boundary after the outpainting and not be hallucinated into bigger objects; b) cars should be placed in a realistic position, and not, e.g., on top of trees; c) the result of outpainting should have as few as possible artefacts in the image. When using the Inpainting model, both a) can be violated (see Fig.~\ref{fig:outpainting_models_qualitative_comparison}, all images in the third column from the left), as well as b) (see Fig.~\ref{fig:outpainting_models_qualitative_comparison}, first and second rows, third column from the left).

\noindent \textbf{Metrics.} To quantify which outpainting method satisfies the described criteria the most, we consider: a) our proposed \emph{SAM IoU} evaluation to measure the first criterion, where we measure the intersection over union (IoU) between the original mask of the non-occluded object and the predicted mask after outpainting with SAM \cite{kirillov2023segany} with one-point guidance. We select this point to be the centroid of the original mask;
b) \emph{TIFA} evaluation \cite{hu2023tifa} to measure the second and the third criterion. This metric uses a visual-question-answering (VQA) (here, mPLUG-large \cite{li2022mplug}) to estimate the faithfulness of an image to the arbitrary set of questions. The questions that we use are related to whether the road is present in the image and whether the cars are generated with the desired attributes (such as color and car type). A complete list of questions is in Appendix~\ref{app:questions_TIFA}; c) both multi-scale structural similarity index measure (MS SSIM) \cite{msssim} and $l_2$ norm (of the difference) between the original masked objects and the result after the outpainting inside of these masks are used to measure the third criterion.

\noindent \textbf{Results.} Using the baseline \emph{Inpainting} model (corresponds to column three from the left in Fig.~\ref{fig:outpainting_models_qualitative_comparison}) can lead to aesthetically pleasing images, which preserves the region inside of the mask well after the outpainting (as can be confirmed by $l_2$ and MS SSIM metrics in Fig.~\ref{fig:outpainting_models_quantitative_comparison}), but which often either increases the object in size (see all rows in the column mentioned above) or generate very unrealistic scenes (see rows one and two in the same column). Both effects make it unreliable to use for the controlled scene synthesis and thus for testing of object detectors.

We observed that training \emph{ControlNet} on $\mathcal{D}$ leads to the preservation of the road control but lack of preservation of the pixel values in the masked region. This can be seen both from SAM IoU, $l_2$, and MS SSIM metrics in Fig.~\ref{fig:outpainting_models_quantitative_comparison} and from images in the fourth column from the left in Fig.~\ref{fig:outpainting_models_qualitative_comparison} (input is in the first column from the left). It can lead to unrealistic artefacts (rows one and two in the fourth column of Fig.~\ref{fig:outpainting_models_qualitative_comparison}). 

\emph{LoRA} fine-tuning %
preserves the pixel values even less (which can be seen from the values of $l_2$ and MS SSIM metrics in Fig.~\ref{fig:outpainting_models_quantitative_comparison}). Moreover, it was not trained with road control. Thus, such outpainting can lead to unrealistic street scenes (rows one and two in the fifth column from the left in Fig.~\ref{fig:outpainting_models_qualitative_comparison}).

Therefore, we use the ControlNet trained as described above together with the \emph{ControlNet+Inpainting} pipeline \cite{von-platen-etal-2022-diffusers}, which allows us to combine both the aesthetically pleasing feature of Inpainting together with the preservation of the region inside of the mask and the control introduced by the ControlNet. This leads to both comparable performance in terms of the preservation of the region inside of the mask (see MS SSIM and $l_2$ subplots of Fig.~\ref{fig:outpainting_models_quantitative_comparison}) as well as outperforming in terms of TIFA and SAM IoU scores. 

\subsection{Systematic Errors in Object Detectors}\label{sec:systematic_errors}

\textbf{Goal.} Now, we apply BEV2EGO pipeline to identify \emph{systematic errors} of object detectors with realistic synthetic data and fine-granular control on the scene level. This would not be possible otherwise as obtaining real images to test object detectors for every possible scene configuration is infeasible. In our setting, \emph{systematic errors} are defined as a set of attributes that uniquely characterize a scene %
and such that a particular object detector consistently performs poorly on images that represent this scene.

\noindent \textbf{Scene generation.} We generate the synthetic dataset $\mathcal{D}$ of 1200 different scenes by controlling the following attributes: \emph{position, rotation angle, type, color, and size} of every car, \emph{road structure}, as well as the \emph{background} using text prompts. During the generation, we uniformly sample a rotation angle $\alpha$ and a center $(m_x, m_z)$ of the first car in BEV. Then we constrain the second car to be close to the first one by sampling i) its rotation angle $\alpha^s$ in the interval $\alpha^s \in [\alpha - 20^{\circ}, \alpha +20^{\circ}]$ and allowing for random flips of $180^\circ$; and ii) its center $(m_x^s, m_z^s)$ such that $m_x^s \in [m_x - 45, m_x - 15] \cup [m_x + 15, m_x + 45]$ and $m_z^s \in [m_z - 45, m_z - 15] \cup [m_z + 15, m_z + 45]$. %
We show more examples as well as examples of more complex scenes with three cars in Appendix~\ref{app:systematic_errors_examples}. %
More details regarding the sampling are in Appendix~\ref{app:BEV2EGO_detailed}. We generate $n_s = 9$ images per scene corresponding to nine different seeds. We do so to identify errors that persist over non-controlled variation in the samples and thus discover true systematic errors.

\noindent \textbf{Metric.} Next, to measure the performance of an object detector on a particular scene represented by $n_s$ seeds, we introduce the Mean Median Score (MMS). The MMS computes the probability per scene (aggregating with a mean over different IoU thresholds and with a median over seeds) that \textit{no} bounding box satisfying the given IoU threshold is of a target class $c=\text{``car''}$. After having computed MMS per scene, we can then average it over any desired group (of scenes): all 1200 scenes or all groups that contain a specific type of car to show the per-group behavior of object detectors.

Concretely, given images $\{x_j\}_{j=1}^{n_s}$ of a particular scene and the lower thresholds $\{\gamma_i\}_{i=1}^{n_t}$ of the IoU of predicted bounding boxes with the ground truth region, we define this metric as follows:
\begin{equation}\label{eq:MMS}
    \frac{1}{n_t}\sum_{i=1}^{n_t} \big ( 1 - \underset{j}{\mathrm{median}}(\max_{k} [ p_\phi(y=c|x_j, \gamma_i, b_k, \psi) ] ) \big),
\end{equation}

for an object detection model (together with the post-processing) $p_\phi$ with weights $\phi$. We obtain a prediction for $x_j$ in a particular bounding box $b_k$, where $k \in \{1, \cdots, B\}$ is an index of a bounding box for a target class of interest $c$ (which is ``car'' in our case) by choosing a desired intersection over union (IoU) threshold for non-maximum suppression (NMS) $\psi$. 
More details are in Appendix~\ref{app:metrics}.

\begin{table}[t]
\centering
\scriptsize 
\begin{tabular}{c | c c c c c}
\hline
Metric & \textbf{FasterRCNN2} & YOLOv5n & YOLOv8n & YOLOv5x6 & RT-DETR-l \\ [0.5ex]
\hline\hline
\textbf{Rep. MAP} $\uparrow$ & 46.7 & 27.5 & 36.6 & \textbf{54.1} & 52.4 \\
\textbf{Rep. mAP@[IoU=0.50]} $\uparrow$ & 67.4 & 45.2 & 52.4 & \textbf{72.0} & 70.6 \\
\textbf{MMS @[IoU=0.50:0.95]} $\downarrow$ & \textbf{34.0} & 64.5 & 56.4 & 46.3 & 35.8 \\
\textbf{MMS @[IoU=0.50]} $\downarrow$ & \textbf{11.2} & 49.5 & 43.2 & 31.5 & 17.6 \\
\hline
\end{tabular}
\\ [1ex]
\begin{tabular}{c | c c c c c}
\hline
$\quad$ \textbf{MMS @[IoU=0.50]} $\downarrow \quad$ & \textbf{FasterRCNN2} & YOLOv5n & YOLOv8n & YOLOv5x6 & RT-DETR-l \\ [0.5ex]
\hline\hline
\textbf{Coupe Car} & \textbf{7.3} & 42.9 & 37.7 & 25.0 & 15.9 \\
\textbf{Sedan} & \textbf{11.3} & 51.6 & 44.8 & 28.5 & 18.4 \\
\textbf{SUV} & \textbf{10.6} & 43.2 & 36.2 & 31.5 & 15.4 \\
\textbf{Smart Car} & \textbf{11.8} & 52.1 & 48.7 & 35.6 & 18.6 \\
\textbf{Sports Car} & \textbf{14.6} & 57.0 & 48.2 & 36.1 & 19.6 \\
\hline
\end{tabular}
\caption{\textbf{Performance of the SOTA object detectors.} \textit{In the first row, upper table:} we reproduce the publicly available results. \textit{In the second row, upper table:} we additionally report  mAP@[IoU=0.50], as it is often used as a standard threshold during the object detection. We use it as well when displaying results on all the Figures. 
\textit{In the third and fourth rows, upper table:} for these models, results on the discovered systematic errors for the scenes with two cars show that \emph{YOLOv5x6} model despite being SOTA in standard metrics, is worse in Mean Median Score (MMS) compared to \emph{FasterRCNN2} and \emph{RT-DETR-l} model.
\textit{Lower table:} A more fine-grained analysis of MMS for different car types. We report results for more models, models trained on Cityscapes \cite{Cordts2016Cityscapes}, as well as more complex scenes with three cars in Appendix \ref{app:systematic_errors_extended}.}

\label{tab:reproduced_AP}
\end{table}

\begin{figure}[hbt!]
    \centering
    \footnotesize %
    \setlength{\tabcolsep}{2pt}
    \renewcommand{\arraystretch}{0.8}
    \begin{tabular}{cc|cc}
        \multicolumn{2}{c|}{Object Detector: \textbf{\textcolor{blue}{YOLOv5n}}} & 
        \multicolumn{2}{c}{Object Detector: \textbf{\textcolor{magenta}{YOLOv5x6}}} \\
        \hline
        \textbf{Car: 0\%} & \textbf{Car: 1\%} & 
        \textbf{Car: 86\%} & \textbf{Car: 87\%} \\
        
        \includegraphics[width=0.22\textwidth]{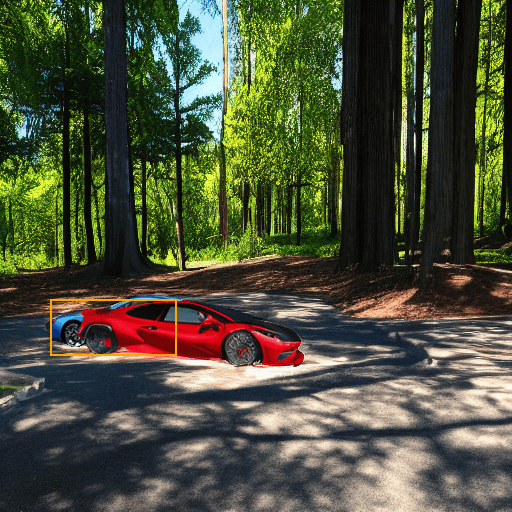} &
        \includegraphics[width=0.22\textwidth]{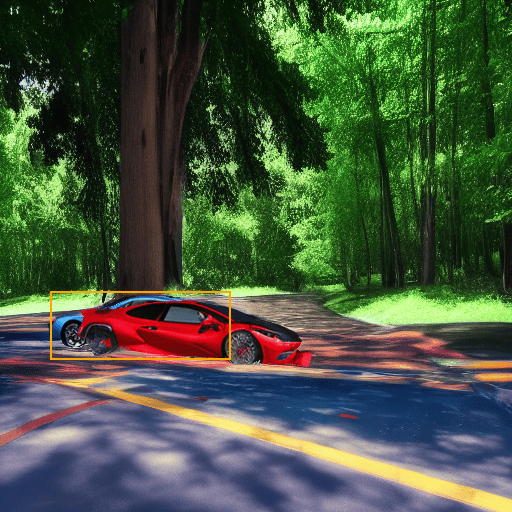} &
        \includegraphics[width=0.22\textwidth]{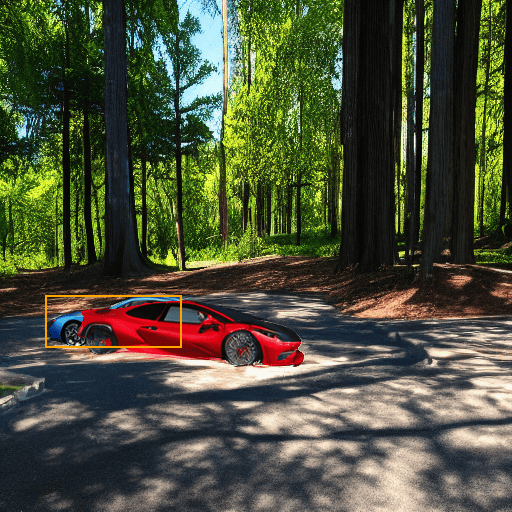} &
        \includegraphics[width=0.22\textwidth]{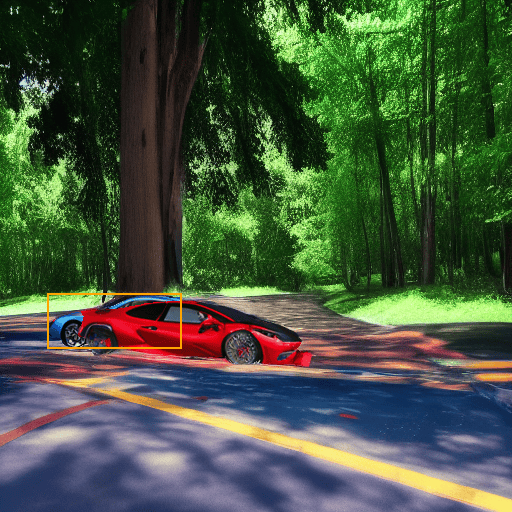} \\
    \hline
    \multicolumn{2}{c|}{Object Detector: \textbf{\textcolor{magenta}{YOLOv5x6}}} & \multicolumn{2}{c|}{Object Detector: \textbf{\textcolor{green}{RT-DETR-l}}}  \\
        \hline
        \textbf{Boat: 13\%} & \textbf{Boat: 37\%} & 
        \textbf{Car: 68\%} & \textbf{Car: 82\%} \\
        \hline
        \includegraphics[width=0.22\textwidth]{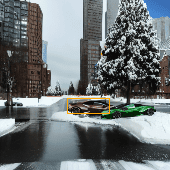} &
        \includegraphics[width=0.22\textwidth]{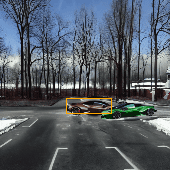} &
        \includegraphics[width=0.22\textwidth]{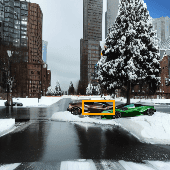} &
        \includegraphics[width=0.22\textwidth]{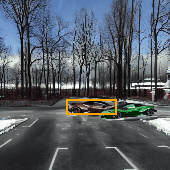}\\
        
    \end{tabular}
    \vspace*{-.25cm}
    \caption{\textbf{Influence of occlusion for YOLOv5x6 vs. YOLOv5n and for YOLOv5x6 vs. RT-DETR-l.} \textit{In the top row}, \emph{YOLOv5x6} performs better than \emph{YOLOv5n}. However, \textit{in the bottom row}, \emph{YOLOv5x6} misclassifies a sports car as a boat, while \emph{RT-DETR-l} classifies it correctly. In the first row, the resolution is 512x512, and the prompt is the same as in Fig.~\ref{fig:BEV2EGO_motivation}. For the second row, the resolution is 170x170, and the prompt is ``cars are driving on snowy street, high resolution, high definition, high quality''.}
    \label{fig:systematic_errors_detection}
\end{figure}

\noindent \textbf{Object detectors.} We measure MMS on $\mathcal{D}$ for the following object detectors:
i) the best object detection model from torchvision \cite{torchvision2016} according to the mean average precision measured at thresholds $\Gamma \coloneqq \{0.5, 0.55, \cdots, 0.95\}$ on COCO~val2017~\cite{lin2015microsoft} (which we will further call MAP, unless specified otherwise) -
FasterRCNN2 \cite{li2021benchmarking}; ii) SOTA object detection models according to their latency and MAP trade-off (YOLOv5 family \cite{yolov5} and YOLOv8 family \cite{yolov8_ultralytics}) - YOLOv5n, YOLOv5x, YOLOv8n, YOLOv8x, and YOLOv5x6 - YOLOv5 model with the best MAP (see Table~\ref{tab:reproduced_AP}) trained on the images of resolution 1280x1280; iii) SOTA object detection models RT-DETR-l and RT-DETR-x \cite{lv2023detrs}; iv) as well as SOTA panoptic segmentation models OneFormer \cite{jain2023oneformer} and Mask2Former \cite{cheng2021mask2former}, trained on a driving dataset Cityscapes \cite{Cordts2016Cityscapes}, which we adapt for the task of 2D detection by constructing the tightest bounding box around the area segmented by the respective algorithm. When choosing models for a driving dataset, we restricted ourselves to recent methods with public weights with permissible licenses that work on 2D images. In the paper, we show analysis on $5$ of them and extend results to the rest in Appendix~\ref{app:systematic_errors_extended}.

\noindent \textbf{Quantitative analysis.} First, we reproduce MAP values of the detection models and their mean average precision at the threshold $0.5$, as this threshold is commonly used. These results are in the top two rows of Table~\ref{tab:reproduced_AP}. The best-performing model according to MAP is YOLOv5x6. While models such as  YOLOv5n and YOLOv8n are the worst in terms of MAP, they are also the fastest in terms of inference, which is relevant in many applications \cite{yolov5}. 

\begin{figure}[t]
    \footnotesize

    \setlength{\tabcolsep}{2pt}
    \begin{tabular}{cc|cc}
        \multicolumn{2}{c|}{Object Detector: \textbf{\textcolor{turquoise}{FasterRCNN2}}} & 
        \multicolumn{2}{c}{Object Detector: \textbf{\textcolor{magenta}{YOLOv5x6}}} \\
        \hline
        \textbf{Boat: 83\%} & \textbf{Boat: 92\%} & 
        \textbf{Car: 33\%} & \textbf{Car: 23\%} \\
        
        \includegraphics[width=0.22\textwidth]{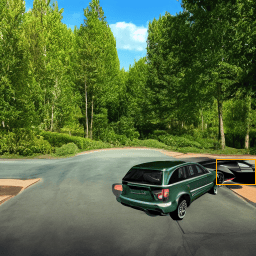} &
        \includegraphics[width=0.22\textwidth]{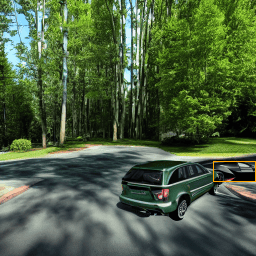} &
        \includegraphics[width=0.22\textwidth]{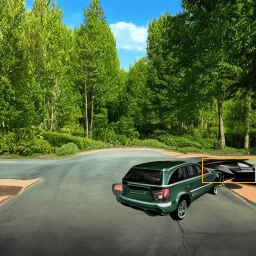} &
        \includegraphics[width=0.22\textwidth]{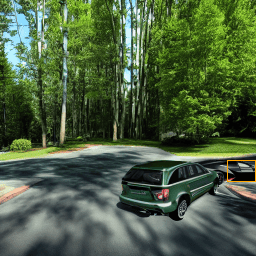} \\
    \hline
    \multicolumn{2}{c|}{Object Detector: \textbf{\textcolor{turquoise}{FasterRCNN2}}} & \multicolumn{2}{c}{Object Detector: \textbf{\textcolor{turquoise}{FasterRCNN2}}}\\
    \hline
    \multicolumn{2}{c|}{Shifting green SUV to the left}
    & \multicolumn{2}{c|}{Changing color of sports car to grey}\\
        \hline
        \textbf{Car: 89\%} & \textbf{Car: 29\%} & 
        \textbf{Car: 99\%} & \textbf{Car: 95\%} \\
        \hline
        \includegraphics[width=0.22\textwidth]{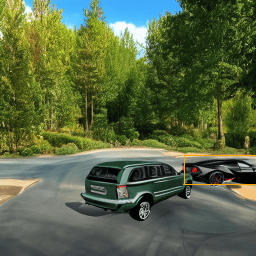} &
        \includegraphics[width=0.22\textwidth]{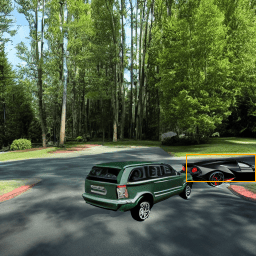} &
        \includegraphics[width=0.22\textwidth]{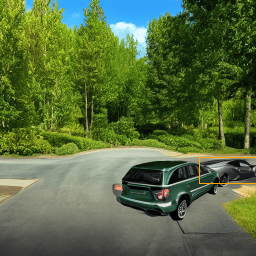} &
        \includegraphics[width=0.22\textwidth]{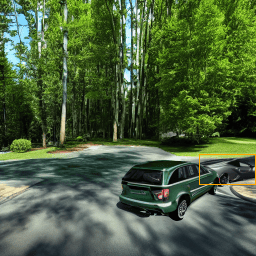} \\
    \end{tabular}
    \vspace*{-.25cm}
    \caption{\textbf{Influence of occlusion and color of the black sports car for FasterRCNN2 vs. YOLOv5x6.} %
    \textit{In the top row}, we show a case where \emph{FasterRCNN2} detects the black sports car occluded by the green SUV as a boat, while \emph{YOLOv5x6} detects it correctly. \textit{In the bottom row}, we show that small changes in the setting from the top row are sufficient to remove the systematic error of \emph{FasterRCNN2}. Here, the resolution is 256x256, and the prompt is ``cars are driving near lake, high resolution, high definition, high quality''.}
    \label{fig:systematic_errors_detection_VCEs}
\end{figure}

Next, we compute MMS at the same threshold values as for MAP in rows three and four of Table~\ref{tab:reproduced_AP}. We can see that surprisingly FasterRCNN2 and RT-DETR-l models outperform the leading MAP model YOLOv5x6 in terms of MMS. This shows the importance of thorough testing of object detectors on the weakest subgroups and not only relying on the standard average-case metrics. %
Additionally, we evaluate MMS on more fine-grained groups, such as different car types. Here, we can see that sports car is the hardest car type for all models but YOLOv8n, for which smart car constitutes the hardest car type to detect. A more detailed analysis is in Appendix~\ref{app:systematic_errors_extended}.

\noindent \textbf{Qualitative analysis.} We have seen already in Fig.~\ref{fig:BEV2EGO_motivation} that YOLOv5n, the fastest but also worst model in terms of both MAP and MMS scores, failed to detect the blue coupe car behind the red sports car. In Fig.~\ref{fig:systematic_errors_detection} (top row), we compare the behavior of YOLOv5n and YOLOv5x6 (the best detector in terms of MAP but not in terms of MMS) on the same scene. We see that occlusion of such degree does not cause a problem for YOLOv5x6.

Next, we search for a systematic error, for which YOLOv5x6 performs worse than RT-DETR-l (a better model in terms of MMS, (see Table~\ref{tab:reproduced_AP}). Because we have seen in Table~\ref{tab:reproduced_AP} that all models have the highest MMS on sports car on average, we focus on this class. In Fig.~\ref{fig:systematic_errors_detection} (bottom row), the sports car is occluded by another one and is generated on a snowy street. This combination of attributes causes YOLOv5x6 to misclassify the detected object as a boat. RT-DETR-l, however, does detect it correctly with high confidence. %

To understand better how systematic errors of object detectors behave under minor scene variations, we do several small changes of a scene corresponding to a systematic error of FasterRCNN2 in Fig.~\ref{fig:systematic_errors_detection_VCEs}. Similarly to what we have seen in Fig.~\ref{fig:BEV2EGO_motivation}, we observe that decreasing the degree of occlusion helps detecting the car correctly. Moreover, simply changing the car's color removes the systematic error and changes the prediction to the correct class with high confidence. More examples of such changes are in Appendix~\ref{app:systematic_errors_examples}.

\section{Evaluation of Sim2Real gap}
\label{sec:Sym2Real}
\begin{figure}[hbt!]
\vspace*{-2em}
  \centering
  \includegraphics[width=\textwidth]{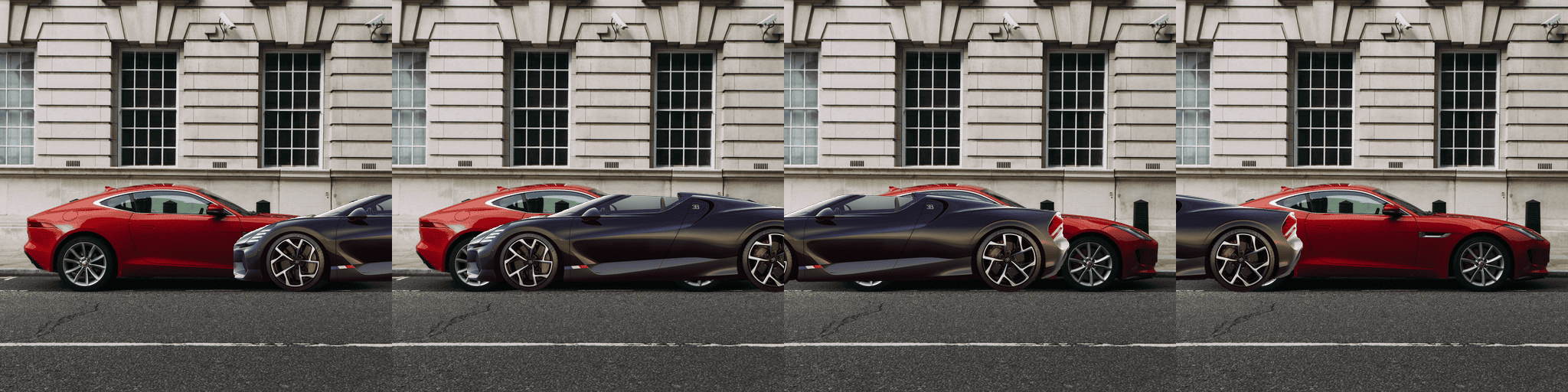}
  \caption{\textbf{Examples of images to measure the Sim2Real gap:} to measure Sim2Real gap, we generate images from real car images, where we can control the degree of occlusion. Quantitative analysis is in the Fig.~\ref{fig:Sim2Real_quantitative}.}
\label{fig:Sim2Real_example}
\end{figure}
\begin{figure}[htbp]
    \centering
    \hspace*{-0.5em}
    \begin{subfigure}[b]{0.495\textwidth}
        \centering  \includegraphics[width=1.05\textwidth,height=3.2cm]{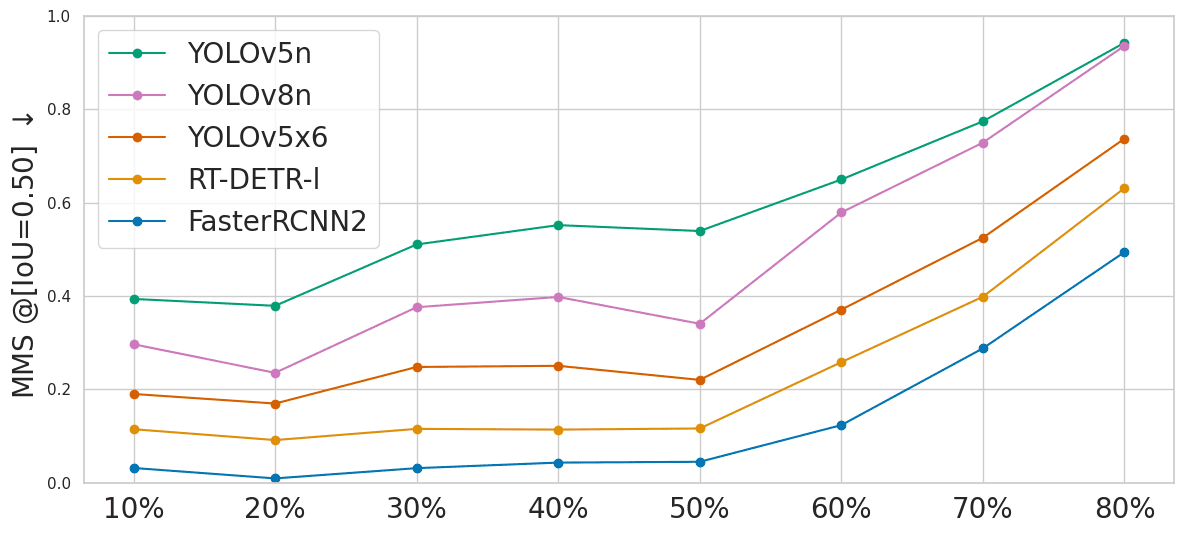}
    \end{subfigure}
    \hfill
    \begin{subfigure}[b]{0.495\textwidth}
        \centering
        \includegraphics[width=\textwidth,height=3.2cm]{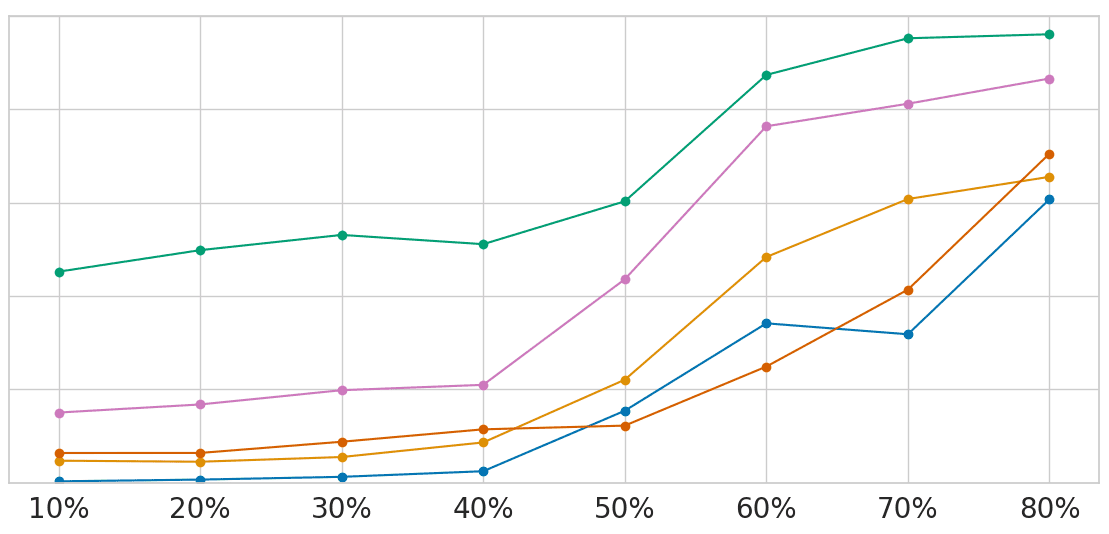}
    \end{subfigure}    
    \caption{\textbf{Measuring the Sim2Real gap:} to estimate the Sim2Real gap, we measure the influence of the degree occlusion on MMS using images as in the Fig.~\ref{fig:Sim2Real_example}. The average Spearman's rank correlation coefficient is $0.91$, which indicates that findings on synthetic data \textit{on the left} translate well to those on the real images \textit{on the right}.}
    \label{fig:Sim2Real_quantitative}
\end{figure}

It is important to quantify the Sim2Real gap and how well the findings discovered on the synthetic scenes transfer to real scenes. This, however, requires both  the availability of \textbf{i)} \textit{data of the real scenes} that match precisely the configurations of our systematic error, and \textbf{ii)} \textit{labels} (such as position, orientation, color, and type of each of the objects on a real scene) that would allow matching synthetic and real images faithfully.
Currently, such a dataset does not exist and this lack of real data is our primary motivation for BEV2EGO. One way, however, is to measure the influence of occlusion on MMS on real scenes and compare it to the influence of occlusion on MMS on synthetic scenes. For this, we take real images of two cars, segment the cars with SAM \cite{kirillov2023segany}, and occlude one car with another. This way, we generate $600$ frames for $6$ different background images with different occluded car types and show $4$ of the frames for one such background image in the Fig.~\ref{fig:Sim2Real_example}. Furthermore, in Fig.~\ref{fig:Sim2Real_quantitative}, we compare quantitatively the influence of occlusion on real images of the side view of two cars to that on synthetic images. For this, we restrict evaluation only to those synthetic cars that were rotated around $y$-axis by angle $\alpha \in [0^{\circ}, 10^{\circ}] \cup [170^{\circ}, 180^{\circ}]$, that is selecting only the cars generated from the side view. The averaged (across occlusion rates) Spearman’s rank correlation coefficient is $0.91$, which indicates that findings on synthetic images generated with BEV2EGO translate well to the real images. \emph{Note,} this approach, however, can be done only for the setting, which requires controlling none of the attributes of interest except for the translation of one object, while BEV2EGO allows to control all the desired attributes, which is why BEV2EGO is an important tool to be able to evaluate the performance of object detectors.

\vspace*{-1em}

\section{Conclusion and Limitations}
\emph{Conclusion.}  We have introduced BEV2EGO, a method to map an abstract 2D bird's-eye view scene definition to realistic image samples in a first-person view. It allows for improved controlled synthesis of realistic scenes and an automatic search of systematic errors of object detectors.

\noindent \emph{Limitations.} Our BEV2EGO relies on external generative models and limitations of those propagate to our method. However, given that overcoming the limitations of those models is an active area of research, BEV2EGO will only improve when incorporating future generative models. 

\noindent \emph{Outlook.} Future directions include extending BEV2EGO to other categories of objects and building upon our analysis to remove systematic errors from existing object detectors, such as using them in training.

\noindent \emph{Potential negative impact.} BEV2EGO can potentially be used for auditing object detection models in self-driving cars or similar applications. As such, one has to be transparent about the configurations used during testing, as otherwise, one can overclaim the robustness of a particular detector to systematic error, which might be detrimental to the end application.

\bibliographystyle{splncs04}

\clearpage
\setcounter{page}{1}

\appendix
\section{Appendix - Overview}
\label{app:appendix_overview}
The appendix has the following structure:
\begin{itemize}
    \item In Appendix~\ref{app:BEV2EGO_detailed}, we provide additional details about the BEV2EGO method used to generate synthetic data.
    \item In Appendix~\ref{app:questions_TIFA}, we present the questions used for the evaluation with TIFA \cite{hu2023tifa} in Section~\ref{sec:generative_outpaintings_comparison}.
    \item In Appendix~\ref{app:metrics}, we elaborate on the MMS.
    \item In Appendix~\ref{app:systematic_errors_extended}, we analyze in more detail the behavior of more object detectors in terms of MMS, compared to the analysis we conducted in Section~\ref{sec:systematic_errors}. Moreover, we repeat the analysis done in Table~\ref{tab:reproduced_AP} for more complex that include three cars and show examples of such scenes.
    \item In Appendix~\ref{app:systematic_errors_examples}, we showcase more examples of the found systematic errors, the influence of small changes on them, and randomly selected scenes generated with BEV2EGO with both two and three cars.
\end{itemize}
\section{Details about BEV2EGO}
\label{app:BEV2EGO_detailed}
In this section, we define the constants used in BEV2EGO.
Specifically, for the EGO view, we position the camera at \((0, h_c, -C)\) in our street scenario, which is a crossroad with a road length of 400 units (and a width of 100 units) for both the vertical and horizontal parts (with respect to the EGO). For the BEV, we place the camera at \((0, h_c, 0)\). We are not restricted to this road type but have fixed it as such because it allows for different driving scenarios.

Note that since we do not have access to the 3D model of the object, we downscale the original height $h_0$ of the object linearly with the distance from the camera $\hat{z}$, that is, $h(\hat{z}) = \frac{h_0 \cdot f}{max(1, \hat{z})}$, assuming a pinhole camera model. Similarly, for the polar angle $\beta$, we compute it using the following heuristic: $\beta(\hat{z}) = \min(15, \max(5, \arctan(\frac{30}{\hat{z}})))$.

For the analysis of systematic errors, we sample the attributes and position of the two cars randomly as described in Section~\ref{sec:BEV_method}. We use the following grid of parameters to sample from, as shown below. Here, whenever the step parameter is indicated, it signifies the discretization step size. Moreover, the values of variables such as \emph{car\_center\_x} and \emph{car\_center\_y} are chosen depending on the car's placement type (variable named \emph{car\_road\_placement}). It can be either horizontal (cars are placed on the horizontal part of the crossroad) or vertical (cars are placed on the vertical part of the crossroad). Additionally, the variable \emph{height\_factor} describes by how much the car will be randomly downscaled, while the variable scale is the factor by which the resolution of the image will be reduced. The last variable is an important parameter as it tests the stability of object detectors in situations where the resolution might be decreased due to sensor limitations.

\definecolor{codegreen}{rgb}{0,0.6,0}
\definecolor{codegray}{rgb}{0.5,0.5,0.5}
\definecolor{codepurple}{rgb}{0.58,0,0.82}
\definecolor{backcolour}{rgb}{0.95,0.95,0.92}

\lstdefinestyle{mystyle}{
   backgroundcolor=\color{backcolour},   
   commentstyle=\color{codegreen},
   keywordstyle=\color{magenta},
   numberstyle=\tiny\color{codegray},
   stringstyle=\color{codepurple},
   basicstyle=\ttfamily\scriptsize,
   breakatwhitespace=false,         
   breaklines=true,                 
   captionpos=b,                    
   keepspaces=true,                 
   numbers=left,                    
   numbersep=5pt,                  
   showspaces=false,                
   showstringspaces=false,
   showtabs=false,                  
   tabsize=2
}

\begin{lstlisting}[style=mystyle]
car_rotation_angle:
  - start: -90
  - end: 90
  - step: 10
car_center_y:
  - start_horizontal: -30
  - end_horizontal: 0 
  - start_vertical: -100
  - end_vertical: 0
  - step: 1
car_center_x:
  - start_horizontal: -100
  - end_horizontal: 100
  - start_vertical: -30
  - end_vertical: 30
  - step: 1
scale:
  - start: 1.0
  - end: 4.0
  - step: 0.5
height_factor:
  - start: 0.8
  - end: 1.2
  - step: 0.1
car_type:
  - sedan
  - sports car
  - smart car
  - coupe car
  - SUV
background_simple:
  - in forest
  - on beach
  - in city
  - on snowy street
  - on highway
  - near lake
color:
  - white
  - black
  - grey
  - yellow
  - red
  - blue
  - green
  - brown
  - pink
  - orange
  - purple
car_road_placement:
  - vertical
  - horizontal
\end{lstlisting}

\section{Questions used in TIFA evaluation}\label{app:questions_TIFA}
TIFA \cite{hu2023tifa} evaluates generative models with a VQA model by reporting the frequency of correctly answering a prepared set of questions for images from a generative model. As already indicated in Section~\ref{sec:generative_outpaintings_comparison}, we use mPLUG-large \cite{li2022mplug} as the VQA model. The questions in our case check if the road and target car types are correctly generated. They are generated with the code provided below and are used for the evaluation of street scenes with two cars.
\newpage

\lstset{style=mystyle}

\begin{lstlisting}
def question_base(car_type_1, car_type_2, car_color_1, car_color_2, common_colors, car_types):
    caption = f'{car_color_1} {car_type_1} and {car_color_2} {car_type_2} are driving on a road.'

    return [
        {
            'caption': caption,
            'element': car_type_1,
            'question': f'is there a {car_type_1}?',
            'choices': ['yes', 'no'],
            'answer': 'yes',
            'element_type': 'object'
        },
        {
            'caption': caption,
            'element': car_type_2,
            'question': f'is there a {car_type_2}?',
            'choices': ['yes', 'no'],
            'answer': 'yes',
            'element_type': 'object'
        },
        {
            'caption': caption,
            'element': 'road',
            'question': 'is there an asphalted road?',
            'choices': ['yes', 'no'],
            'answer': 'yes',
            'element_type': 'location'
        },
        {
            'caption': caption,
            'element': 'road',
            'question': 'what type of path is this?',
            'choices': ['trail', 'grass', 'sand', 'asphalted road', 'dirt', 'snow', 'mountain', 'river', 'sea', 'forest'],
            'answer': 'asphalted road',
            'element_type': 'location'
        },
        {
            'caption': caption,
            'element': car_color_1,
            'question': f'is the {car_type_1} {car_color_1}?',
            'choices': ['yes', 'no'],
            'answer': 'yes',
            'element_type': 'color'
        },
        {
            'caption': caption,
            'element': car_color_1,
            'question': f'what color is the {car_type_1}?',
            'choices': common_colors,
            'answer': car_color_1,
            'element_type': 'color'
        },
        {
            'caption': caption,
            'element': car_color_2,
            'question': f'is the {car_type_2} {car_color_2}?',
            'choices': ['yes', 'no'],
            'answer': 'yes',
            'element_type': 'color'
        },
        {
            'caption': caption,
            'element': car_color_2,
            'question': f'what color is the {car_type_2}?',
            'choices': common_colors,
            'answer': car_color_2,
            'element_type': 'color'
        },
        {
            'caption': caption,
            'element': 'driving',
            'question': f'are the {car_type_1} and {car_type_2} driving?',
            'choices': ['yes', 'no'],
            'answer': 'yes',
            'element_type': 'activity'
        }
    ]
\end{lstlisting}
\section{Details about the Mean Median Score (MMS)}\label{app:metrics}
The MMS metric, introduced in Equation~\ref{eq:MMS}, aggregates the predicted confidences in the target class ``car'' by an object detector over the nine seeds with the median, and over all the lower thresholds of the IoU of the predicted bounding boxes with the ground truth region in $\Gamma$, as described in Section~\ref{sec:systematic_errors}, with the mean. We perform the aggregation with the median over the seeds for a more robust estimation with respect to outliers. However, we estimate over the thresholds with the mean, as we want potential outliers at threshold $0.5$ to dominate the score, as this threshold is relevant in applications. We report in the table MMS aggregated both over all the thresholds from $\Gamma$ and only at the threshold $\gamma_0=0.5$, while we search for the systematic errors that are displayed in Figs.~\ref{fig:systematic_errors_detection}, \ref{fig:systematic_errors_detection_VCEs}, \ref{fig:systematic_errors_detection_extended}, and \ref{fig:systematic_errors_detection_VCEs_extended} based only on MMS at the threshold $\gamma_0=0.5$, as we use this threshold when choosing the predicted bounding box on the displayed images.

Namely, to display a systematic error, we take all the bounding boxes with the IoU greater than or equal to $0.5$ and choose the prediction with the highest predicted confidence. The ground truth bounding box is the tightest bounding box around the respective ``car'' object. When filtering bounding boxes this way for occluded objects, we take IoU as a maximum over the IoU of the predicted bounding boxes with the ground truth region and the IoU of the predicted bounding box with the visible ground truth region. We do this, as some object detectors might correctly detect the ``car'' object only for the visible ground truth region, while others extend the bounding box to the full ground truth size of the occluded object.

\section{Extended analysis of systematic errors in object detectors}\label{app:systematic_errors_extended}
Here, we provide a more detailed quantitative analysis of the systematic errors, thus extending the analysis in Section~\ref{sec:systematic_errors}. As in Section~\ref{sec:systematic_errors}, we provide an evaluation for 1200 different scenes generated with BEV2EGO with attributes sampled as described in Section~\ref{sec:BEV_method} and Appendix~\ref{app:BEV2EGO_detailed}. We begin with a quantitative analysis and then display some of the systematic errors, as well as the influence of small changes in attributes on object behavior for the systematic errors in question in Appendix~\ref{app:systematic_errors_extended}.

\noindent \textbf{MMS for more object detectors.} First, in Table~\ref{tab:reproduced_AP_complete}, we complete the analysis started in the Table~\ref{tab:reproduced_AP}, by doing it for more detectors. We do the following observations: \textbf{i)} OneFormer and Mask2Former trained on realistic driving scenes respond well to our synthetic images, indicated by low MMS scores compared to object detectors trained on COCO; \textbf{ii)} they do not consistently outperform FasterRCNN2, the best object detector according to MMS, which is trained on COCO. This experiment indicates that our proposed pipeline can be useful to investigate the biases of models trained on different datasets. 
\begin{table}[t]
\centering
\scriptsize 
\begin{tabular}{c | c c c c c}
\hline
Metric & \textbf{FasterRCNN2} & YOLOv5n & YOLOv8n & YOLOv5x6 & RT-DETR-l \\ [0.5ex]
\hline\hline
\textbf{Rep. MAP} $\uparrow$ & 46.7 & 27.5 & 36.6 & \textbf{54.1} & 52.4 \\
\textbf{Rep. mAP@[IoU=0.50]} $\uparrow$ & 67.4 & 45.2 & 52.4 & \textbf{72.0} & 70.6 \\
\textbf{MMS @[IoU=0.50:0.95]} $\downarrow$ & \textbf{34.0} & 64.5 & 56.4 & 46.3 & 35.8 \\
\textbf{MMS @[IoU=0.50]} $\downarrow$ & 11.2 & 49.5 & 43.2 & 31.5 & 17.6 \\
\hline
\end{tabular}
\\ [1ex]
\begin{tabular}{c | c c c c c}
\hline
$\quad$ \textbf{MMS @[IoU=0.50]} $\downarrow \quad$ & \textbf{FasterRCNN2} & YOLOv5n & YOLOv8n & YOLOv5x6 & RT-DETR-l \\ [0.5ex]
\hline\hline
\textbf{Coupe Car} & \textbf{7.3} & 42.9 & 37.7 & 25.0 & 15.9 \\
\textbf{Sedan} & 11.3 & 51.6 & 44.8 & 28.5 & 18.4 \\
\textbf{SUV} & 10.6 & 43.2 & 36.2 & 31.5 & 15.4 \\
\textbf{Smart Car} & 11.8 & 52.1 & 48.7 & 35.6 & 18.6 \\
\textbf{Sports Car} & 14.6 & 57.0 & 48.2 & 36.1 & 19.6 \\
\hline
\end{tabular}
\\ [2ex]
\begin{tabular}{c | c c c c c}
    \hline
    Metric & YOLOv5x & YOLOv8x & RT-DETR-x & OneFormer & Mask2Former\\ [0.5ex]
    \hline\hline
    \textbf{Rep. MAP} $\uparrow$ & 49.8 & 52.8 & 54.0 & - & - \\
    \textbf{Rep. mAP@[IoU=0.50]} $\uparrow$ & 68.2 & 69.9 & 72.0 & - & - \\
    \textbf{MMS @[IoU=0.50:0.95]} $\downarrow$ & 47.5 & 46.3 & 35.5 & 40.3 & 47.4  \\
    \textbf{MMS @[IoU=0.50]} $\downarrow$ & 33.1 & 32.6 & 17.9 & \textbf{8.6} & 14.7 \\
    \hline
    \end{tabular}
   \\ [1ex] 
    \centering
    \begin{tabular}{c | c c c c c}
    \hline
    \textbf{MMS @[IoU=0.50]} $\downarrow \quad \quad $ & YOLOv5x & YOLOv8x & RT-DETR-x & OneFormer & Mask2Former \\ [0.5ex]
    \hline\hline
    \textbf{Coupe Car} & 28.0 & 27.0 & 16.5 & 8.5 & 14.6 \\
    \textbf{Sedan} & 31.0 & 30.4 & 18.0 & \textbf{10.6} & 16.1 \\
    \textbf{SUV} & 31.2 & 29.7 & 15.6 & \textbf{4.1} & 5.2 \\
    \textbf{Smart Car} & 36.4 & 36.7 & 19.3 & \textbf{9.1} & 19.6 \\
    \textbf{Sports Car} & 38.6 & 38.4 & 20.1 & \textbf{10.7} & 17.9 \\
    \hline
\end{tabular}
\caption{\textbf{Performance of all the investigated SOTA object detectors.} We insert Table~\ref{tab:reproduced_AP} from the main paper such that the Appendix is self-contained. \textit{In the first row, first and third tables:} we reproduce the publicly available results except for OneFormer and Mask2Former as they are panoptic segmentation networks. \textit{In the second row, first and third tables:} we additionally report  mAP@[IoU=0.50], as it is often used as a standard threshold during object detection. We use it as well when displaying results on all the Figures. 
\textit{In the third and fourth rows, first and third tables:} for these models, results on the discovered systematic errors show that \textbf{YOLOv5x6} model despite being SOTA in standard metrics, is worse in Mean Median Score (MMS) compared to \textbf{FasterRCNN2} and both \textbf{RT-DETR} models. 
\textit{Second and fourth tables:} A more fine-grained analysis of MMS for different car types.}
\label{tab:reproduced_AP_complete}
\end{table}

\begin{figure}[htbp]
    \centering
    \hspace*{-0.5em}
    \begin{subfigure}[b]{0.495\textwidth}
        \centering          \includegraphics[width=1.05\textwidth,height=3.2cm]{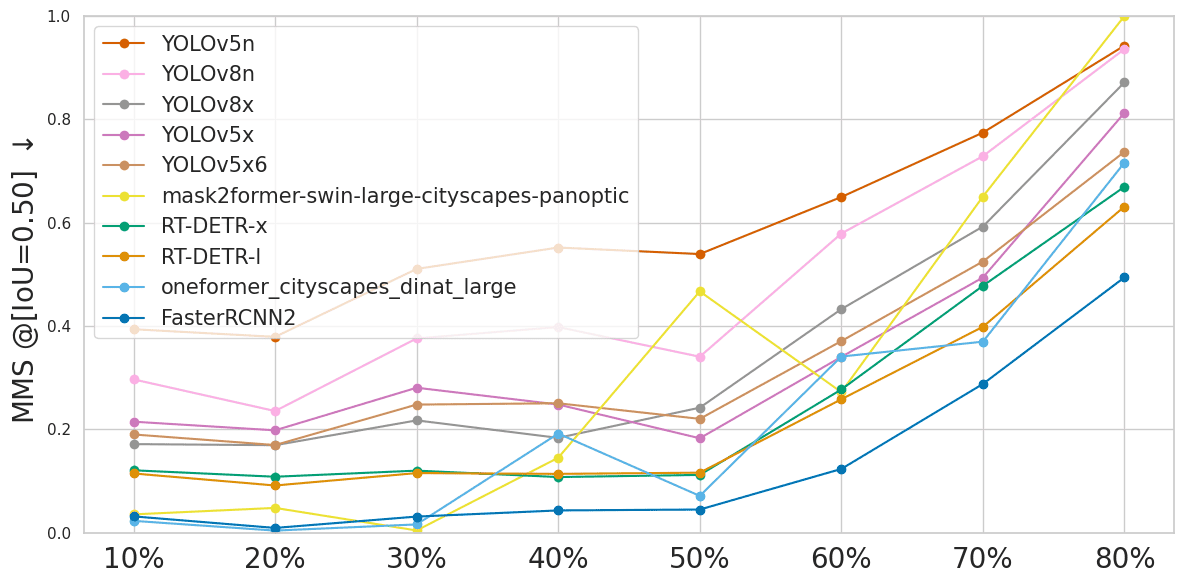}
    \end{subfigure}
    \hfill
    \begin{subfigure}[b]{0.495\textwidth}
        \centering
        \includegraphics[width=\textwidth,height=3.2cm]{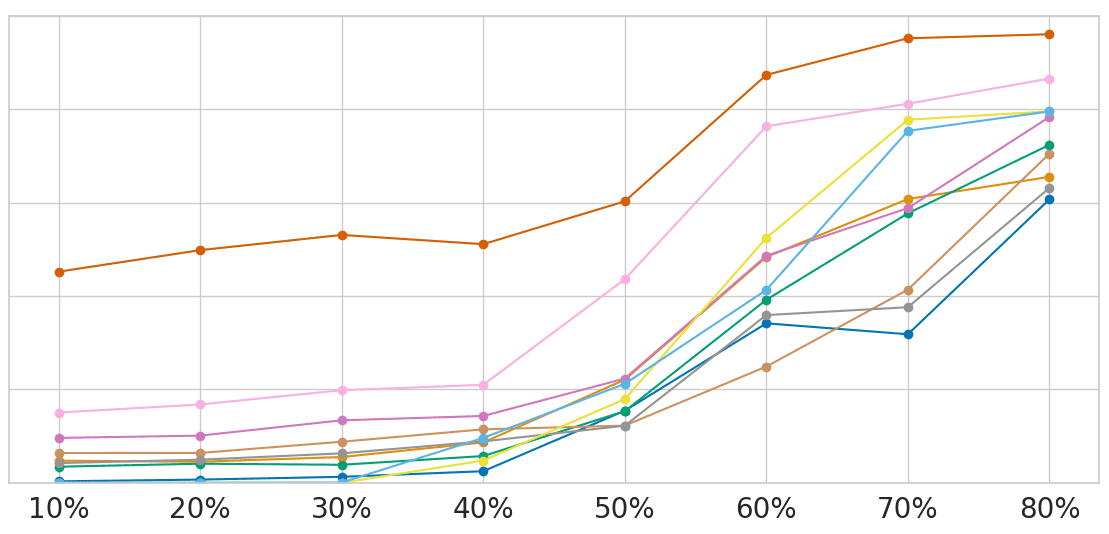}
    \end{subfigure}    
    \caption{\textbf{Measuring the Sim2Real gap for all the investigated SOTA object detectors:} we repeat the experiment presented in Fig.~\ref{fig:Sim2Real_quantitative}, but for all the investigated object detectors. The average Spearman's rank correlation coefficient is 0.66, which indicates that findings on synthetic data \textit{on the left} translate well to those on the real images \textit{on the right} also when using more object detectors.}
    \label{fig:Sim2Real_quantitative_extended}
\end{figure}

\noindent \textbf{Sim2Real gap for more object detectors.} Furthermore, we repeat the experiment that measures the Sim2Real gap with more object detectors in Fig.~\ref{fig:Sim2Real_quantitative_extended}. Here, we see that even though the average over the object detectors Spearman's rank correlation coefficient of 0.66 is lower, it still indicates that the findings from the synthetic images translate well to those on the real images.

\begin{figure}[htb!]
\centering
\includegraphics[width=\linewidth]{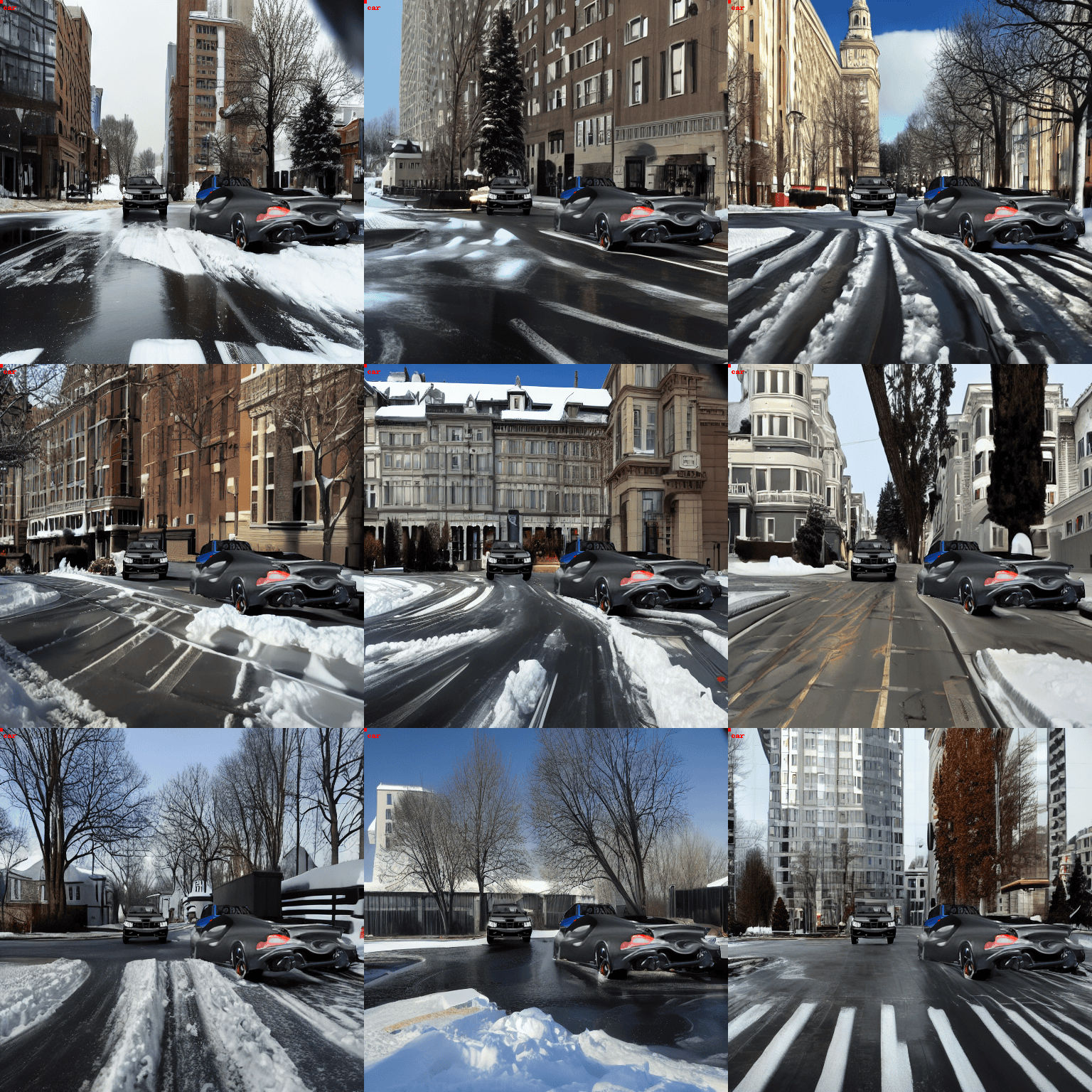}
\caption{\textbf{Selected scene containing three cars generated with BEV2EGO.}}
\label{fig:selected_scenes_3_cars_0}
\end{figure}

\begin{figure}[htb!]
\centering
\includegraphics[width=\linewidth]{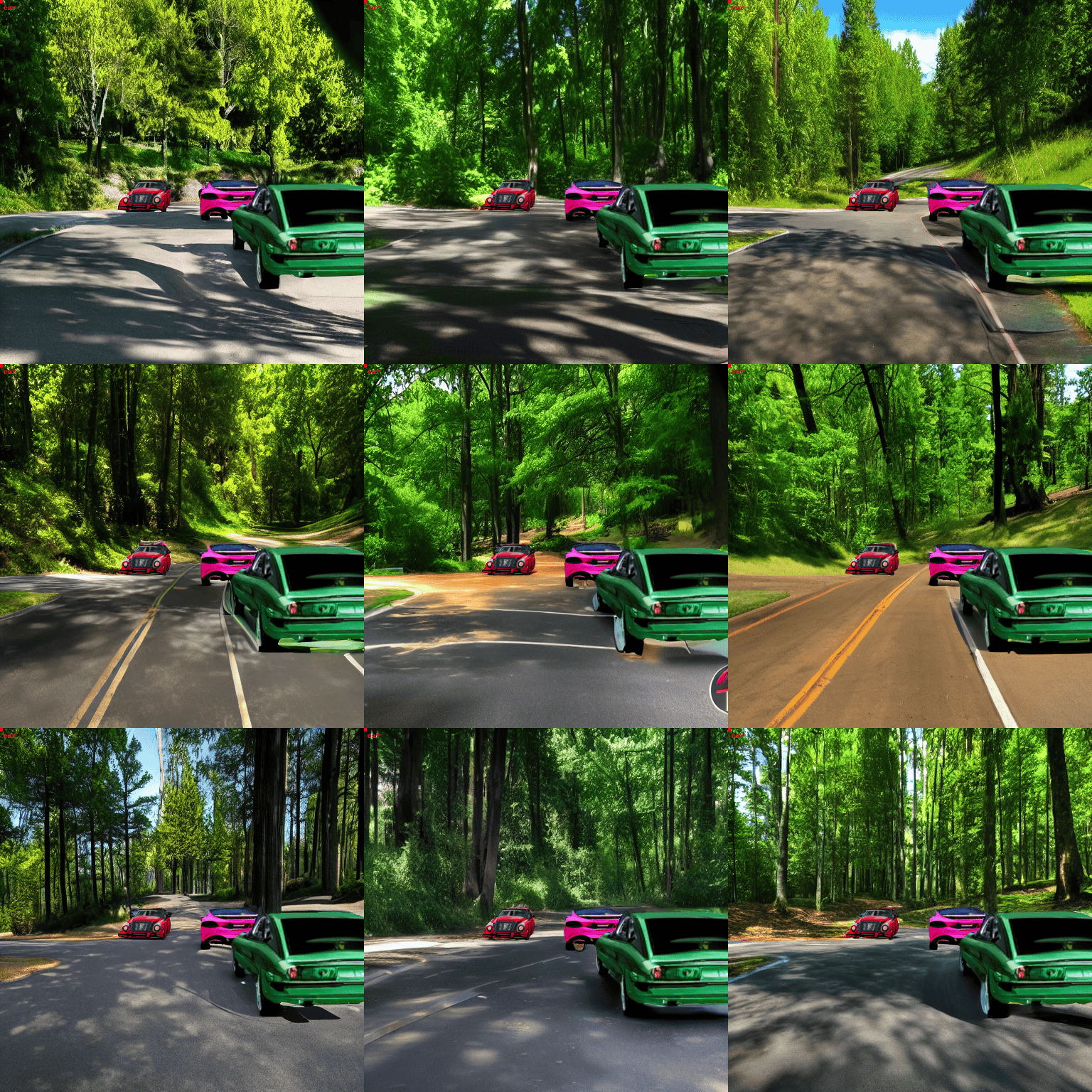}
\caption{\textbf{Selected scene containing three cars generated with BEV2EGO.}}
\label{fig:selected_scenes_3_cars_1}
\end{figure}

\begin{figure}[htb!]
\centering
\includegraphics[width=\linewidth]{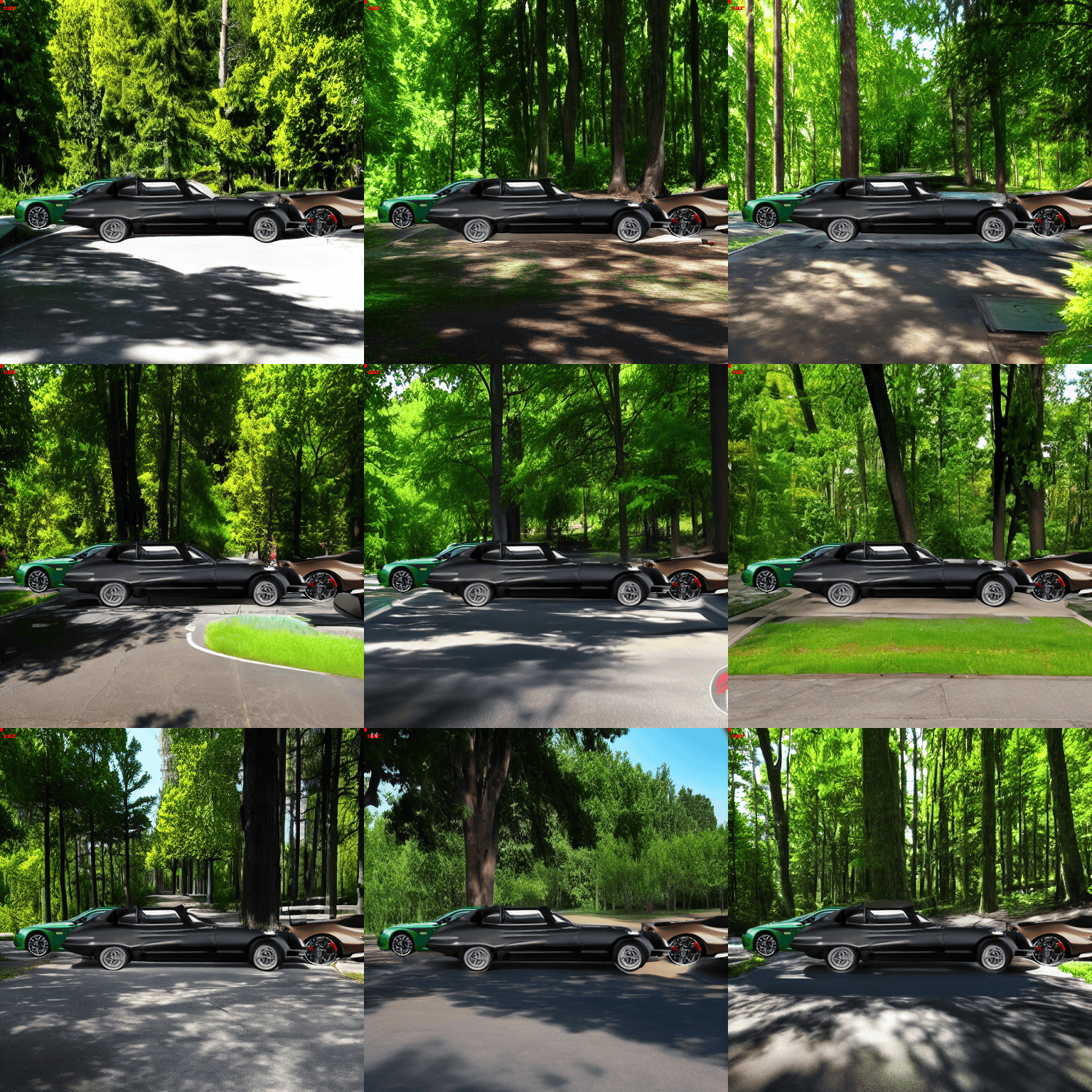}
\caption{\textbf{Selected scene containing three cars generated with BEV2EGO.}}
\label{fig:selected_scenes_3_cars_2}
\end{figure}

\begin{figure}[htb!]
\centering
\includegraphics[width=\linewidth]{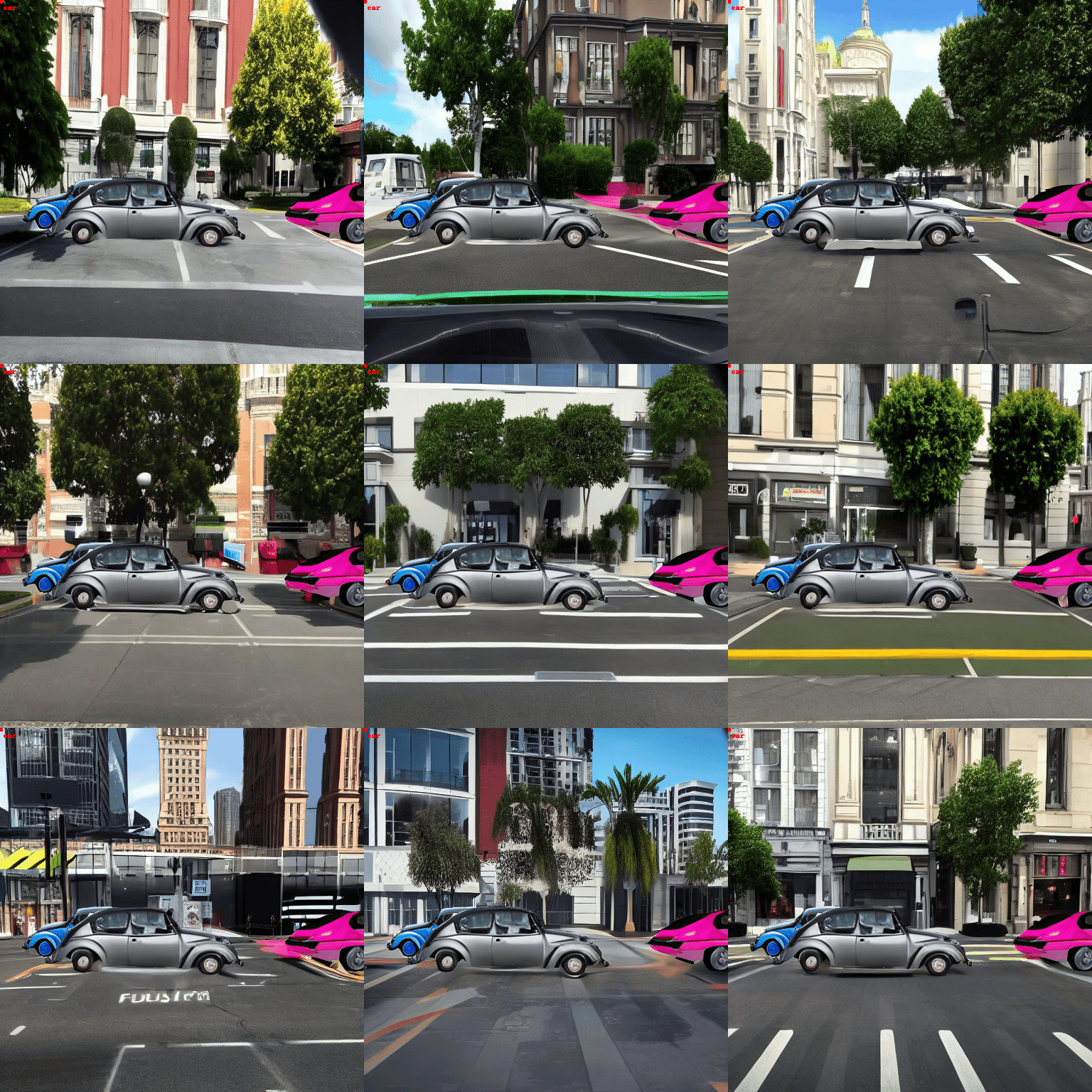}
\caption{\textbf{Selected scene containing three cars generated with BEV2EGO.}}
\label{fig:selected_scenes_3_cars_3}
\end{figure}

\noindent \textbf{Scenes with three cars for more object detectors.} Next, we show similar results for the scenes with more complex scenes containing three cars in Table~\ref{tab:reproduced_AP_complete_3cars} with the examples of scenes provided in \Cref{fig:selected_scenes_3_cars_0,fig:selected_scenes_3_cars_1,fig:selected_scenes_3_cars_2,fig:selected_scenes_3_cars_3}. Here, the setting for the scene generation is similar to that described in Section~\ref{sec:systematic_errors}, however, we increase the size of all the cars, randomly generate three cars and constrain two of them to be on the same lane following each other and the third - to drive in the opposite direction on another lane. This displays that BEV2EGO is a flexible method that allows to construct 2D scenes of different complexity. We observe that the results in Table~\ref{tab:reproduced_AP_complete_3cars} have average (across rows \textbf{MMS @[IoU=0.50:0.95]} and
\textbf{MMS @[IoU=0.50]}) Spearman's rank correlation coefficient of $0.78$, which indicates that findings on the scenes with two cars translate well to those on the scenes with three cars.

\begin{table}[t]
\centering
\scriptsize 
\begin{tabular}{c | c c c c c}
\hline
Metric & \textbf{FasterRCNN2} & YOLOv5n & YOLOv8n & YOLOv5x6 & RT-DETR-l \\ [0.5ex]
\hline\hline
\textbf{MMS @[IoU=0.50:0.95]} $\downarrow$ & \textbf{43.0} & 69.1 & 58.5 & 51.4 & 45.2 \\
\textbf{MMS @[IoU=0.50]} $\downarrow$ & 16.1 & 53.1 & 41.7 & 31.7 & 23.4 \\
\hline
\end{tabular}
\\ [1ex]
\begin{tabular}{c | c c c c c}
\hline
$\quad$ \textbf{MMS @[IoU=0.50]} $\downarrow \quad$ & \textbf{FasterRCNN2} & YOLOv5n & YOLOv8n & YOLOv5x6 & RT-DETR-l \\ [0.5ex]
\hline\hline
\textbf{Coupe Car} & \textbf{13.1} & 45.5 & 32.8 & 24.9 & 22.7 \\
\textbf{Sedan} & \textbf{17.8} & 57.0 & 46.9 & 33.1 & 25.4 \\
\textbf{SUV} & 17.0 & 52.1 & 41.1 & 33.9 & 23.5 \\
\textbf{Smart Car} & 13.3 & 49.8 & 40.2 & 32.7 & 20.7 \\
\textbf{Sports Car} & 19.2 & 60.9 & 47.4 & 33.4 & 25.2 \\
\hline
\end{tabular}
\\ [2ex]
\begin{tabular}{c | c c c c c}
\hline
Metric & YOLOv5x & YOLOv8x & RT-DETR-x & OneFormer & Mask2Former \\ [0.5ex]
\hline\hline
\textbf{MMS @[IoU=0.50:0.95]} $\downarrow$ & 51.7 & 47.1 & 44.7 & 46.7 & 47.5 \\
\textbf{MMS @[IoU=0.50]} $\downarrow$ & 32.2 & 28.4 & 23.4 & \textbf{14.4} & 16.6 \\
\hline
\end{tabular}
   \\ [1ex] 
    \centering
\begin{tabular}{c | c c c c c}
\hline
$\quad$ \textbf{MMS @[IoU=0.50]} $\downarrow \quad$ & YOLOv5x & YOLOv8x & RT-DETR-x & OneFormer & Mask2Former \\ [0.5ex]
\hline\hline
\textbf{Coupe Car} & 24.9 & 22.5 & 22.5 & 13.5 & 14.8 \\
\textbf{Sedan} & 34.3 & 29.9 & 25.2 & 18.2 & 20.2 \\
\textbf{SUV} & 32.7 & 30.0 & 24.7 & \textbf{14.5} & 18.3 \\
\textbf{Smart Car} & 33.1 & 28.6 & 20.1 & \textbf{10.4} & 13.2 \\
\textbf{Sports Car} & 35.7 & 30.7 & 24.8 & \textbf{16.0} & 16.8 \\
\hline
\end{tabular}
\caption{\textbf{Performance of all the investigated SOTA object detectors for more complex scenes with three cars.}  
\textit{In the first and second rows, first and third tables:} for these models, results on the discovered systematic errors show that \textbf{YOLOv5x6} model despite being SOTA in standard metrics, is worse in Mean Median Score (MMS) compared to \textbf{FasterRCNN2} and \textbf{RT-DETR} models. The order of the performance of the models has Spearman's rank correlation coefficient of $0.78$ with the order of the same methods evaluated on the scenes with two cars, which indicates that findings on the scenes with two cars translate well to those on the scenes with three cars.
\textit{Second and fourth tables:} A more fine-grained analysis of MMS for different car types.}
\label{tab:reproduced_AP_complete_3cars}
\end{table}

\begin{table}[b]
\begin{flushleft}
\subsection{Influence of color per car type}\label{app:systematic_errors_extended_quant1}
First, we examine how color influences the MMS depending on the car type. This aspect is important, as color is a common attribute in real-life scenarios of the object ``car''. In the table, we observe that no single dominant best and worst-performing colors exist across different car types, necessitating a more thorough evaluation. Here, we shorten the names of the models, so that the table would fit the page.
\end{flushleft}
\centering
\scriptsize 
\setlength{\tabcolsep}{2pt}
\renewcommand{\arraystretch}{0.9}
\begin{tabular}{c | c c c c c c c c c c | c}
\hline MMS by color and car type & FR2 & Y5n & Y5x & Y5x6 & Y8n & Y8x & RD-l & RD-x & OF & M2F & Average \\ 
\hline white coupe car & 8.1 & \textcolor{red}{52.6} & 29.6 & 27.6 & 40.5 & 29.8 & 14.8 & 14.9 & 6.8 & 19.7 & 24.4  \\ black coupe car & 6.5 & 47.1 & 33.5 & \textcolor{red}{28.8} & 38.6 & \textcolor{red}{35.0} & 15.8 & 15.8 & 7.7 & 13.5 & 24.2  \\ grey coupe car & 5.7 & 42.4 & 25.5 & 21.9 & 38.2 & 22.6 & 17.6 & 17.8 & 7.3 & \textcolor{green}{7.0} & 20.6  \\ yellow coupe car & 9.1 & 52.0 & 25.0 & 23.3 & 42.7 & 26.3 & 15.9 & 14.7 & \textcolor{green}{3.9} & 9.9 & 22.3  \\ red coupe car & 5.4 & 38.0 & 27.5 & \textcolor{green}{20.9} & \textcolor{green}{28.3} & 27.4 & \textcolor{green}{12.1} & \textcolor{green}{13.0} & 4.5 & 14.3 & 19.1  \\ blue coupe car & 6.1 & 38.9 & 28.4 & 21.7 & 33.9 & 28.2 & 16.7 & 17.2 & 9.3 & \textcolor{red}{28.7} & 22.9  \\ green coupe car & 8.8 & 42.6 & 28.2 & 25.8 & 39.0 & 31.4 & \textcolor{red}{17.9} & 18.1 & 11.8 & 17.3 & 24.1  \\ \textbf{brown coupe car} & \textcolor{red}{10.9} & 48.1 & \textcolor{red}{36.7} & 28.6 & \textcolor{red}{44.7} & 29.6 & 17.3 & \textcolor{red}{19.9} & \textcolor{red}{16.6} & 17.5 & \textbf{27.0}  \\ pink coupe car & 8.2 & \textcolor{green}{33.3} & \textcolor{green}{22.4} & 26.3 & 34.3 & 23.9 & 14.5 & 16.1 & 6.7 & 10.2 & 19.6  \\ orange coupe car & \textcolor{green}{4.8} & 42.8 & 30.8 & 26.2 & 38.2 & 24.0 & 14.3 & 14.8 & 9.0 & 14.1 & 21.9  \\ purple coupe car & 6.1 & 36.7 & 22.5 & 23.6 & 34.3 & \textcolor{green}{22.1} & 15.7 & 16.0 & 6.9 & 11.5 & 19.5  \\ \hline
\hline \textbf{white sedan} & \textcolor{red}{28.2} & \textcolor{red}{64.9} & 38.1 & \textcolor{red}{43.0} & 53.6 & \textcolor{red}{42.3} & \textcolor{red}{33.2} & \textcolor{red}{33.0} & \textcolor{red}{22.4} & 22.7 & \textbf{38.1}  \\ black sedan & 14.4 & 56.1 & 35.4 & 29.4 & 49.6 & 31.4 & 21.4 & 21.1 & 10.5 & 14.3 & 28.4  \\ grey sedan & \textcolor{green}{5.0} & 50.7 & 21.5 & 19.2 & 42.6 & \textcolor{green}{19.4} & 14.3 & 14.0 & \textcolor{green}{4.4} & 14.1 & 20.5  \\ yellow sedan & 8.4 & 61.6 & 34.5 & 28.4 & 49.0 & 32.9 & 15.7 & 15.1 & 5.5 & 14.1 & 26.5  \\ red sedan & 11.1 & \textcolor{green}{37.7} & 31.6 & 28.6 & \textcolor{green}{33.2} & 31.4 & 18.7 & 17.6 & 11.2 & \textcolor{red}{26.6} & 24.8  \\ blue sedan & 6.1 & 41.6 & \textcolor{green}{20.0} & \textcolor{green}{16.9} & 35.4 & 20.2 & \textcolor{green}{13.7} & \textcolor{green}{11.1} & 6.2 & \textcolor{green}{8.7} & 18.0  \\ green sedan & 11.0 & 63.1 & 36.5 & 30.7 & 54.0 & 35.0 & 20.1 & 17.5 & 10.1 & 17.9 & 29.6  \\ brown sedan & 9.6 & 51.7 & 27.8 & 27.6 & 44.8 & 27.9 & 18.2 & 19.0 & 11.0 & 9.9 & 24.8  \\ pink sedan & 13.5 & 42.0 & 28.1 & 26.4 & 37.9 & 32.2 & 18.3 & 17.6 & 14.8 & 8.8 & 23.9  \\ orange sedan & 12.1 & 55.9 & \textcolor{red}{43.0} & 41.2 & \textcolor{red}{55.0} & 40.1 & 17.2 & 18.6 & 9.1 & 19.5 & 31.2  \\ purple sedan & 9.4 & 44.3 & 25.6 & 24.8 & 37.9 & 25.5 & 16.1 & 16.2 & 14.4 & 20.7 & 23.5  \\ \hline
\hline white sports car & 16.5 & 63.5 & 38.2 & 34.9 & 52.6 & 38.2 & 18.3 & 19.7 & \textcolor{green}{3.8} & 19.4 & 30.5  \\ \textbf{black sports car} & \textcolor{red}{26.0} & \textcolor{red}{70.3} & 47.4 & 44.4 & \textcolor{red}{65.0} & \textcolor{red}{47.6} & \textcolor{red}{29.1} & \textcolor{red}{28.4} & 12.5 & \textcolor{red}{27.2} & \textbf{39.8}  \\ grey sports car & \textcolor{green}{7.6} & 59.1 & \textcolor{green}{33.2} & \textcolor{green}{26.5} & 45.3 & 36.0 & 21.4 & 20.0 & 8.1 & 19.3 & 27.6  \\ yellow sports car & 16.9 & 68.6 & 35.9 & 33.3 & 48.9 & 32.8 & 16.2 & 19.9 & 13.1 & 21.0 & 30.7  \\ red sports car & 11.2 & \textcolor{green}{43.9} & 37.0 & 39.7 & \textcolor{green}{38.0} & \textcolor{green}{32.1} & 18.5 & 18.6 & 9.5 & 12.7 & 26.1  \\ blue sports car & 8.5 & 52.1 & 35.1 & 26.8 & 43.1 & 37.1 & 16.9 & \textcolor{green}{15.7} & 9.2 & 16.4 & 26.1  \\ green sports car & 8.6 & 52.6 & 36.4 & 33.1 & 49.2 & 37.4 & \textcolor{green}{15.6} & 17.9 & 15.5 & 16.1 & 28.2  \\ brown sports car & 13.9 & 64.6 & 40.1 & 38.4 & 51.1 & 39.8 & 19.7 & 19.9 & \textcolor{red}{17.6} & 18.2 & 32.3  \\ pink sports car & 16.4 & 46.9 & 34.6 & 38.5 & 43.2 & 41.9 & 16.4 & 17.5 & 7.7 & \textcolor{green}{8.3} & 27.1  \\ orange sports car & 14.9 & 52.3 & \textcolor{red}{48.0} & \textcolor{red}{50.1} & 49.3 & 41.5 & 18.9 & 20.0 & 6.2 & 12.6 & 31.4  \\ purple sports car & 20.4 & 55.0 & 38.4 & 33.1 & 44.3 & 37.8 & 22.3 & 22.4 & 13.3 & 23.6 & 31.1  \\ \hline
\hline white smart car & 13.7 & \textcolor{red}{64.0} & 40.0 & 41.0 & 52.4 & 39.0 & 21.4 & 19.2 & 10.0 & 26.3 & 32.7  \\ black smart car & 11.6 & 51.0 & 32.7 & 30.6 & 47.0 & 36.9 & 21.2 & \textcolor{red}{24.0} & 9.4 & 17.9 & 28.2  \\ grey smart car & \textcolor{green}{7.4} & 45.4 & \textcolor{green}{23.7} & \textcolor{green}{24.0} & \textcolor{green}{38.5} & \textcolor{green}{26.9} & 15.5 & 16.3 & 5.0 & 15.3 & 21.8  \\ yellow smart car & 12.0 & 56.6 & 36.8 & 31.7 & 45.6 & 30.0 & 17.9 & 21.1 & 8.7 & 24.1 & 28.4  \\ red smart car & 10.2 & \textcolor{green}{39.0} & 36.4 & 38.3 & 40.5 & 37.5 & 18.4 & 17.9 & 7.5 & \textcolor{green}{10.1} & 25.6  \\ blue smart car & 8.1 & 50.6 & 32.7 & 31.6 & 46.1 & 33.7 & \textcolor{green}{14.5} & \textcolor{green}{14.1} & \textcolor{green}{4.3} & 10.9 & 24.7  \\ green smart car & 13.7 & 53.7 & 39.7 & 39.1 & 52.8 & 38.9 & 15.2 & 18.6 & 7.5 & \textcolor{red}{30.4} & 30.9  \\ brown smart car & \textcolor{red}{13.8} & 48.8 & 38.0 & 37.4 & 50.5 & 37.2 & 21.2 & 22.8 & 12.4 & 16.6 & 29.9  \\ \textbf{pink smart car} & 13.7 & 57.5 & \textcolor{red}{47.0} & \textcolor{red}{48.1} & \textcolor{red}{57.3} & \textcolor{red}{47.2} & 19.4 & 17.8 & 13.4 & 16.4 & \textbf{33.8}  \\ orange smart car & 11.0 & 46.7 & 35.9 & 37.7 & 48.0 & 36.1 & 17.0 & 16.5 & 7.2 & 18.8 & 27.5  \\ purple smart car & 13.2 & 55.4 & 36.5 & 31.9 & 52.6 & 39.8 & \textcolor{red}{22.4} & 23.8 & \textcolor{red}{13.6} & 22.1 & 31.1  \\ \hline
\hline white SUV & 14.7 & 44.1 & 34.4 & 35.2 & 34.6 & 34.4 & \textcolor{red}{19.9} & \textcolor{red}{22.0} & \textcolor{red}{10.9} & 6.5 & 25.7  \\ black SUV & 8.5 & 41.7 & 30.2 & 28.5 & 37.2 & 26.8 & 17.1 & 14.9 & 4.4 & \textcolor{red}{8.6} & 21.8  \\ grey SUV & 9.8 & 38.2 & 28.6 & 28.3 & 37.9 & 26.9 & 16.2 & 15.8 & 5.3 & 8.3 & 21.5  \\ yellow SUV & 14.4 & \textcolor{red}{55.6} & 37.2 & 36.4 & 35.3 & 30.8 & 14.6 & 16.1 & 2.1 & 7.3 & 25.0  \\ red SUV & 10.5 & 34.8 & 30.9 & 38.3 & 37.4 & \textcolor{red}{36.3} & 16.4 & 14.0 & \textcolor{green}{0.2} & 2.2 & 22.1  \\ blue SUV & 6.0 & 42.6 & 27.2 & \textcolor{green}{20.6} & 30.4 & \textcolor{green}{22.9} & 12.3 & \textcolor{green}{12.0} & 2.5 & 4.8 & 18.1  \\ green SUV & 7.3 & 43.5 & 24.5 & 29.1 & 34.7 & 25.7 & 13.1 & 15.1 & 3.2 & 4.6 & 20.1  \\ brown SUV & \textcolor{green}{5.3} & 45.0 & 32.6 & 28.3 & 44.6 & 31.8 & 15.0 & 15.8 & 3.7 & 3.3 & 22.5  \\ pink SUV & 10.9 & \textcolor{green}{32.7} & 29.9 & 29.2 & \textcolor{green}{26.2} & 29.0 & \textcolor{green}{11.8} & 12.0 & 2.2 & \textcolor{green}{0.0} & 18.4  \\ \textbf{orange SUV} & \textcolor{red}{21.2} & 54.9 & \textcolor{red}{44.0} & \textcolor{red}{40.8} & \textcolor{red}{47.3} & 34.2 & 18.2 & 18.8 & 3.4 & 5.9 & \textbf{28.9}  \\ purple SUV & 6.8 & 41.2 & \textcolor{green}{23.4} & 30.2 & 38.3 & 28.7 & 14.7 & 14.8 & 7.5 & 4.9 & 21.0  \\ \hline
\end{tabular}
\newline
 \caption{\textbf{Influence of color and car type on MMS@[IoU=0.50] $\downarrow$.} This table highlights in red the combination of car type and color with the highest MMS, and in green, the combination with the lowest MMS for each object detector. The combination that, on average across detectors, has the highest MMS is marked in bold}
\label{tab:MMS_influence_color}
\end{table}

\newpage

\begin{table}[t]
\begin{flushleft}
\subsection{Influence of background per car type}\label{app:systematic_errors_extended_quant2}
Next, we analyze how background type influences the MMS depending on the car type. This analysis is crucial, as the object ``car'' is placed against various backgrounds in real-life scenarios. In the table, we observe that no single dominant best and worst-performing background types exist across different car types, which requires a more thorough evaluation.  Here, we shorten the names of the models, so that the table would fit the page. 
\end{flushleft}
\centering
\scriptsize 
\setlength{\tabcolsep}{2pt}
\renewcommand{\arraystretch}{0.9}
\begin{tabular}{c | c c c c c c c c c c | c}
\hline MMS by color and car type & FR2 & Y5n & Y5x & Y5x6 & Y8n & Y8x & RD-l & RD-x & OF & M2F & Average \\ 
\hline coupe car in forest & 6.5 & 47.1 & 27.4 & 24.9 & 39.0 & 25.4 & 15.6 & 15.7 & 6.6 & 14.1 & 22.2  \\ coupe car on beach & 6.2 & 44.8 & 29.2 & 25.8 & 38.2 & 25.6 & 14.8 & 15.5 & \textcolor{green}{5.8} & 15.5 & 22.1  \\ coupe car in city & 10.6 & 41.1 & 29.6 & 26.0 & 36.0 & 29.4 & 18.4 & 19.6 & \textcolor{red}{14.6} & \textcolor{red}{23.1} & 24.8  \\ coupe car on snowy street & \textcolor{green}{4.5} & \textcolor{green}{35.7} & \textcolor{green}{20.4} & \textcolor{green}{21.2} & \textcolor{green}{33.2} & \textcolor{green}{21.3} & \textcolor{green}{13.1} & \textcolor{green}{13.7} & 6.2 & \textcolor{green}{10.2} & 17.9  \\ \textbf{coupe car on highway} & \textcolor{red}{11.1} & \textcolor{red}{50.3} & \textcolor{red}{35.7} & \textcolor{red}{30.9} & \textcolor{red}{45.3} & \textcolor{red}{36.1} & \textcolor{red}{19.7} & \textcolor{red}{20.2} & 12.9 & 16.0 & \textbf{27.8}  \\ coupe car near lake & 5.8 & 39.2 & 26.7 & 21.8 & 35.0 & 25.4 & 14.7 & 14.8 & 5.9 & 10.8 & 20.0  \\ \hline
\hline sedan in forest & 9.0 & 52.1 & 32.9 & 27.6 & 46.6 & 30.9 & 16.7 & 17.1 & 11.0 & 16.2 & 26.0  \\ \textbf{sedan on beach} & \textcolor{red}{18.1} & \textcolor{red}{54.9} & \textcolor{red}{38.9} & \textcolor{red}{35.0} & \textcolor{red}{48.1} & \textcolor{red}{36.0} & \textcolor{red}{22.4} & \textcolor{red}{21.8} & \textcolor{red}{16.8} & 16.2 & \textbf{30.8}  \\ sedan in city & 8.5 & 48.5 & 24.9 & 25.6 & \textcolor{green}{38.4} & 25.3 & 16.5 & 15.9 & 10.3 & \textcolor{red}{23.2} & 23.7  \\ sedan on snowy street & 12.3 & 53.5 & 31.2 & 28.5 & 45.6 & 31.6 & 18.6 & 18.7 & 11.2 & 22.3 & 27.4  \\ sedan on highway & 12.9 & 53.9 & 32.2 & 31.0 & 46.2 & 33.2 & 20.4 & 18.8 & 8.4 & \textcolor{green}{9.1} & 26.6  \\ sedan near lake & \textcolor{green}{7.6} & \textcolor{green}{45.3} & \textcolor{green}{24.8} & \textcolor{green}{23.1} & 42.7 & \textcolor{green}{24.6} & \textcolor{green}{16.1} & \textcolor{green}{15.9} & \textcolor{green}{6.7} & 9.6 & 21.7  \\ \hline
\hline sports car in forest & 15.3 & 55.7 & 41.1 & 34.5 & 48.0 & 39.4 & \textcolor{red}{21.2} & 22.8 & 10.5 & \textcolor{green}{14.4} & 30.3  \\ sports car on beach & \textcolor{red}{18.8} & \textcolor{red}{59.7} & 40.0 & 36.4 & 48.4 & \textcolor{red}{41.1} & 20.6 & \textcolor{red}{23.2} & \textcolor{green}{7.4} & 16.8 & 31.2  \\ sports car in city & 13.1 & \textcolor{green}{51.5} & 35.4 & 36.6 & 45.9 & 39.0 & 19.9 & 19.2 & 11.5 & \textcolor{red}{25.6} & 29.8  \\ sports car on snowy street & 12.6 & 59.6 & \textcolor{green}{33.6} & \textcolor{green}{32.3} & \textcolor{green}{45.8} & \textcolor{green}{35.5} & \textcolor{green}{17.7} & 18.3 & \textcolor{red}{14.2} & 18.2 & 28.8  \\ \textbf{sports car on highway} & 16.3 & 58.3 & \textcolor{red}{42.5} & \textcolor{red}{43.3} & \textcolor{red}{51.5} & 37.9 & 20.3 & 18.7 & 12.3 & 17.4 & \textbf{31.9}  \\ sports car near lake & \textcolor{green}{11.8} & 56.4 & 38.3 & 34.5 & 49.7 & 37.9 & 17.8 & \textcolor{green}{18.1} & 8.1 & 16.8 & 28.9  \\ \hline
\hline smart car in forest & 10.0 & \textcolor{red}{54.8} & 34.8 & \textcolor{green}{33.7} & 50.9 & \textcolor{green}{32.1} & 17.6 & 20.2 & \textcolor{green}{6.8} & \textcolor{green}{15.0} & 27.6  \\ smart car on beach & 10.8 & \textcolor{green}{49.7} & 38.3 & 34.7 & \textcolor{green}{45.2} & \textcolor{red}{38.6} & 18.4 & 19.1 & 10.1 & 15.8 & 28.1  \\ smart car in city & 12.5 & 53.7 & 35.3 & 34.4 & 47.9 & 37.1 & \textcolor{green}{17.5} & \textcolor{green}{18.0} & 11.2 & \textcolor{red}{28.8} & 29.6  \\ smart car on snowy street & \textcolor{green}{9.6} & 50.7 & \textcolor{green}{31.9} & 35.8 & 47.9 & 35.4 & 19.8 & 18.8 & 6.8 & 15.4 & 27.2  \\ \textbf{smart car on highway} & \textcolor{red}{15.4} & 53.3 & 35.7 & 36.4 & \textcolor{red}{50.9} & 38.3 & \textcolor{red}{20.0} & \textcolor{red}{21.2} & \textcolor{red}{12.3} & 26.4 & \textbf{31.0}  \\ smart car near lake & 12.2 & 51.1 & \textcolor{red}{42.1} & \textcolor{red}{37.8} & 49.4 & 38.1 & 17.8 & 18.8 & 7.6 & 16.7 & 29.2  \\ \hline
\hline SUV in forest & 9.3 & 47.6 & 33.4 & 32.7 & 38.7 & 30.2 & 16.5 & 17.5 & 7.4 & \textcolor{red}{8.8} & 24.2  \\ SUV on beach & \textcolor{red}{13.7} & 46.0 & 34.6 & \textcolor{red}{38.1} & 39.8 & \textcolor{red}{35.6} & 17.0 & 16.4 & 4.3 & 7.1 & 25.3  \\ SUV in city & 9.2 & \textcolor{green}{39.0} & \textcolor{green}{26.9} & \textcolor{green}{26.8} & \textcolor{green}{31.6} & \textcolor{green}{24.8} & 14.1 & 14.3 & 2.6 & 5.9 & 19.5  \\ SUV on snowy street & 10.2 & 39.9 & 29.5 & 29.0 & 33.1 & 27.9 & 13.7 & 14.4 & 3.9 & 3.4 & 20.5  \\ \textbf{SUV on highway} & 13.6 & \textcolor{red}{49.8} & \textcolor{red}{36.8} & 33.3 & \textcolor{red}{40.1} & 34.4 & \textcolor{red}{18.3} & \textcolor{red}{18.1} & \textcolor{red}{7.5} & 7.2 & \textbf{25.9}  \\ SUV near lake & \textcolor{green}{8.4} & 39.4 & 27.9 & 30.6 & 35.7 & 27.3 & \textcolor{green}{13.6} & \textcolor{green}{13.7} & \textcolor{green}{0.1} & \textcolor{green}{0.1} & 19.7  \\ \hline
\end{tabular}
\newline
\caption{\textbf{Influence of background and car type on MMS@[IoU=0.50] $\downarrow$.} This table highlights in red the combination of car and background type with the highest MMS, and in green, the combination with the lowest MMS for each object detector. The combination that, on average across detectors, has the highest MMS is marked in bold}
\label{tab:MMS_influence_background}
\end{table}

\clearpage

\begin{figure}[ht]  %
\begin{flushleft}
\subsection{Influence of rotation angle per car type}\label{app:systematic_errors_extended_quant4}
In this section, we separately examine for each object detector the behavior of MMS per rotation angle. Similar to Fig.~\ref{fig:Sim2Real_quantitative}, we bin all possible rotation angles after taking the absolute value that a car was visible from in the interval $[0^{\circ}, 180^{\circ}]$ into nine bins and plot the values attained in the respective bin per endpoint of the respective bin. In the first figure below, we display the behavior on average across different car types, and then examine this behavior for the car type ``SUV'' in the second figure below, which exhibits the highest MMS across all but one object detector - \emph{YOLOv5x} - in the bin $[80^{\circ}, 100^{\circ}]$, centered around the angle $90^{\circ}$ in absolute value, thus a rear or front view. OneFormer as well as Mask2Former seem to be able to detect the car perfectly for this car type. 

\end{flushleft}

  \centering
  \begin{subfigure}[b]{\columnwidth}
    \includegraphics[width=\linewidth]{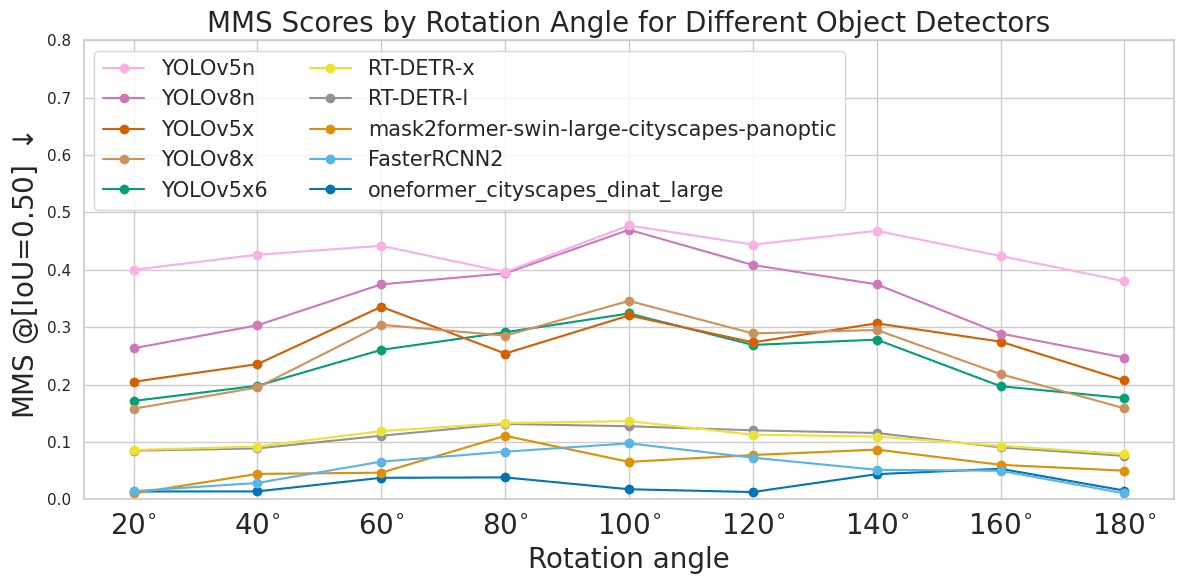}
    \label{fig:rotation_angle}
  \end{subfigure}
  \hfill  %
  \begin{subfigure}[b]{\columnwidth}
    \includegraphics[width=\linewidth]{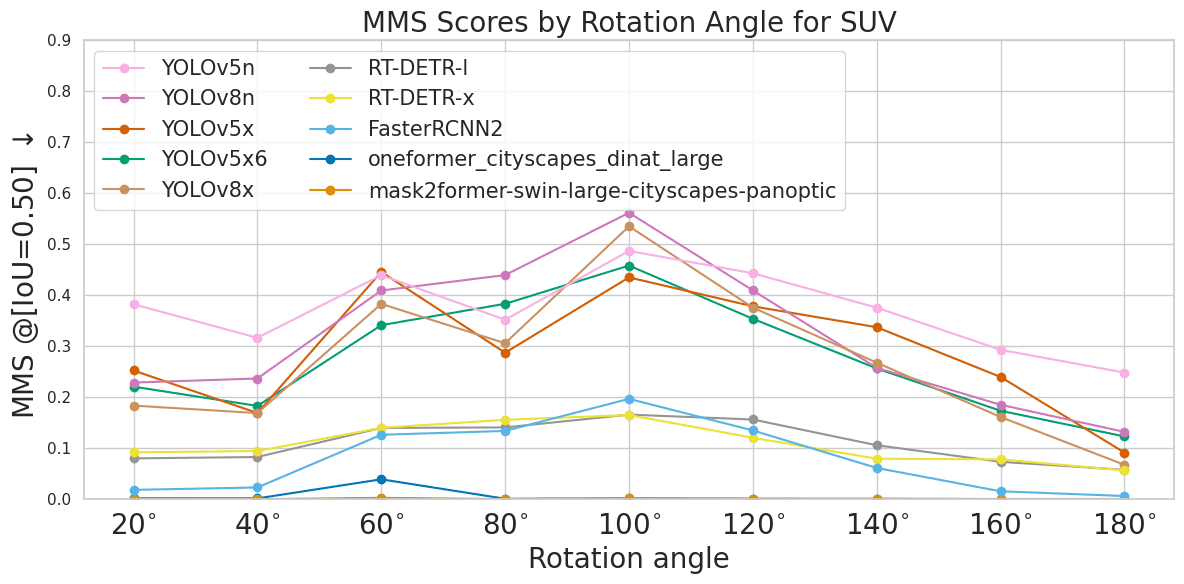}
    \label{fig:rotation_angle_SUV}
  \end{subfigure}
   \caption{\textbf{Comparative analysis of dependence of Mean Median Score (MMS) on rotation angles.}}
  \label{fig:comparative_rotation_angles}
\end{figure}

\clearpage

\begin{figure}[hbt!]
\begin{flushleft}
\section{More examples of the generated scenes with BEV2EGO}\label{app:systematic_errors_examples}
\subsection{Systematic errors per object detector}
In this section, in Fig.~\ref{fig:systematic_errors_detection_extended}, we display systematic errors for four additional object detectors used in the evaluation in Table~\ref{tab:reproduced_AP}, thus extending the examples shown in Fig.~\ref{fig:systematic_errors_detection} and Fig.~\ref{fig:systematic_errors_detection_VCEs}.
\end{flushleft}
    \centering
    \footnotesize
    \setlength{\tabcolsep}{2pt}
    \renewcommand{\arraystretch}{0.8}
    \begin{tabular}{cc|cc}
        \multicolumn{2}{c|}{Object Detector: \textbf{\textcolor{green}{YOLOv8x}}} & 
        \multicolumn{2}{c}{Object Detector: \textbf{\textcolor{orange}{RT-DETR-x}}} \\
        \hline
        \textbf{Backpack: 57\%} & \textbf{Backpack: 45\%} & 
        \textbf{Car: 93\%} & \textbf{Car: 93\%} \\
        
        \includegraphics[width=0.22\textwidth]{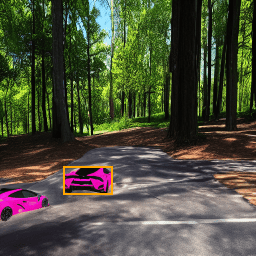} &
        \includegraphics[width=0.22\textwidth]{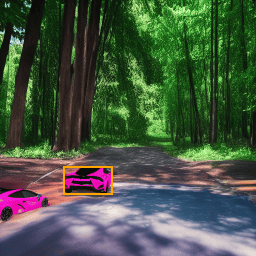} &
        \includegraphics[width=0.22\textwidth]{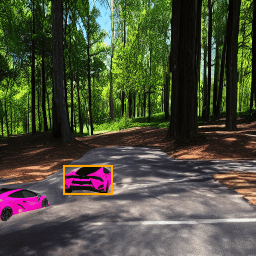} &
        \includegraphics[width=0.22\textwidth]{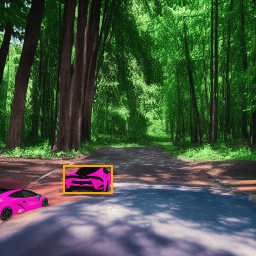} \\
    \hline
        \multicolumn{2}{c|}{Object Detector: \textbf{\textcolor{red}{YOLOv8n}}} & 
        \multicolumn{2}{c}{Object Detector: \textbf{\textcolor{cyan}{RT-DETR-l}}} \\
        \hline
        \textbf{Boat: 23\%} & \textbf{Train: 52\%} & 
        \textbf{Car: 60\%} & \textbf{Car: 83\%} \\
        
        \includegraphics[width=0.22\textwidth]{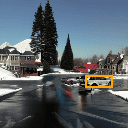} &
        \includegraphics[width=0.22\textwidth]{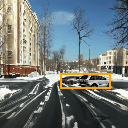} &
        \includegraphics[width=0.22\textwidth]{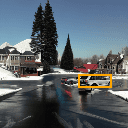} &
        \includegraphics[width=0.22\textwidth]{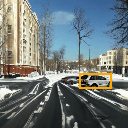} \\
    \hline
        \multicolumn{2}{c|}{Object Detector: \textbf{\textcolor{purple}{YOLOv5x}}} & 
        \multicolumn{2}{c}{Object Detector: \textbf{\textcolor{orange}{RT-DETR-x}}} \\
        \hline
        \textbf{Boat: 75\%} & \textbf{Boat: 48\%} & 
        \textbf{Car: 89\%} & \textbf{Car: 90\%} \\
        
        \includegraphics[width=0.22\textwidth]{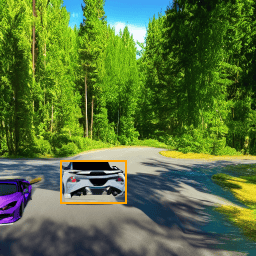} &
        \includegraphics[width=0.22\textwidth]{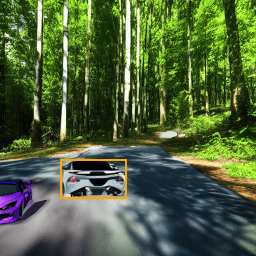} &
        \includegraphics[width=0.22\textwidth]{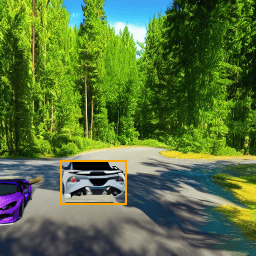} &
        \includegraphics[width=0.22\textwidth]{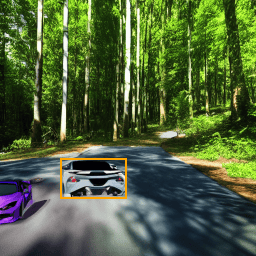} \\

        \multicolumn{2}{c|}{Object Detector: \textbf{\textcolor{orange}{RT-DETR-x}}} & 
        \multicolumn{2}{c}{Object Detector: \textbf{\textcolor{magenta}{FasterRCNN2}}} \\
        \hline
        \textbf{Cow: 43\%} & \textbf{Cow: 70\%} & 
        \textbf{Car: 97\%} & \textbf{Car: 91\%} \\
        
        \includegraphics[width=0.22\textwidth]{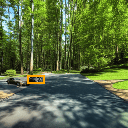} &
        \includegraphics[width=0.22\textwidth]{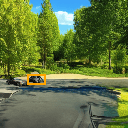} &
        \includegraphics[width=0.22\textwidth]{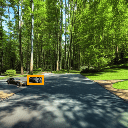} &
        \includegraphics[width=0.22\textwidth]{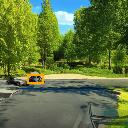} \\

    \end{tabular}
    \caption{\textbf{Influence of different of \emph{color} and \emph{resolution} values on the behaviour of different object detectors which results in the systematic errors, left row.} In each figure, we show the behaviour of object detectors for which this systematic error was problematic respectively in the left column and the behaviour of an object detector, which does not have this error - in the right.}
    \label{fig:systematic_errors_detection_extended}
\end{figure}

\begin{figure}[hbt!]
\begin{flushleft}
\subsection{Influence of small changes in attributes on object detectors}
In this section, in Fig.~\ref{fig:systematic_errors_detection_VCEs_extended}, we display how small changes in the attributes affect the behavior of the object detectors regarding the systematic errors from the previous section, thus extending the examples shown in Fig.~\ref{fig:systematic_errors_detection_VCEs}.

\end{flushleft}

    \centering
    \footnotesize
    \setlength{\tabcolsep}{2pt}
    \renewcommand{\arraystretch}{0.8}
    \begin{tabular}{cc|cc}
        \multicolumn{2}{c|}{Object Detector: \textbf{\textcolor{green}{YOLOv8x}}} & 
        \multicolumn{2}{c}{Object Detector: \textbf{\textcolor{green}{YOLOv8x}}} \\
        \hline
        \textbf{Backpack: 57\%} & \textbf{Backpack: 45\%} & 
        \textbf{Car: 82\%} & \textbf{Car: 61\%} \\
        
        \includegraphics[width=0.22\textwidth]{images/Appendix/Systematic_errors/YOLOv8x/app_YOLOv8x_SE1_1.png} &
        \includegraphics[width=0.22\textwidth]{images/Appendix/Systematic_errors/YOLOv8x/app_YOLOv8x_SE1_2.png} &
        \includegraphics[width=0.22\textwidth]{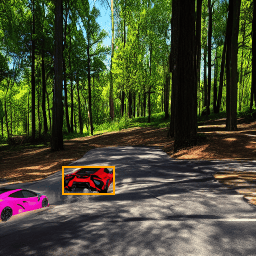} &
        \includegraphics[width=0.22\textwidth]{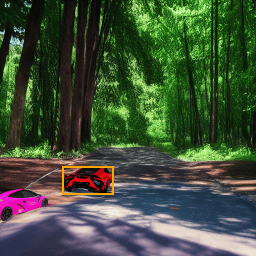} \\
    \hline
        \multicolumn{2}{c|}{Object Detector: \textbf{\textcolor{red}{YOLOv8n}}} & 
        \multicolumn{2}{c}{Object Detector: \textbf{\textcolor{red}{YOLOv8n}}} \\
        \hline
        \textbf{Boat: 23\%} & \textbf{Train: 52\%} & 
        \textbf{Car: 65\%} & \textbf{Car: 71\%} \\
        
        \includegraphics[width=0.22\textwidth]{images/Appendix/Systematic_errors/YOLOv8n/app_YOLOv8n_SE1_1.png} &
        \includegraphics[width=0.22\textwidth]{images/Appendix/Systematic_errors/YOLOv8n/app_YOLOv8n_SE1_2.png} &
        \includegraphics[width=0.22\textwidth]{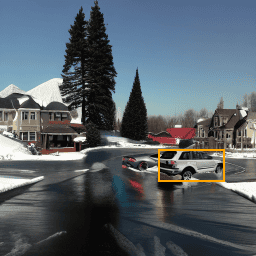} &
        \includegraphics[width=0.22\textwidth]{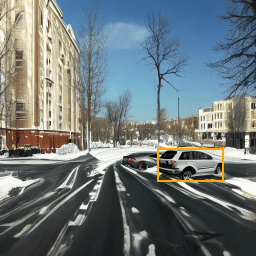} \\
    \hline
        \multicolumn{2}{c|}{Object Detector: \textbf{\textcolor{purple}{YOLOv5x}}} & 
        \multicolumn{2}{c}{Object Detector: \textbf{\textcolor{purple}{YOLOv5x}}} \\
        \hline
        \textbf{Boat: 75\%} & \textbf{Boat: 48\%} & 
        \textbf{Car: 80\%} & \textbf{Car: 86\%} \\
        
        \includegraphics[width=0.22\textwidth]{images/Appendix/Systematic_errors/YOLOv5x/app_YOLOv5x_SE1_1.png} &
        \includegraphics[width=0.22\textwidth]{images/Appendix/Systematic_errors/YOLOv5x/app_YOLOv5x_SE1_2.png} &
        \includegraphics[width=0.22\textwidth]{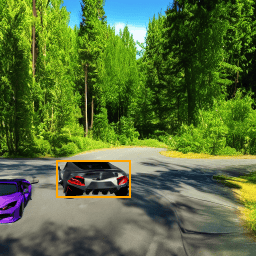} &
        \includegraphics[width=0.22\textwidth]{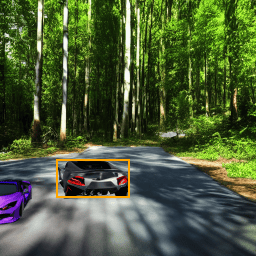} \\

        \multicolumn{2}{c|}{Object Detector: \textbf{\textcolor{blue}{RT-DETR-x}}} & 
        \multicolumn{2}{c}{Object Detector: \textbf{\textcolor{magenta}{RT-DETR-x}}} \\
        \hline
        \textbf{Cow: 43\%} & \textbf{Cow: 70\%} & 
        \textbf{Car: 83\%} & \textbf{Car: 57\%} \\
        
        \includegraphics[width=0.22\textwidth]{images/Appendix/Systematic_errors/RTDETRx/app_RTDETRx_SE1_1.png} &
        \includegraphics[width=0.22\textwidth]{images/Appendix/Systematic_errors/RTDETRx/app_RTDETRx_SE1_2.png} &
        \includegraphics[width=0.22\textwidth]{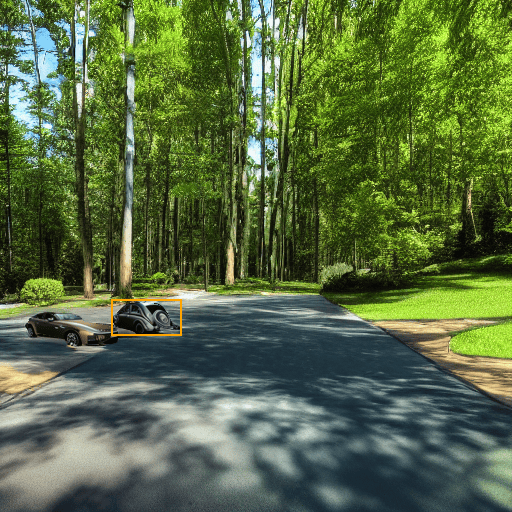} &
        \includegraphics[width=0.22\textwidth]{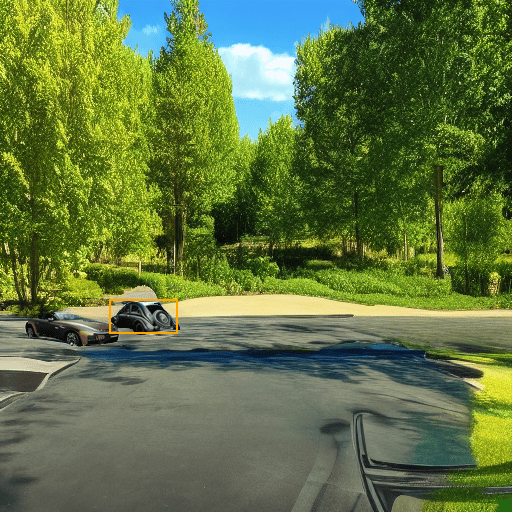} \\

    \end{tabular}
    \caption{\textbf{Influence of different of \emph{color} and \emph{resolution} on the behaviour of different object detectors which results in the systematic errors, left row.} In each figure, we show the behaviour of object detectors for which this systematic error was problematic respectively in the left column and the behaviour of the same object detector after the change of one of the attributes - in the right. The attribute changes are respectively in each row as follows: i) color: pink $\rightarrow$ red, ii) resolution $128\times128 \rightarrow 256\times256$, iii) color: white $\rightarrow$ grey, iv) resolution $128\times128 \rightarrow 512\times512$. In all cases, the objects are correctly detected after the respective change in only one of the attributes.}
    \label{fig:systematic_errors_detection_VCEs_extended}
\end{figure}

\begin{figure}[htb!]
\begin{flushleft}
\subsection{Randomly selected scenes}
In this section, in \Cref{fig:random_scenes_1,fig:random_scenes_2,fig:random_scenes_3,fig:random_scenes_4,fig:random_scenes_5,fig:random_scenes_6}, we display randomly selected scenes generated with BEV2EGO for all 9 seeds.
\end{flushleft}
\centering
\includegraphics[width=\linewidth]{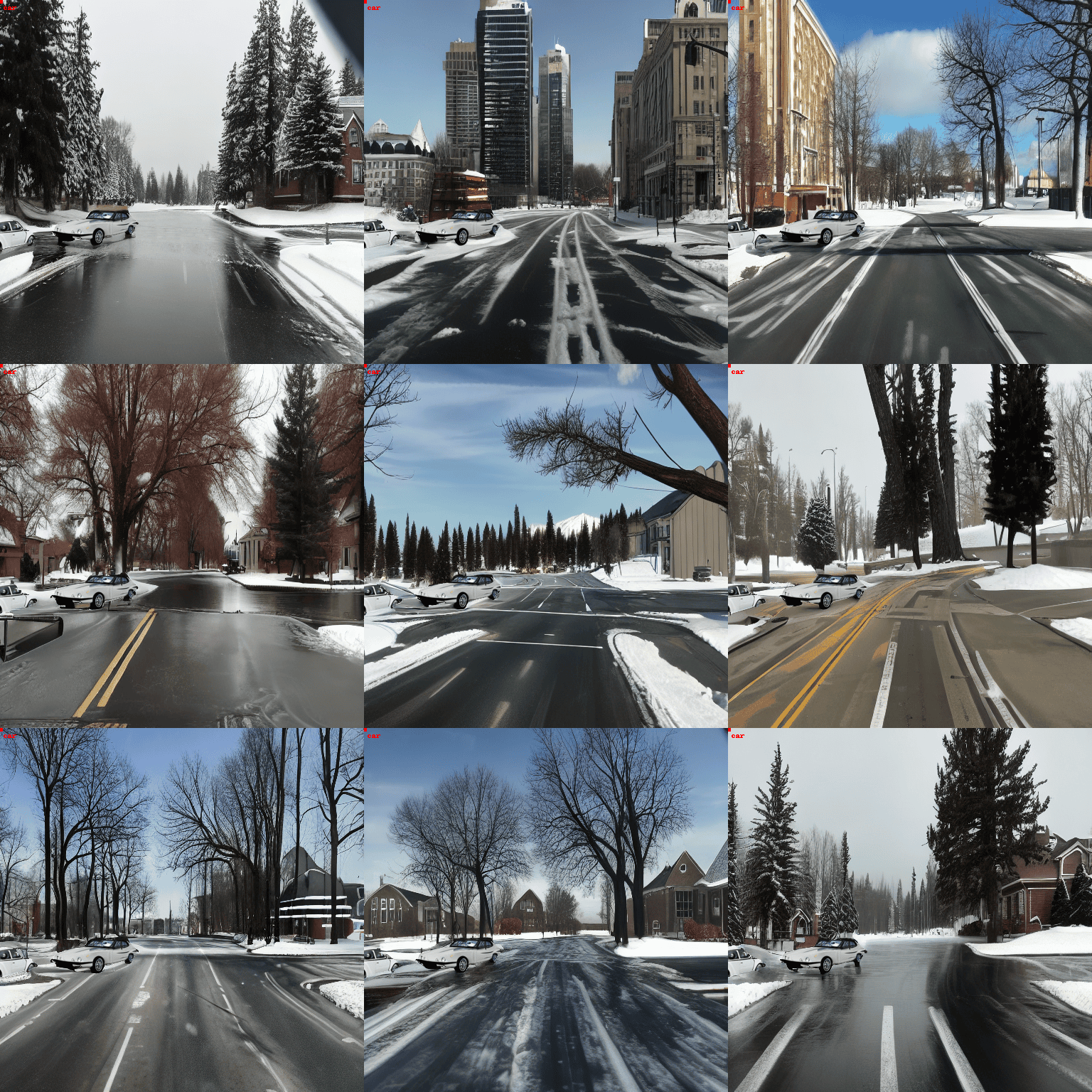}
\caption{\textbf{Randomly selected scene generated with BEV2EGO.}}
\label{fig:random_scenes_1}
\end{figure}

\begin{figure}[htb!]
\centering
\includegraphics[width=\linewidth]{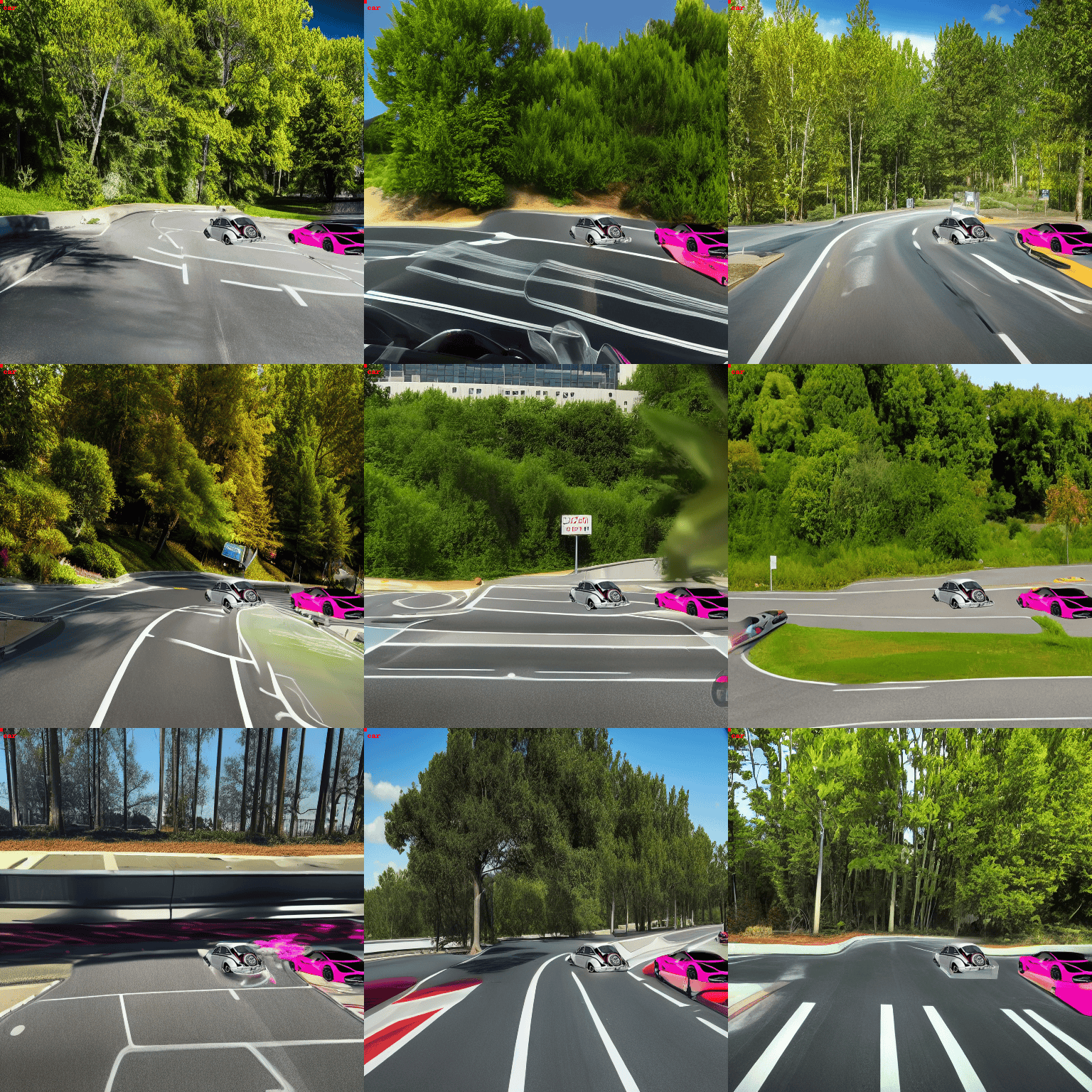}
\caption{\textbf{Randomly selected scene generated with BEV2EGO.}}
\label{fig:random_scenes_2}
\end{figure}

\begin{figure}[htb!]
\centering
\includegraphics[width=\linewidth]{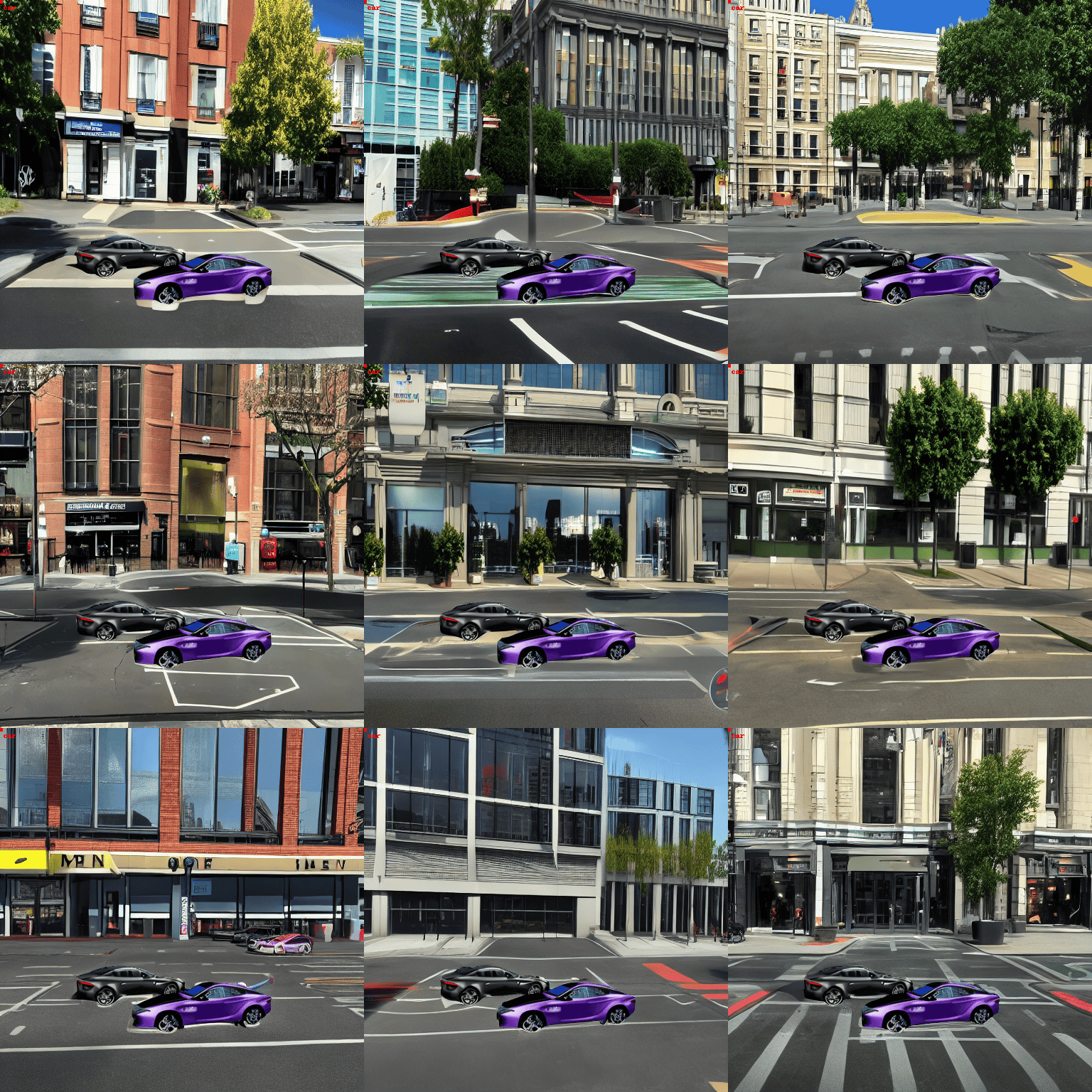}
\caption{\textbf{Randomly selected scene generated with BEV2EGO.}}
\label{fig:random_scenes_3}
\end{figure}

\begin{figure}[htb!]
\centering
\includegraphics[width=\linewidth]{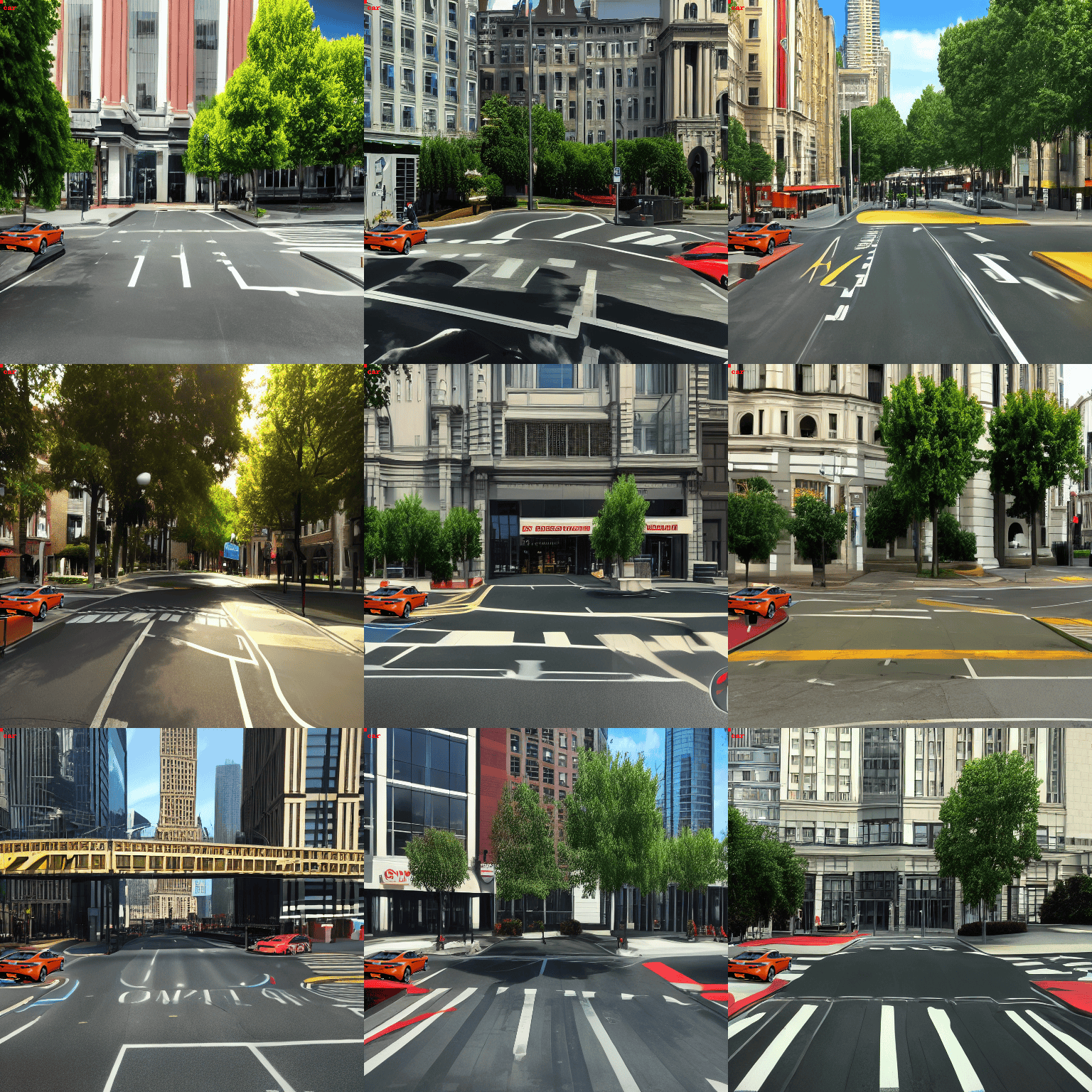}
\caption{\textbf{Randomly selected scene generated with BEV2EGO.}}
\label{fig:random_scenes_4}
\end{figure}

\begin{figure}[htb!]
\centering
\includegraphics[width=\linewidth]{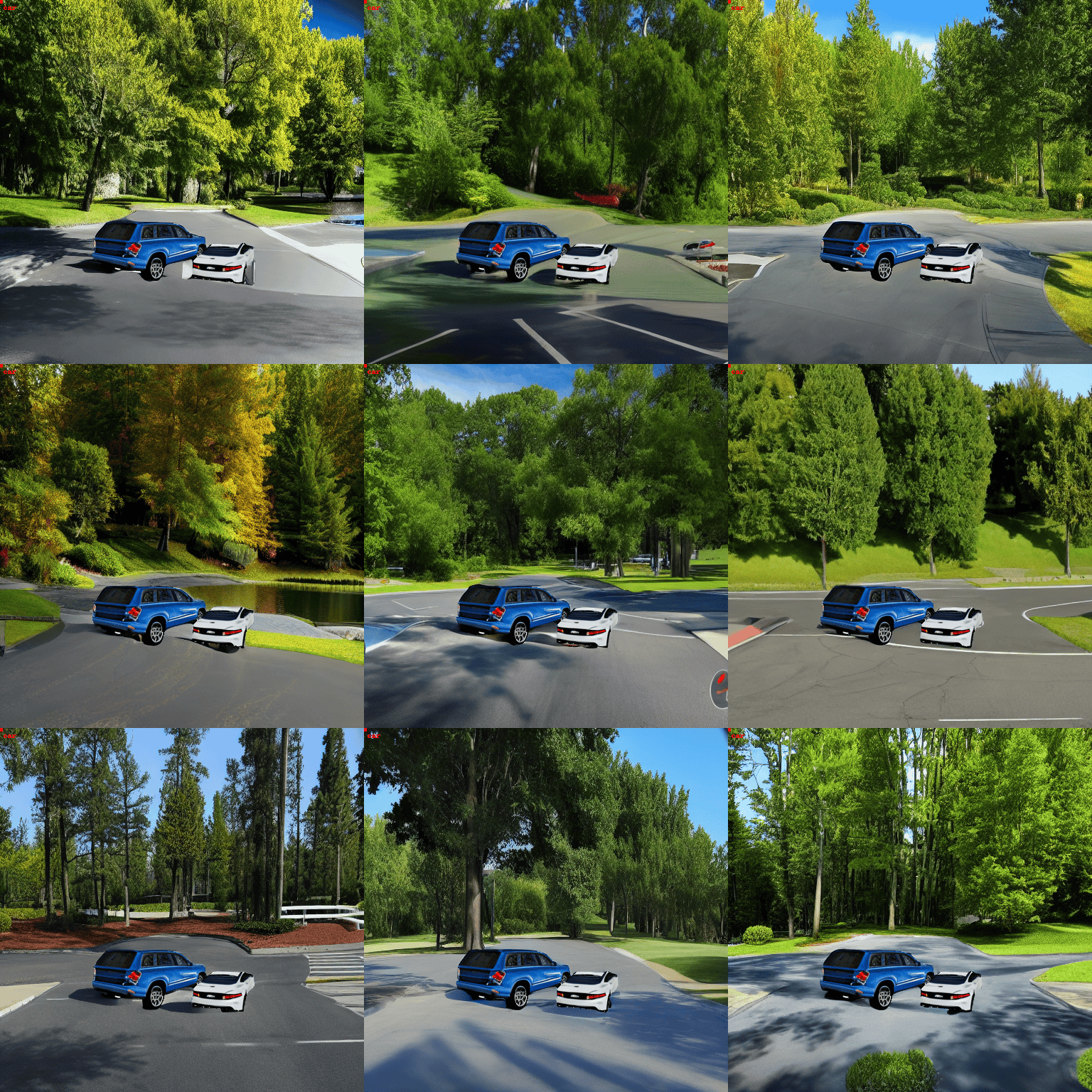}
\caption{\textbf{Randomly selected scene generated with BEV2EGO.}}
\label{fig:random_scenes_5}
\end{figure}

\begin{figure}[htb!]
\centering
\includegraphics[width=\linewidth]{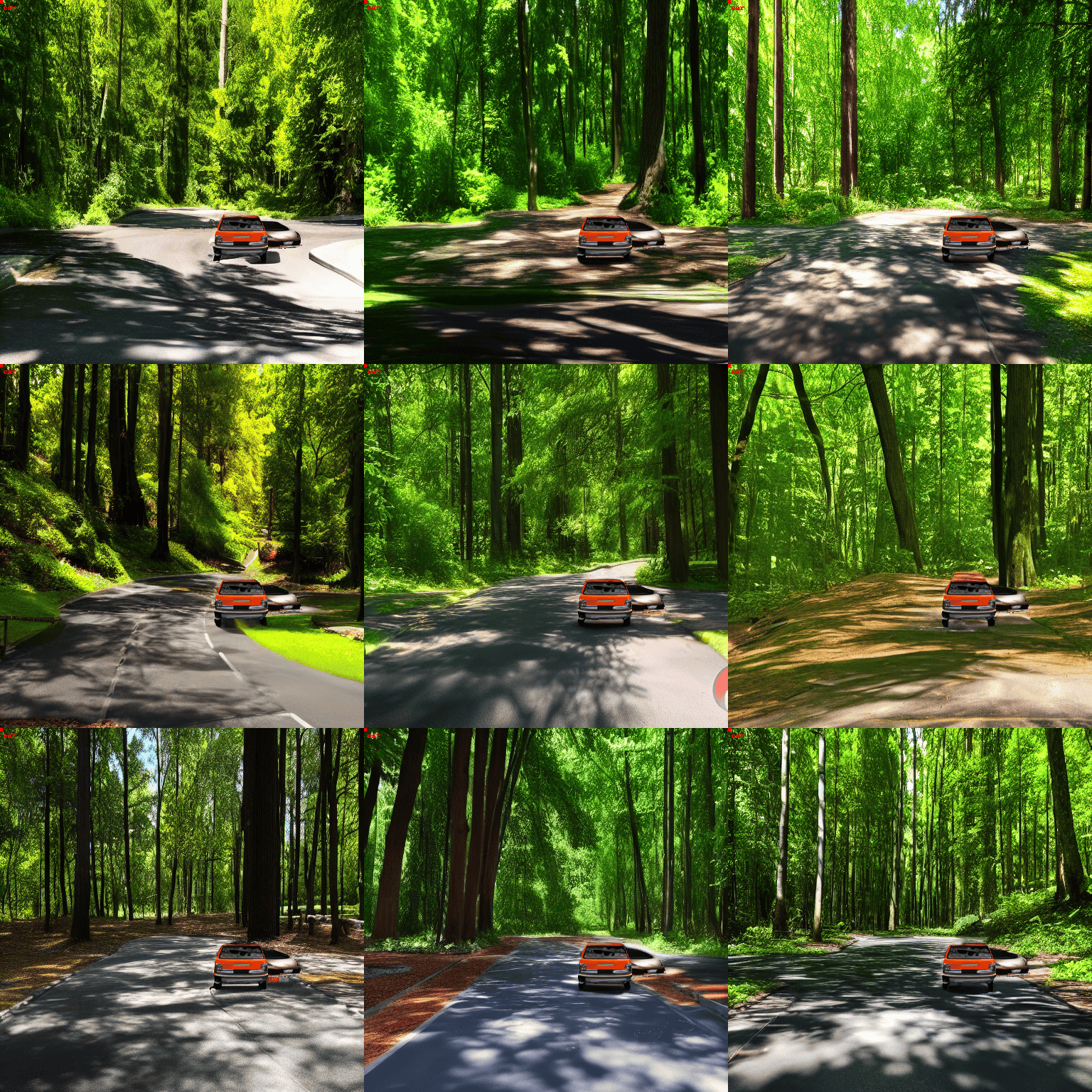}
\caption{\textbf{Randomly selected scene generated with BEV2EGO.}}
\label{fig:random_scenes_6}
\end{figure}

\begin{figure}[htb!]
\begin{flushleft}
\subsection{Randomly selected scenes with three cars}
In this section, in \Cref{fig:3_cars_random_scenes_1,fig:3_cars_random_scenes_2,fig:3_cars_random_scenes_3,fig:3_cars_random_scenes_4,fig:3_cars_random_scenes_5,fig:3_cars_random_scenes_6}, we display randomly selected scenes generated with BEV2EGO for all 9 seeds.
\end{flushleft}
\centering
\includegraphics[width=\linewidth]{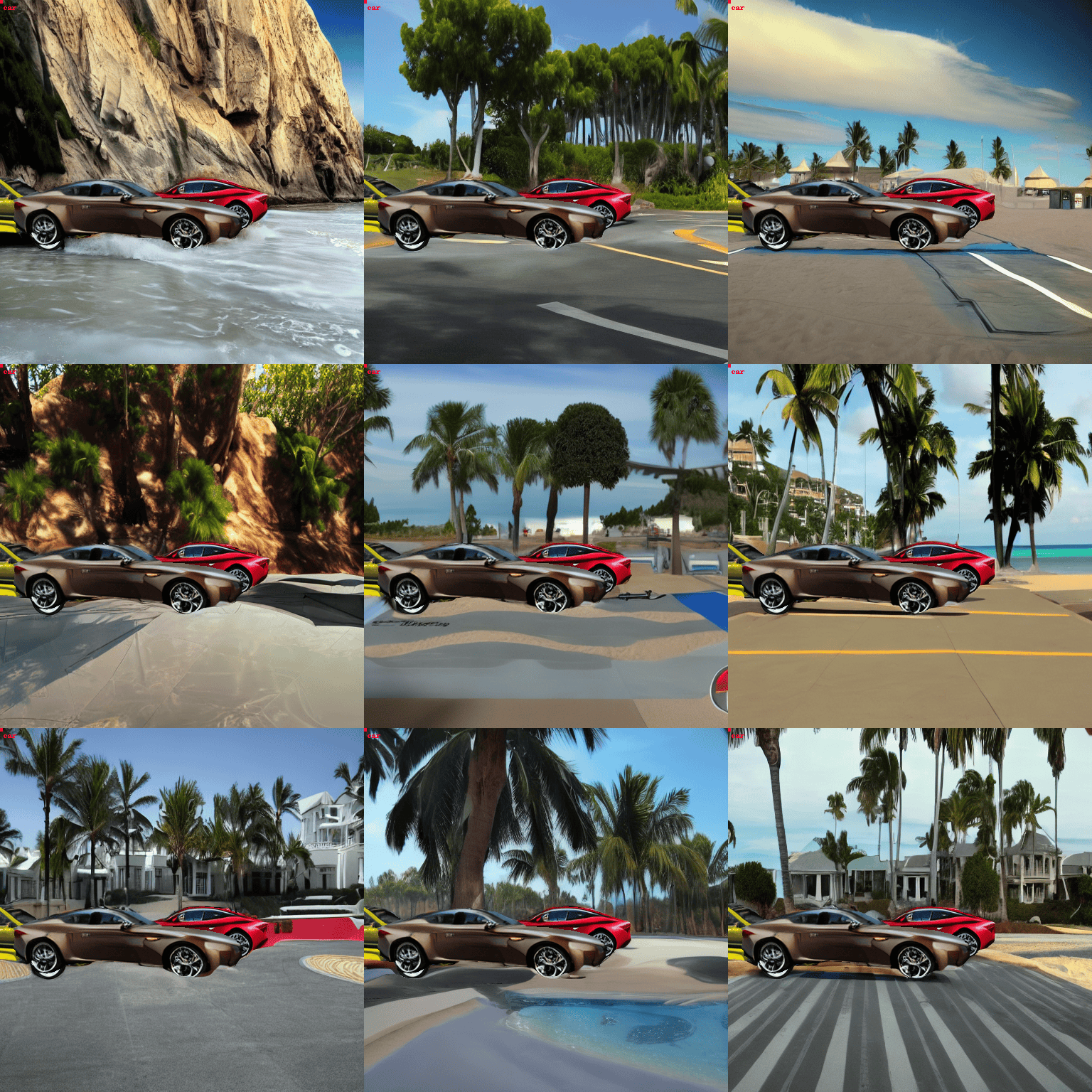}
\caption{\textbf{Randomly selected scene containing three cars generated with BEV2EGO.}}
\label{fig:3_cars_random_scenes_1}
\end{figure}

\begin{figure}[htb!]
\centering
\includegraphics[width=\linewidth]{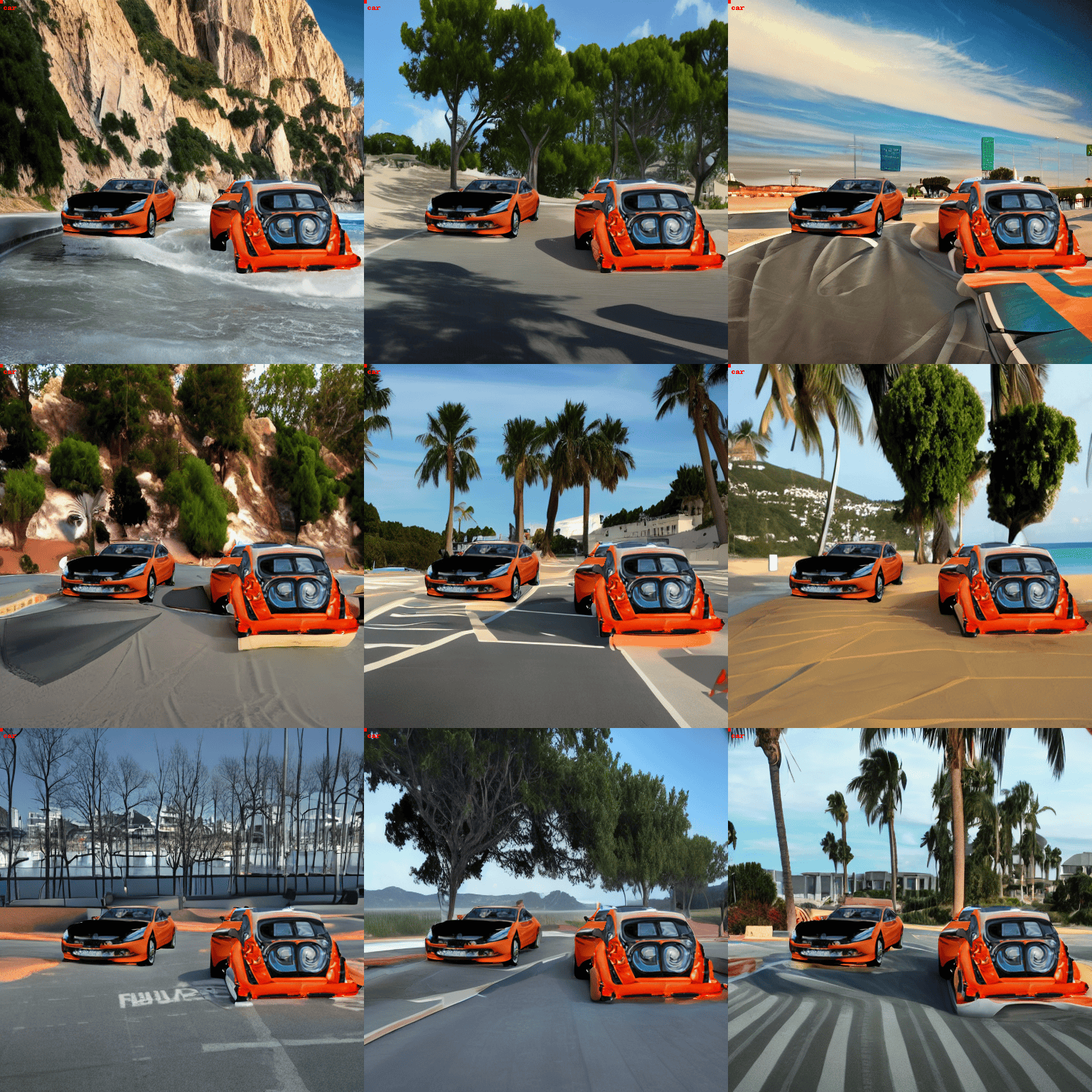}
\caption{\textbf{Randomly selected scene containing three cars generated with BEV2EGO.}}
\label{fig:3_cars_random_scenes_2}
\end{figure}

\begin{figure}[htb!]
\centering
\includegraphics[width=\linewidth]{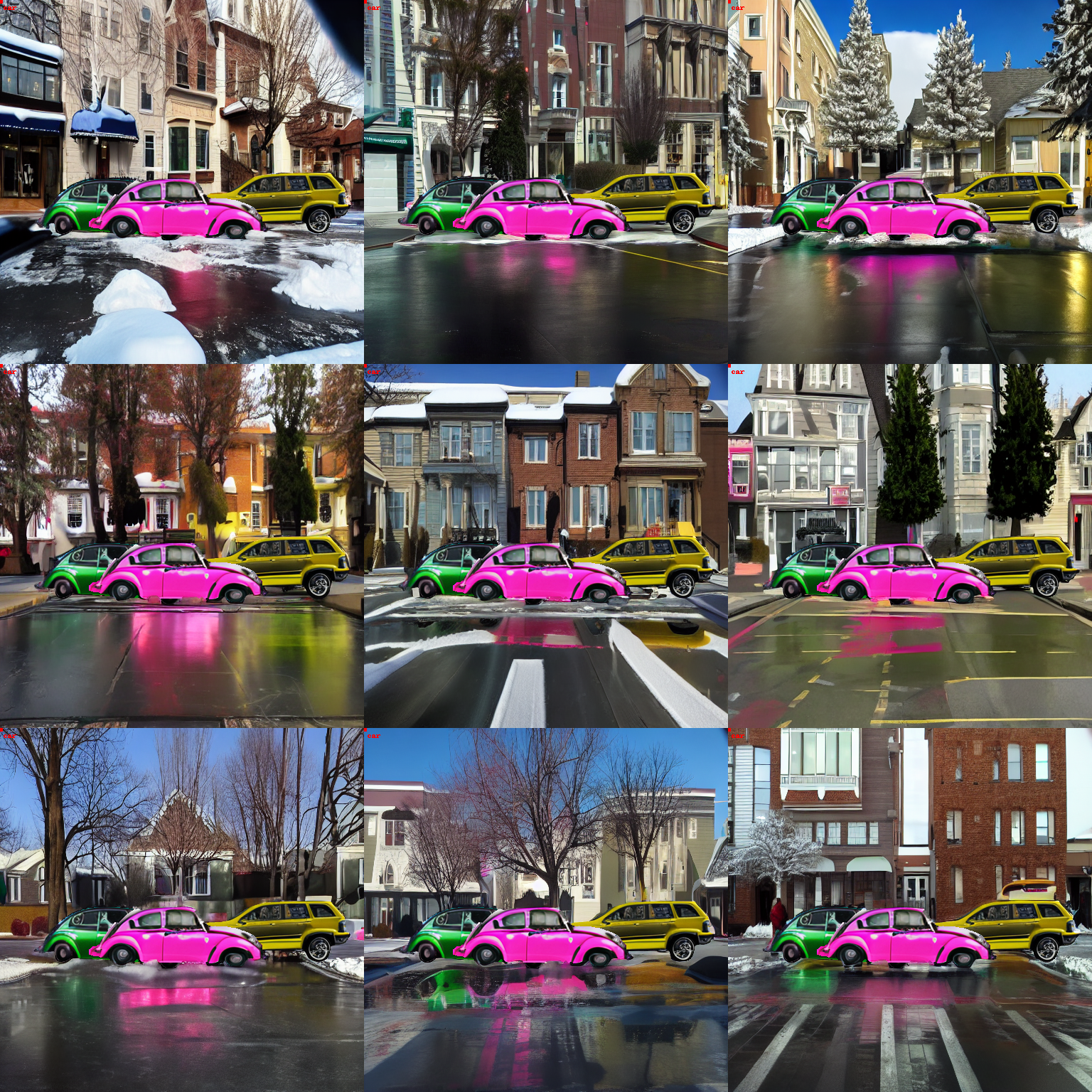}
\caption{\textbf{Randomly selected scene containing three cars generated with BEV2EGO.}}
\label{fig:3_cars_random_scenes_3}
\end{figure}

\begin{figure}[htb!]
\centering
\includegraphics[width=\linewidth]{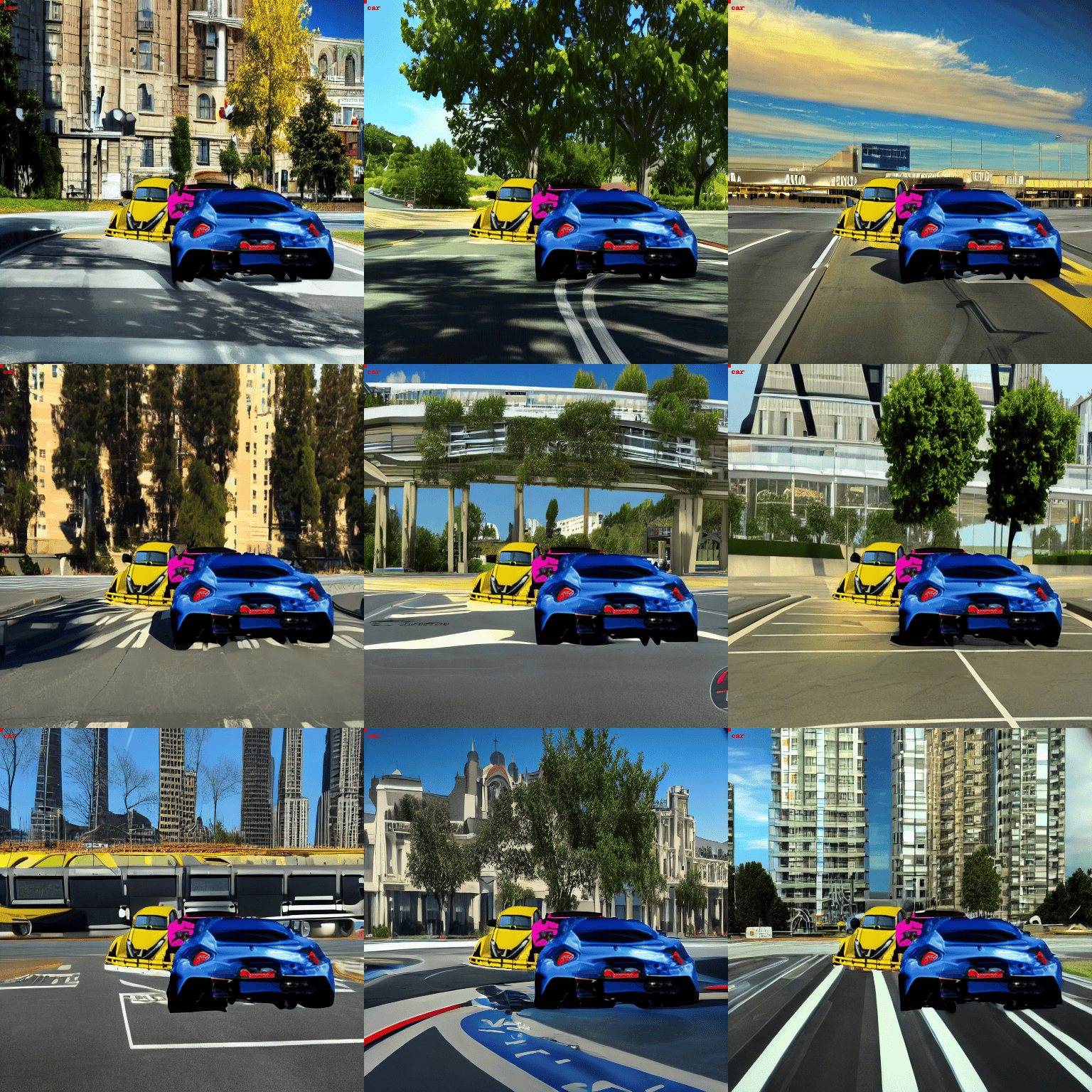}
\caption{\textbf{Randomly selected scene containing three cars generated with BEV2EGO.}}
\label{fig:3_cars_random_scenes_4}
\end{figure}

\begin{figure}[htb!]
\centering
\includegraphics[width=\linewidth]{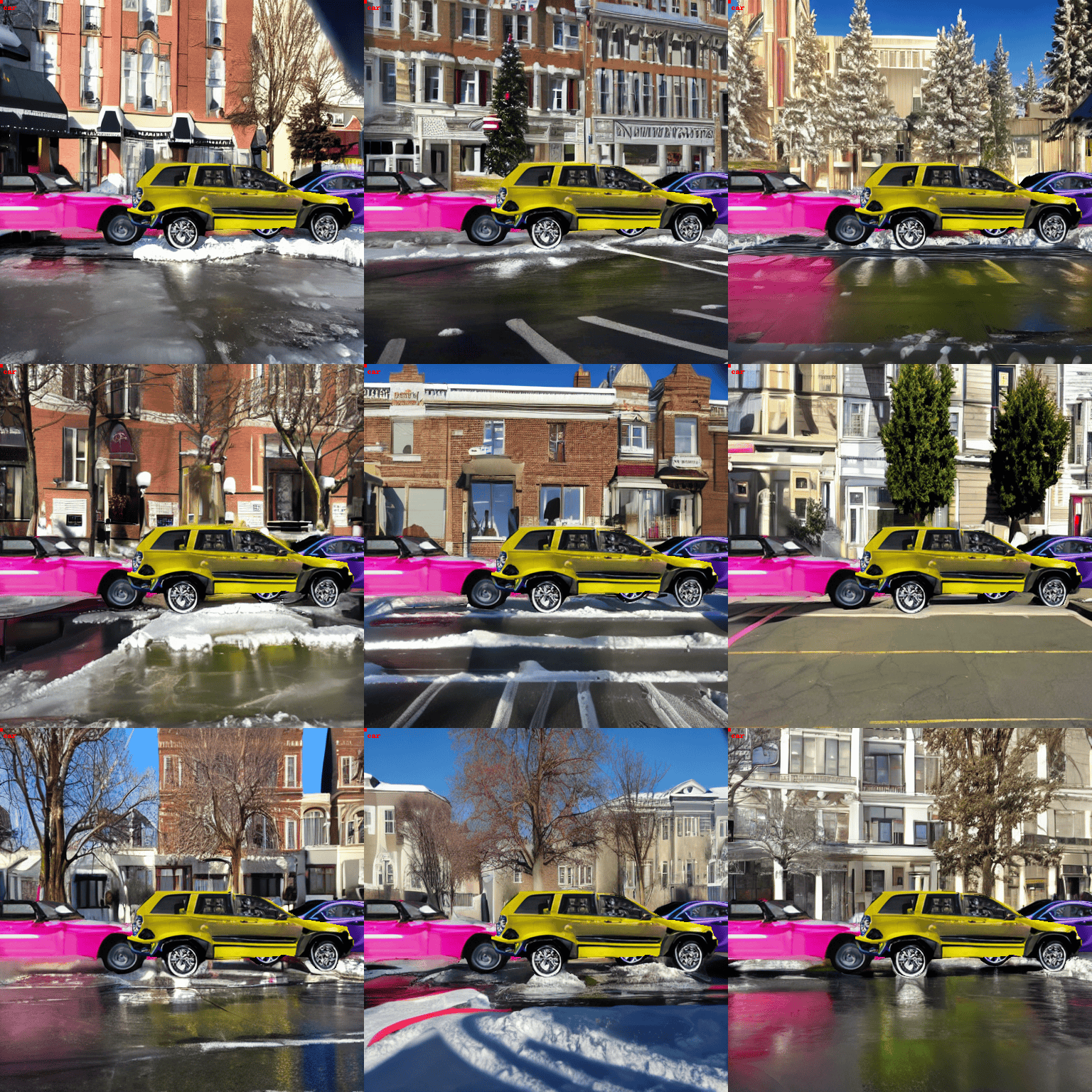}
\caption{\textbf{Randomly selected scene containing three cars generated with BEV2EGO.}}
\label{fig:3_cars_random_scenes_5}
\end{figure}

\begin{figure}[htb!]
\centering
\includegraphics[width=\linewidth]{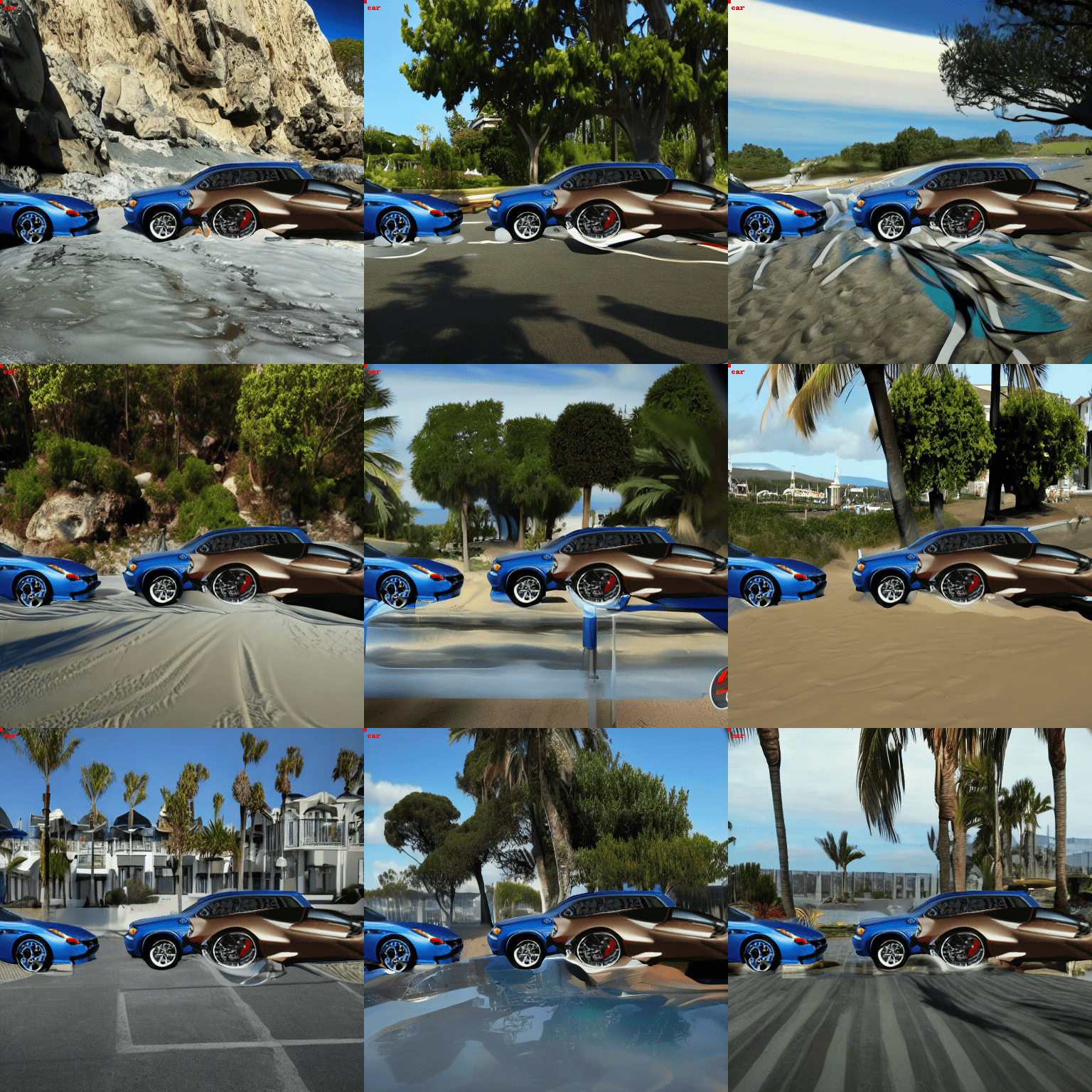}
\caption{\textbf{Randomly selected scene containing three cars generated with BEV2EGO.}}
\label{fig:3_cars_random_scenes_6}
\end{figure}


\begin{thebibliography}{10}
\providecommand{\url}[1]{\texttt{#1}}
\providecommand{\urlprefix}{URL }
\providecommand{\doi}[1]{https://doi.org/#1}

\bibitem{beyer2020imagenet}
Beyer, L., Hénaff, O.J., Kolesnikov, A., Zhai, X., van~den Oord, A.: Are we done with imagenet? arXiv:2006.07159  (2020)

\bibitem{bitterwolf2022classifiers}
Bitterwolf, J., Meinke, A., Boreiko, V., Hein, M.: Classifiers should do well even on their worst classes. In: ICML 2022 Shift Happens Workshop (2022), \url{https://openreview.net/forum?id=QxIXCVYJ2WP}

\bibitem{Blank_Hueger_Mock_Stauner_2022}
Blank, F., H{\"u}ger, F., Mock, M., Stauner, T.: Assurance methodology for in-vehicle {AI}. ATZ worldwide  \textbf{124},  54--59 (07 2022)

\bibitem{bordes2023pug}
Bordes, F., Shekhar, S., Ibrahim, M., Bouchacourt, D., Vincent, P., Morcos, A.S.: Pug: Photorealistic and semantically controllable synthetic data for representation learning. arXiv:2308.03977  (2022)

\bibitem{boreiko2023SCROD}
Boreiko, V., Hein, M., Metzen, J.H.: Identifying systematic errors in object detectors with the {SCROD} pipeline. In: ICCV Workshops (2023)

\bibitem{nuscenes}
Caesar, H., Bankiti, V., Lang, A.H., Vora, S., Liong, V.E., Xu, Q., Krishnan, A., Pan, Y., Baldan, G., Beijbom, O.: nuscenes: A multimodal dataset for autonomous driving. In: CVPR (2020)

\bibitem{chen2023integrating}
Chen, K., Xie, E., Chen, Z., Hong, L., Li, Z., Yeung, D.Y.: Geodiffusion: Text-prompted geometric control for object detection data generation. In: ICLR (2024)

\bibitem{cheng2021mask2former}
Cheng, B., Misra, I., Schwing, A.G., Kirillov, A., Girdhar, R.: Masked-attention mask transformer for universal image segmentation. In: CVPR (2022)

\bibitem{cheng2023layoutdiffuse}
Cheng, J., Liang, X., Shi, X., He, T., Xiao, T., Li, M.: Layoutdiffuse: Adapting foundational diffusion models for layout-to-image generation. arXiv:2302.08908  (2023)

\bibitem{Cordts2016Cityscapes}
Cordts, M., Omran, M., Ramos, S., Rehfeld, T., Enzweiler, M., Benenson, R., Franke, U., Roth, S., Schiele, B.: The cityscapes dataset for semantic urban scene understanding. In: CVPR (2016)

\bibitem{croce2021robustbench}
Croce, F., Andriushchenko, M., Sehwag, V., Debenedetti, E., Flammarion, N., Chiang, M., Mittal, P., Hein, M.: Robustbench: a standardized adversarial robustness benchmark. In: NeurIPS Datasets and Benchmarks Track (2021)

\bibitem{objaverseXL}
Deitke, M., Liu, R., Wallingford, M., Ngo, H., Michel, O., Kusupati, A., Fan, A., Laforte, C., Voleti, V., Gadre, S.Y., VanderBilt, E., Kembhavi, A., Vondrick, C., Gkioxari, G., Ehsani, K., Schmidt, L., Farhadi, A.: Objaverse-xl: A universe of 10m+ 3d objects. arXiv:2307.05663  (2023)

\bibitem{eyuboglu_domino:_2022}
Eyuboglu, S., Varma, M., Saab, K.K., Delbrouck, J.B., Lee-Messer, C., Dunnmon, J., Zou, J., Re, C.: Domino: {Discovering} {Systematic} {Errors} with {Cross}-{Modal} {Embeddings}. In: ICLR (2022)

\bibitem{gao2022adaptive}
Gao, I., Ilharco, G., Lundberg, S., Ribeiro, M.T.: Adaptive testing of computer vision models. arXiv:2212.02774  (2022)

\bibitem{gao2023magicdrive}
Gao, R., Chen, K., Xie, E., Hong, L., Li, Z., Yeung, D.Y., Xu, Q.: {MagicDrive}: Street view generation with diverse 3d geometry control. In: ICLR (2024)

\bibitem{DeepFashion2}
Ge, Y., Zhang, R., Wu, L., Wang, X., Tang, X., Luo, P.: A versatile benchmark for detection, pose estimation, segmentation and re-identification of clothing images. In: CVPR (2019)

\bibitem{geirhos2020shortcut}
Geirhos, R., Jacobsen, J.H., Michaelis, C., Zemel, R., Brendel, W., Bethge, M., Wichmann, F.A.: Shortcut learning in deep neural networks. Nature Machine Intelligence  \textbf{2}(11),  665--673 (2020)

\bibitem{geirhos2018}
Geirhos, R., Rubisch, P., Michaelis, C., Bethge, M., Wichmann, F.A., Brendel, W.: Imagenet-trained {CNN}s are biased towards texture; increasing shape bias improves accuracy and robustness. In: ICLR (2019)

\bibitem{hartley_zisserman_2004}
Hartley, R., Zisserman, A.: Multiple View Geometry in Computer Vision. Cambridge University Press (2004)

\bibitem{hendrycks2021manyfaces}
Hendrycks, D., Basart, S., Mu, N., Kadavath, S., Wang, F., Dorundo, E., Desai, R., Zhu, T., Parajuli, S., Guo, M., Song, D., Steinhardt, J., Gilmer, J.: The many faces of robustness: A critical analysis of out-of-distribution generalization. In: ICCV (2021)

\bibitem{hendrycks2018benchmarking}
Hendrycks, D., Dietterich, T.: Benchmarking neural network robustness to common corruptions and perturbations. In: ICLR (2019)

\bibitem{hu2022lora}
Hu, E.J., Shen, Y., Wallis, P., Allen-Zhu, Z., Li, Y., Wang, S., Wang, L., Chen, W.: Lo{RA}: Low-rank adaptation of large language models. In: ICLR (2022)

\bibitem{hu2023tifa}
Hu, Y., Liu, B., Kasai, J., Wang, Y., Ostendorf, M., Krishna, R., Smith, N.A.: Tifa: Accurate and interpretable text-to-image faithfulness evaluation with question answering. arXiv:2303.11897  (2023)

\bibitem{Idrissi2022ImageNetX}
Idrissi, B.Y., Bouchacourt, D., Balestriero, R., Evtimov, I., Hazirbas, C., Ballas, N., Vincent, P., Drozdzal, M., Lopez-Paz, D., Ibrahim, M.: Imagenet-x: Understanding model mistakes with factor of variation annotations. arXiv:2211.01866  (2022)

\bibitem{gansteerability}
Jahanian, A., Chai, L., Isola, P.: On the "steerability" of generative adversarial networks. In: ICLR (2020)

\bibitem{jain2023oneformer}
Jain, J., Li, J., Chiu, M., Hassani, A., Orlov, N., Shi, H.: {OneFormer: One Transformer to Rule Universal Image Segmentation} (2023)

\bibitem{jain_distilling_2022}
Jain, S., Lawrence, H., Moitra, A., Madry, A.: Distilling {Model} {Failures} as {Directions} in {Latent} {Space}. arXiv:2206.14754  (2022)

\bibitem{yolov5}
Jocher, G.: Yolov5 by ultralytics (2020). \doi{10.5281/zenodo.3908559}, \url{https://github.com/ultralytics/yolov5}

\bibitem{yolov8_ultralytics}
Jocher, G., Chaurasia, A., Qiu, J.: Ultralytics yolov8 (2023), \url{https://github.com/ultralytics/ultralytics}

\bibitem{kirillov2023segany}
Kirillov, A., Mintun, E., Ravi, N., Mao, H., Rolland, C., Gustafson, L., Xiao, T., Whitehead, S., Berg, A.C., Lo, W.Y., Doll{\'a}r, P., Girshick, R.: Segment anything. arXiv:2304.02643  (2023)

\bibitem{explaining_in_style}
Lang, O., Gandelsman, Y., Yarom, M., Wald, Y., Elidan, G., Hassidim, A., Freeman, W.T., Isola, P., Globerson, A., Irani, M., Mosseri, I.: Explaining in style: Training a gan to explain a classifier in stylespace. In: ICCV (2021)

\bibitem{leclerc2021three}
Leclerc, G., Salman, H., Ilyas, A., Vemprala, S., Engstrom, L., Vineet, V., Xiao, K., Zhang, P., Santurkar, S., Yang, G., Kapoor, A., Madry, A.: 3db: A framework for debugging computer vision models. arXiv:2106.03805  (2021)

\bibitem{li2022mplug}
Li, C., Xu, H., Tian, J., Wang, W., Yan, M., Bi, B., Ye, J., Chen, H., Xu, G., Cao, Z., et~al.: mplug: Effective and efficient vision-language learning by cross-modal skip-connections. arXiv:2205.12005  (2022)

\bibitem{li2023blip2}
Li, J., Li, D., Savarese, S., Hoi, S.: {BLIP-2:} bootstrapping language-image pre-training with frozen image encoders and large language models. In: ICML (2023)

\bibitem{li2023imagenete}
Li, X., Chen, Y., Zhu, Y., Wang, S., Zhang, R., Xue, H.: Imagenet-e: Benchmarking neural network robustness via attribute editing. In: CVPR (2023)

\bibitem{li2021benchmarking}
Li, Y., Xie, S., Chen, X., Dollar, P., He, K., Girshick, R.: Benchmarking detection transfer learning with vision transformers. arXiv:2111.11429  (2021)

\bibitem{li2023gligen}
Li, Y., Liu, H., Wu, Q., Mu, F., Yang, J., Gao, J., Li, C., Lee, Y.J.: Gligen: Open-set grounded text-to-image generation. In: CVPR (2023)

\bibitem{li_2021_discover}
Li, Z., Xu, C.: Discover the unknown biased attribute of an image classifier. In: ICCV (2021)

\bibitem{lin2015microsoft}
Lin, T.Y., Maire, M., Belongie, S., Bourdev, L., Girshick, R., Hays, J., Perona, P., Ramanan, D., Zitnick, C.L., Dollár, P.: Microsoft coco: Common objects in context. arXiv:1405.0312  (2015)

\bibitem{liu2023one}
Liu, M., Xu, C., Jin, H., Chen, L., Xu, Z., Su, H., et~al.: One-2-3-45: Any single image to 3d mesh in 45 seconds without per-shape optimization. arXiv:2306.16928  (2023)

\bibitem{liu2023zero1to3}
Liu, R., Wu, R., Hoorick, B.V., Tokmakov, P., Zakharov, S., Vondrick, C.: Zero-1-to-3: Zero-shot one image to 3d object. arXiv:2303.11328  (2023)

\bibitem{lv2023detrs}
Lv, W., Xu, S., Zhao, Y., Wang, G., Wei, J., Cui, C., Du, Y., Dang, Q., Liu, Y.: Detrs beat yolos on real-time object detection. arXiv:2304.08069  (2023)

\bibitem{torchvision2016}
maintainers, T., contributors: Torchvision: Pytorch's computer vision library. \url{https://github.com/pytorch/vision}

\bibitem{mao2023coco}
Mao, X., Chen, Y., Zhu, Y., Chen, D., Su, H., Zhang, R., Xue, H.: Coco-o: A benchmark for object detectors under natural distribution shifts. In: ICCV (2023)

\bibitem{Marathe_2023_CVPR}
Marathe, A., Ramanan, D., Walambe, R., Kotecha, K.: Wedge: A multi-weather autonomous driving dataset built from generative vision-language models. In: CVPR Workshops (2023)

\bibitem{Metzen_Systematic_Errors}
{Metzen}, J.H., {Hutmacher}, R., {Hua}, N.G., {Boreiko}, V., {Zhang}, D.: Identification of systematic errors of image classifiers on rare subgroups. In: ICCV (2023)

\bibitem{mou2023t2iadapter}
Mou, C., Wang, X., Xie, L., Wu, Y., Zhang, J., Qi, Z., Shan, Y., Qie, X.: T2i-adapter: Learning adapters to dig out more controllable ability for text-to-image diffusion models. arXiv:2302.08453  (2023)

\bibitem{MouradDAWN}
Mourad A.~Kenk, M.H.: Dawn: Vehicle detection in adverse weather nature dataset. \url{https://data.mendeley.com/datasets/766ygrbt8y/3}

\bibitem{yannic2023spurious}
Neuhaus, Y., Augustin, M., Boreiko, V., Hein, M.: Spurious features everywhere -- large-scale detection of harmful spurious features in imagenet. In: ICCV (2023)

\bibitem{Niemeyer2020GIRAFFE}
Niemeyer, M., Geiger, A.: Giraffe: Representing scenes as compositional generative neural feature fields. In: CVPR (2021)

\bibitem{peychev2023automated}
Peychev, M., Müller, M.N., Fischer, M., Vechev, M.: Automated classification of model errors on imagenet. In: NeurIPS (2023)

\bibitem{von-platen-etal-2022-diffusers}
von Platen, P., Patil, S., Lozhkov, A., Cuenca, P., Lambert, N., Rasul, K., Davaadorj, M., Wolf, T.: Diffusers: State-of-the-art diffusion models. \url{https://github.com/huggingface/diffusers} (2022)

\bibitem{qian2023magic123}
Qian, G., Mai, J., Hamdi, A., Ren, J., Siarohin, A., Li, B., Lee, H.Y., Skorokhodov, I., Wonka, P., Tulyakov, S., Ghanem, B.: Magic123: One image to high-quality 3d object generation using both 2d and 3d diffusion priors. arXiv:2306.17843  (2023)

\bibitem{pmlr-v139-radford21a}
Radford, A., Kim, J.W., Hallacy, C., Ramesh, A., Goh, G., Agarwal, S., Sastry, G., Askell, A., Mishkin, P., Clark, J., Krueger, G., Sutskever, I.: Learning transferable visual models from natural language supervision. In: ICML (2021)

\bibitem{Resnick2021CBERT_Wokrshop}
Resnick, C., Litany, O., Kar, A., Kreis, K., Lucas, J., Cho, K., Fidler, S.: Causal bert: Improving object detection by searching for challenging groups. In: ICCV Workshops (2021)

\bibitem{rombach2021highresolution}
Rombach, R., Blattmann, A., Lorenz, D., Esser, P., Ommer, B.: High-resolution image synthesis with latent diffusion models. In: CVPR (2022)

\bibitem{shen2020interpreting}
Shen, Y., Gu, J., Tang, X., Zhou, B.: Interpreting the latent space of gans for semantic face editing. In: CVPR (2020)

\bibitem{Shoshan_2021_ICCV}
Shoshan, A., Bhonker, N., Kviatkovsky, I., Medioni, G.: Gan-control: Explicitly controllable gans. In: ICCV (2021)

\bibitem{singla2022salient}
Singla, S., Feizi, S.: Salient imagenet: How to discover spurious features in deep learning? In: ICLR (2022)

\bibitem{Tong_Mass_Producing}
{Tong}, S., {Jones}, E., {Steinhardt}, J.: Mass-producing failures of multimodal systems with language models. arXiv:2306.12105  (2023)

\bibitem{tsipras2020imagenet}
Tsipras, D., Santurkar, S., Engstrom, L., Ilyas, A., Madry, A.: From imagenet to image classification: Contextualizing progress on benchmarks. In: ICML (2020)

\bibitem{vasudevan2022does}
Vasudevan, V., Caine, B., Gontijo-Lopes, R., Fridovich-Keil, S., Roelofs, R.: When does dough become a bagel? analyzing the remaining mistakes on imagenet. In: NeurIPS (2022)

\bibitem{wallace2023endtoend}
Wallace, B., Gokul, A., Ermon, S., Naik, N.: End-to-end diffusion latent optimization improves classifier guidance. In: ICCV (2023)

\bibitem{msssim}
Wang, Z., Simoncelli, E., Bovik, A.: Multiscale structural similarity for image quality assessment. In: Asilomar Conference on Signals, Systems \& Computers (2003)

\bibitem{wiles2022discovering}
Wiles, O., Albuquerque, I., Gowal, S.: Discovering bugs in vision models using off-the-shelf image generation and captioning. In: NeurIPS Workshops (2022)

\bibitem{Xu_2023_CVPR}
Xu, Y., Chai, M., Shi, Z., Peng, S., Skorokhodov, I., Siarohin, A., Yang, C., Shen, Y., Lee, H.Y., Zhou, B., Tulyakov, S.: Discoscene: Spatially disentangled generative radiance fields for controllable 3d-aware scene synthesis. In: CVPR (2023)

\bibitem{xue2023freestylenet}
Xue, H., Huang, Z., Sun, Q., Song, L., Zhang, W.: Freestyle layout-to-image synthesis. In: CVPR (2023)

\bibitem{bdd100k}
Yu, F., Chen, H., Wang, X., Xian, W., Chen, Y., Liu, F., Madhavan, V., Darrell, T.: Bdd100k: A diverse driving dataset for heterogeneous multitask learning. In: CVPR (2020)

\bibitem{zhang2023adding}
Zhang, L., Agrawala, M.: Adding conditional control to text-to-image diffusion models. arXiv:2302.05543  (2023)

\end{thebibliography}
\end{document}